\documentclass[Afour,sageh,times]{sagej}
\usepackage[inline]{enumitem}
\usepackage{amsmath}
\usepackage{amssymb,amsfonts}
\usepackage[ruled,vlined]{algorithm2e}
\usepackage{graphicx}
\usepackage{subfloat}
\usepackage{textcomp}
\usepackage{xcolor}
\usepackage{verbatim} 
\usepackage{mathrsfs}
\usepackage{subfigure}
\usepackage{empheq}
\usepackage{courier}
\usepackage{lipsum}
\usepackage{comment}
\usepackage{gensymb}
\usepackage{mathtools}
\usepackage{bm}
\usepackage{ wasysym } 
\usepackage{soul}
\usepackage[table]{xcolor}

\usepackage[colorlinks,bookmarksopen,bookmarksnumbered,citecolor=red,urlcolor=red]{hyperref}

\newcommand\BibTeX{{\rmfamily B\kern-.05em \textsc{i\kern-.025em b}\kern-.08em
T\kern-.1667em\lower.7ex\hbox{E}\kern-.125emX}}

\long\def\revised#1{{\color{black}#1}}

\newcommand{\call}{\mathcal}

\def\cA{{\call A}}
\def\cB{{\call B}}

\def\cI{{\call I}}

\def\cO{{\call O}}

\def\cX{{\call X}}
\def\cY{{\call Y}}
\def\cZ{{\call Z}}

\newcommand{\naturals}{\mathbb{N}}

\newcommand{\ap}{AP}
\newcommand{\true}{{\tt{True}}}
\newcommand{\false}{{\tt{False}}}
\newcommand{\var}{\pi}
\newcommand{\inp}{\cI}

\newcommand{\inpcen}{\inp_{\textrm{robot}}}
\newcommand{\inpcentrans}{\inp_{\textrm{robot}}^{\textrm{trans}}}
\newcommand{\inpcenorien}{\inp_{\textrm{robot}}^{\textrm{orien}}}
\newcommand{\inpterrain}{\inp_{\textrm{terrain}}}
\newcommand{\inpterrainbool}{\inp_{\textrm{terrain}}^{\cB}}

\newcommand{\inpstatesterrainexprposs}{\dom_{\textrm{terrain}}^{\textrm{poss}}}

\newcommand{\inprequest}{\inp_{\textrm{req}}}
\newcommand{\inprequesttrans}{\inp_{\textrm{req}}^{\textrm{trans}}}
\newcommand{\inprequestorien}{\inp_{\textrm{req}}^{\textrm{orien}}}
\newcommand{\tvar}[2]{n_{#1}^{#2}}

\newcommand{\xcells}{\cX}
\newcommand{\ycells}{\cY}
\newcommand{\out}{\cO}
\newcommand{\skillcustom}[1]{\skill_{#1}}
\newcommand{\skillprecustom}[1]{\Sigma_{#1}^{\textrm{pre}}}
\newcommand{\skillpostcustom}[1]{\Sigma_{#1}^{\textrm{post}}}
\newcommand{\skill}{o}
\newcommand{\skillpre}{\skillprecustom{\skill}}
\newcommand{\skillpost}{\skillpostcustom{\skill}}
\newcommand{\skillpretrans}{\skillprecustom{\skill_{\textrm{trans}}}}
\newcommand{\skillposttrans}{\skillpostcustom{\skill_{\textrm{trans}}}}
\newcommand{\skillpreorien}{\skillprecustom{\skill_{\textrm{orien}}}}
\newcommand{\skillpostorien}{\skillpostcustom{\skill_{\textrm{orien}}}}

\newcommand{\dom}{\Sigma}
\newcommand{\statevar}{\sigma}
\newcommand{\inpstate}{\statevar_\inp}
\newcommand{\inpstates}{\dom_\inp}
\newcommand{\inpstateterrain}{\statevar_{\textrm{terrain}}}
\newcommand{\inpstateterrainbool}{\statevar_{\textrm{terrain}}^{\cB}}

\newcommand{\inpstatesterrain}{\dom_{\textrm{terrain}}}

\newcommand{\inpstatecen}{\statevar_{\textrm{robot}}}
\newcommand{\inpstatescen}{\dom_{\textrm{robot}}}
\newcommand{\inpstaterequest}{\statevar_{\textrm{req}}}
\newcommand{\inpstatesrequest}{\dom_{\textrm{req}}}
\newcommand{\inpstatesrequestposs}{\dom_{\textrm{req}}^{\textrm{poss}}}

\newcounter{examplecounter}
\newcommand{\examplelabel}[1]{%
  \refstepcounter{examplecounter}%
  \label{#1}%
  \theexamplecounter%
}

\newcounter{problemcounter}
\newcommand{\problemlabel}[1]{%
  \refstepcounter{problemcounter}%
  \label{#1}%
  \theproblemcounter%
}

\newcounter{defcounter}
\newcommand{\deflabel}[1]{
    \refstepcounter{defcounter}%
    \label{#1}%
    \thedefcounter%
}

\newcommand{\envspec}{\varphi_\textrm{e}}

\newcommand{\sysspec}{\varphi_\textrm{s}}
\newcommand{\envinit}{\varphi_\textrm{e}^\textrm{i}}
\newcommand{\sysinit}{\varphi_\textrm{s}^\textrm{i}}
\newcommand{\alphainit}{\varphi_\alpha^\textrm{i}}
\newcommand{\envsafety}{\varphi_\textrm{e}^\textrm{t}}

\newcommand{\alphasafety}{\varphi_\alpha^\textrm{t}}

\newcommand{\envsafetynothard}{\varphi_\textrm{e}^\textrm{t,skill}}
\newcommand{\envsafetyhard}{\varphi_\textrm{e}^\textrm{t,hard}}

\newcommand{\syssafety}{\varphi_\textrm{s}^\textrm{t}}
\newcommand{\syssafetynothard}{\varphi_\textrm{s}^\textrm{t,skill}}
\newcommand{\syssafetyhard}{\varphi_\textrm{s}^\textrm{t,hard}}

\newcommand{\envlive}{\varphi_\textrm{e}^\textrm{g}}
\newcommand{\syslive}{\varphi_\textrm{s}^\textrm{g}}
\newcommand{\alphalive}{\varphi_\alpha^\textrm{g}}

\newcommand{\notallowedrepair}{\varphi_\textrm{disallow}}
\newcommand{\disallowedtrans}{\notallowedrepair}

\newcommand{\spec}{\varphi}

\newcommand{\range}[1]{[#1]}

\newcommand{\strategy}{\cA}

\newcommand{\grounding}{\textrm{G}}
\newcommand{\inversegrounding}{\grounding^{-1}}
\newcommand{\ps}{\cZ}
\newcommand{\physicalstate}{z}
\newcommand{\reals}{\mathbb{R}}

\newcommand{\vg}[1]{\bm{#1}}
\renewcommand{\v}[1]{\mathbf{#1}}

\def\volumeyear{2026}

\begin{document}

\runninghead{Zhou et al.}

\title{Physically-Feasible Reactive Synthesis for Terrain-Adaptive Locomotion}

\author{Ziyi Zhou\affilnum{1}, Qian Meng\affilnum{2}, Hadas Kress-Gazit\affilnum{2}, and Ye Zhao\affilnum{1}}

\affiliation{\affilnum{1}Georgia Institute of Technology, Atlanta, GA, USA\\
\affilnum{2}Cornell University, Ithaca, NY, USA}

\corrauth{Ye Zhao, Georgia Institute of Technology, Atlanta, GA 30332, USA.}

\email{ye.zhao@me.gatech.edu}

\setcounter{secnumdepth}{3}

\begin{abstract}
We present an integrated planning framework for quadrupedal locomotion over \revised{discrete, unforeseen terrain configurations revealed at runtime}.
Existing methods often depend on heuristics for real-time foothold selection---limiting robustness and adaptability---or rely on computationally intensive trajectory optimization across complex terrains and long horizons. In contrast, our approach combines reactive synthesis for generating correct-by-construction symbolic-level controllers with mixed-integer convex programming (MICP) for physically feasible footstep planning during each symbolic transition. To reduce the reliance on costly MICP solves and accommodate specifications that may be violated due to physical infeasibility, we adopt a symbolic repair mechanism that selectively generates only the required symbolic transitions. During execution, real-time MICP replanning based on actual terrain data, runtime symbolic repair, \revised{kinematic-feasibility re-targeting, and a delay-aware coordination scheme} enable seamless bridging between offline synthesis and online operation. Through extensive simulation and hardware experiments, we validate the framework's ability to identify missing locomotion skills and respond effectively in safety-critical environments, including scattered stepping stones and rebar scenarios. Code and experimental videos are available at \url{https://synthesis-micp.github.io/}.
\end{abstract}


\keywords{Legged Robots, Optimization and Optimal Control,
Formal Methods in Robotics and Automation}

\maketitle

\section{Introduction}
Terrain-adaptive locomotion is essential for enhancing the mobility of legged robots and moving beyond blind walking strategies \citep{di2018dynamic,kim2019highly, zhou2022momentum}. Many existing methods achieve terrain adaptability by adjusting footholds around a nominal trajectory in real time \citep{fankhauser2018robust,grandia2023perceptive}. To further boost traversability, some approaches have explored variable gait patterns. In \citep{Park2015OnlinePF, kim2020vision,gilroy2021autonomous}, jumping motions are precomputed offline and triggered heuristically upon detecting obstacles. Other methods embed gait switching—such as transitions from walking to jumping—within simplified dynamics models during trajectory sampling \citep{norby2020fast}, or rely on pre-trained classifiers to assess feasibility \citep{chignoli2022rapid}. However, despite the importance of safety in high-risk settings such as disaster zones and construction sites, relatively few works address formal safety guarantees for agile, variable-gait locomotion \citep{zhao2022reactive,shamsah2023integrated} that fully account for the robot's physical limitations.

\begin{figure}[t]
    \centering
    \includegraphics[width=1\linewidth]{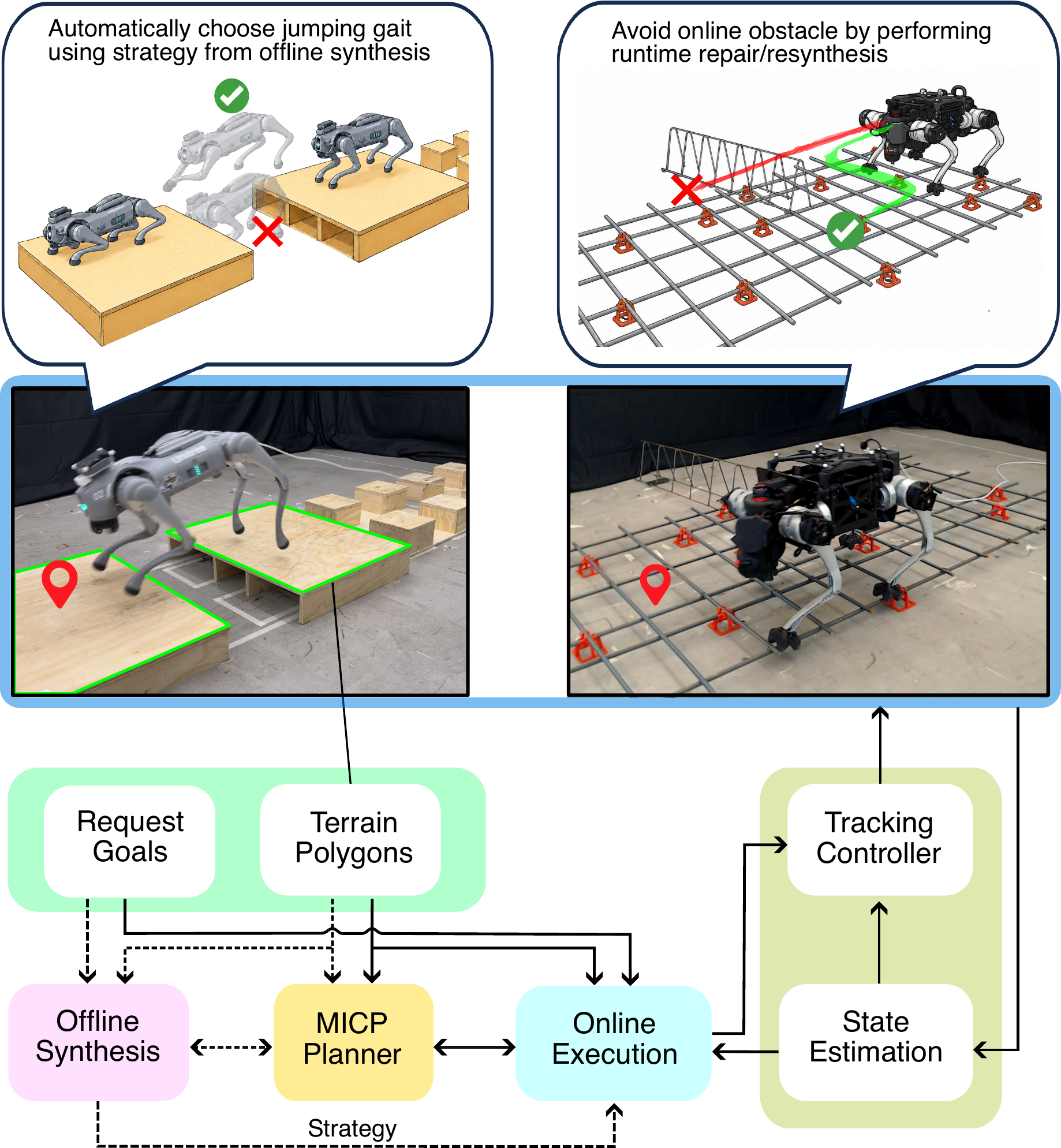}
    \caption{System architecture overview. The solid lines indicate online communication, while dashed lines represent offline processes.
    The request goals in the offline phase are user-defined, while those in the online phase are provided by a global planner based on the global goal.
    }
    \label{fig:system_diagram}
\end{figure}

Mixed-integer programming (MIP)-based methods \citep{valenzuela2016mixed,shirai2022simultaneous} model both contact state and contact surface selection for each leg and timestep as binary decision variables. This allows them to explicitly capture the interaction between terrain geometry and the robot's kinematic and dynamic constraints \citep{deits2014footstep}. When dynamics and constraints are suitably relaxed, the resulting convex approximation (MICP) provides a global certificate of feasibility upon convergence \citep{dai2019global}, making it a powerful tool for formally assessing locomotion viability.
However, the computational demands of MIP-based approaches scale poorly with increased terrain complexity and longer planning horizons, posing a major hurdle for real-time applications. To mitigate this, various simplifications have been proposed, such as fixing contact sequences \citep{ding2020kinodynamic,fey20243d}, constraining timing \citep{ponton2016convex,risbourg2022real,acosta2023bipedal}, or limiting the number of footsteps \citep{aceituno2017simultaneous}. Despite these efforts, scalability remains a challenge in cluttered environments with various terrain segments and features.

To simplify navigation across complex terrains, one effective strategy is to decompose long-horizon, global tasks into a sequence of localized, robot-centric environments centered around intermediate waypoints \citep{fankhauser2018robust}. Despite this simplification, the associated MIP formulation still faces computational challenges, as it must account for all nearby terrain elements.
To address this, we discretize the local environment and constrain each MIP to solving only a single discrete transition at a time. These discrete transitions are then composed using a Linear Temporal Logic (LTL)-based reactive synthesis framework \citep{bloem2012synthesis}, which enables us to systematically construct a correct-by-construction strategy that guides the robot through local decisions toward achieving its intermediate goals.
\revised{For readers unfamiliar with formal methods, LTL is a logic for expressing temporal properties of system behaviors using operators such as ``always,'' ``eventually,'' and ``until''~\citep{piterman2006synthesis,bloem2012synthesis}; reactive synthesis automatically constructs a controller (named as a \emph{strategy automaton} in this paper) that satisfies an LTL specification against any admissible environment behavior, providing correct-by-construction guarantees at the symbolic level.}

In this work, we present an integrated planning framework that unifies reactive synthesis with MIP-based optimization to enable legged robots to respond safely and efficiently to \revised{unforeseen terrain configurations revealed at runtime}. We abstract both the continuous robot states and local environmental context into symbolic representations, allowing the robot to make high-level decisions in real time for negotiating complex obstacles such as large gaps or cluttered terrain (see Fig.~\ref{fig:system_diagram}). Each symbolic transition is validated through an MICP to ensure the physical feasibility of the locomotion process. This architecture not only connects high-level symbolic reasoning with low-level dynamic feasibility, but also mitigates the computational challenges of solving long-horizon MICP problems.
By leveraging guidance from the LTL-based symbolic planner, each MICP instance only needs to handle short horizons and a limited set of terrain features, which substantially reduces computational load.
The framework is two-stage: in the offline synthesis stage, we employ \textit{gait-free} MICP formulations that jointly optimize over contact mode and foothold placement to precompute a diverse set of locomotion skills for symbolic transitions. During online deployment, we switch to lightweight \textit{gait-fixed} MICP formulations using these precomputed gaits, enabling real-time generation of physically consistent motions with real-world terrain data.

\revised{The two-stage architecture exposes two scalability challenges that we address with two complementary mechanisms. First, the full task specification has a state space that grows exponentially with the number of variables encoding the surrounding terrains, so direct synthesis is computationally impractical~\citep{bloem2012synthesis}; to address this, we propose a high-level manager that fixes the local terrain and goal information and only retains the skills needed for the current local environment, yielding an $80.9$--$90.1\%$ reduction in the symbolic variable count in our experiments. Second, gait-free MICP---used to certify dynamic feasibility---is itself computationally expensive; we therefore use a symbolic repair process that suggests only the missing transitions whose feasibility must be checked, yielding a $71.6$--$97.6\%$ reduction in costly gait-free MICP solves relative to exhaustive enumeration. During online execution, the gait-fixed MICP is re-solved against the actual perceived terrain to generate a new nominal trajectory, and runtime symbolic repair handles any unforeseen terrain that the offline phase did not cover.}

\revised{Closing the loop between this synthesis-MICP framework and a real robot poses two practical challenges that we address through two additional online-execution mechanisms. (i) Online MICP solve times are inherently variable, ranging from hundreds of milliseconds to several seconds, and can exceed the trajectory horizon used by the tracking controller. We propose a \emph{delay-aware coordination} scheme that asynchronously appends or stalls the tracking trajectory based on a formal two-case analysis of the per-transition timing budget; without it, the closed loop between the discrete strategy automaton and the continuous-time tracker is ill-defined under bounded solve time. (ii) The online MICP tends to become infeasible when the offline-checked target pose mismatches the actual terrain. We add a lightweight, zero-horizon \emph{kinematic-feasibility re-targeting} step that pre-checks and adjusts the desired pose at each transition, decoupling offline pose specification from online terrain geometry; the same online MICP is also augmented with additional collision-avoidance constraints for swing-foot clearance during a transition. Together with the offline scalability machinery, these mechanisms make the framework end-to-end deployable, which we validate on two robot hardware platforms and benchmark against both a pure-MIP planner and a heuristic foothold planner.}

The proposed framework remains flexible and modular---it can be driven by commands from a higher-level planner \citep{wermelinger2016navigation, fernbach2017kinodynamic,norby2020fast,chignoli2022rapid,liao2023walking,feng2023gpf} or directly from a user, and it easily accommodates new behaviors by expanding the symbolic transition set.

The main contributions of this work are summarized as follows:
\begin{itemize}
    \item \textbf{Integrated reactive-synthesis and MICP framework for terrain-adaptive locomotion.}
    Symbolic-level guidance limits the planning horizon and the number of terrain features any single MICP must consider, bridging high-level symbolic reasoning with low-level physical feasibility.

    \item \textbf{Scalable offline synthesis via a high-level manager and symbolic repair with dynamic feasibility checks.}
    The high-level manager reduces the symbolic state space by $80.9$--$90.1\%$, and symbolic repair, equipped with dynamic feasibility checks via gait-free MICP, generates only the necessary skills---reducing costly gait-free MICP solves by $71.6$--$97.6\%$ versus exhaustive enumeration.

    \item \textbf{Online execution module that bridges offline synthesis and real-world deployment.}
    The module re-solves a gait-fixed MICP against the actual perceived terrain at each symbolic transition, with \revised{delay-aware coordination between the strategy automaton and the tracking controller, online kinematic-feasibility re-targeting of the desired pose, and additional collision-avoidance constraints in the online MICP. Runtime symbolic repair handles previously unseen scenarios that were not considered offline.}

    \item \revised{\textbf{Comprehensive experimental validation.} We report simulation across four terrain scenarios with full per-module timing, comparisons against a pure MIP planner and a heuristic foothold planner, hardware deployment of the integrated stack on two robot platforms (Unitree Go2 and SkyMul Chotu), and a feasibility study of yaw orientation as a generalization case study.}
\end{itemize}

A conference version of this work is presented in \citep{zhou2025physically}. Beyond the conference paper, the current article contributes the delay-aware coordination and online re-targeting mechanisms motivated above, full hardware validation of the integrated framework on two robot platforms, systematic benchmarking against a pure-MIP planner and a heuristic foothold planner, a preliminary yaw-extension feasibility study, and substantially expanded preliminaries, related-work survey, and discussion of limitations.


\section{Related Work}
\subsection{Hierarchical Planning for Terrain-Adaptive Locomotion}
Planning for terrain-adaptive locomotion can be boiled down into solving contact sequence and timing (equivalently gait), location (equivalently foothold), and robot motion (e.g., Center of Mass (CoM) or base pose), given the chosen kinematics/dynamics model and the perceived terrain information. Hierarchical approaches first separately or simultaneously address part of the above problems and then solve for the rest, which allows for higher computational efficiency.  
Kinematic footstep planning focuses on the first two problems, neglecting the robot dynamics through graph-based approaches \citep{kolter2008control,kalakrishnan2010fast,winkler2014path,mastalli2015line,fankhauser2018robust,griffin2019footstep} or optimization-based approaches \citep{deits2014footstep,tonneau2020sl1m,risbourg2022real, ren2025accelerating}.
Although non-fixed gait/contact plans can be generated for very rough terrains \citep{hauser2005non,bretl2006motion,tonneau2018efficient}, conservative quasi-static motions are generated for challenging terrains due to the kinematic simplification during  footstep planning. 

Given the recent advances in optimization-based approaches, notably Model Predictive Control (MPC), another focus of the literature is on the local foothold adaptation and the integration of robot dynamics for dynamic locomotion.
Instantaneous foothold adaptation around a nominal location, with a fixed and cyclic gait, is performed based on heuristic search \citep{jenelten2020perceptive,kim2020vision,agrawal2022vision} or learning approaches \citep{magana2019fast, villarreal2020mpc,yu2021visual,gangapurwala2022rloc, wu2025learn, Kamohara2025RLMPC}. 
In \citep{grandia2021multi,jenelten2022tamols,grandia2023perceptive}, the base pose and foothold are optimized jointly, demonstrating impressive terrain traversing capabilities on rough terrains. More recently, MPC has been employed within hierarchically integrated planning frameworks for legged navigation over rough terrain, incorporating explicit terrain uncertainty quantification \citep{muenprasitivej2024bipedal, shamsah2025terrain, muenprasitivej2026bipedal}.

However, the above dynamic locomotion works consider a fixed gait pattern that usually prohibits achieving versatile behaviors, such as jumping.
In \citep{Park2015OnlinePF, kim2020vision,gilroy2021autonomous}, jumping motions are generated offline and triggered through heuristics.
In \citep{fernbach2017kinodynamic,norby2020fast,chignoli2022rapid}, RRT-Connect is used for rapid kino-dynamic locomotion planning, and the switch between normal walking and jumping is either embedded inside a reduced-order model during sampling \citep{norby2020fast} or decided by a pre-trained feasibility classifier \citep{chignoli2022rapid}.
Beyond limited gait modes, other works focus on online gait planning and/or foothold selection and deploy MPC to track the plan,
which can be further categorized into model-free \citep{da2021learning,yang2022fast,xie2022glide,pmlr-v164-margolis22a,yu2024learning} and model-based methods \citep{boussema2019online,wang2023and,sun2023free}.

\revised{A complementary line of work uses end-to-end reinforcement learning (RL) to train perceptive locomotion policies that directly map terrain observations to joint commands, achieving impressive robustness on stairs, gaps, and unstructured outdoor terrain~\citep{miki2022learning,agarwal2023legged,zhuang2023robot,hoeller2023anymal,cheng2024extreme,luo2024pie}. Compared with these data-driven approaches, our model-based, formal-methods framework offers explicit symbolic-level realizability guarantees and physical-feasibility certificates from MICP, at the cost of weaker generalization and the need for a terrain abstraction; we view the RL approaches and ours as complementary rather than competing.}

\subsection{Simultaneous Planning via Optimization}
Planning for terrain-adaptive locomotion problem can also be formulated as a combinatorial optimization problem that simultaneously optimizes for gait, foothold, and robot dynamics.
One approach is to incorporate rigid \citep{posa2014direct} or smoothed \citep{mordatch2012discovery, drnach2021robust} complementarity constraints, which accurately model the physical contact interaction behavior but remains computationally inefficient due to the non-smoothness and high nonlinearity. Promising online contact-implicit MPC results are reported in \citep{aydinoglu2022real,le2024fast, drnach2022mediating} but still limited to low-dimensional system or static environment. 

Another way of encoding discrete contact decisions is to treat both the contact state and the contact plane selection for each leg at each timestep as binary variables \citep{valenzuela2016mixed,shirai2022simultaneous,jiang2023locomotion}. The underlying problem can be formulated as Mixed Integer Program (MIP). Convex approximations of nonlinear dynamics, usually centroidal dynamics, and constraints are typically required to transform the MIP into a more tractable Mixed Integer Convex Program (MICP), albeit with an increased number of binary variables. 
A global certificate for the approximated convex problem exists upon convergence \citep{dai2019global}, providing an ideal means to determine the feasibility in our proposed method.
To relieve the computational burden due to the introduction of many binary variables, the works of \citep{ponton2016convex,ding2020kinodynamic,acosta2023bipedal} assume the contact sequence and timing are chosen \textit{a priori}, and only optimize the contact plane selection. Aceituno-Cabezas et al. \citep{aceituno2017simultaneous} optimize the contact sequence within a certain gait cycle that fixes the number of footsteps. 
However, it is inevitable that the problem complexity grows exponentially when the number of discrete contact options increases along with the time horizon and terrain features. Our work aims to alleviate computational burden through a two-stage approach. In the offline phase, expensive MICPs that optimize both contact state and foothold selection are solved to generate necessary locomotion gaits. In the online phase, efficient MICPs with adaptive and predefined gaits are used to produce dynamically feasible motions and footholds. Additionally, guided by a synthesis-based task planner, each MICP in our work considers shorter time horizons and fewer terrain features compared to solving a single large MICP problem.

\subsection{Task and Motion Planning for Contact-Rich Planning}
Task and Motion Planning (TAMP) formally defines symbolic-level tasks and searches through a graph of predefined motion primitives that enable feasible symbolic transitions \citep{garrett2021integrated, zhao2024survey}. For more complex physical contact resoning, Toussaint et al. propose Logic Geometric Programming (LGP) \citep{toussaint2015logic} and embed the high-level
logic representation into the low-level motion planner, demonstrating contact-rich tool-use behaviors \citep{toussaint2018differentiable}
after defining abundant action primitives. 
Building on this, a broader range of motions has been showcased, including versatile manipulation \citep{migimatsu2020object,zhao2021sydebo} and loco-manipulation tasks \citep{sleiman2023versatile}.
In a similar vein, works such as \citep{ding2021hybrid,shirai2021lto,jelavic2021combined,amatucci2022monte,chen2021trajectotree,zhang2023simultaneous,zhu2023efficient,asselmeier2024hierarchical} integrate graph-search or sampling-based methods with optimization-based approaches in a holistic manner, abstracting contact modes within a discrete domain. However, the combinatorial nature of these problems often leads to poor scalability due to the explosion of contact modes. Deploying such approaches online in dynamic environments remains an open challenge. From a different perspective, we abstract a smaller set of task-level contact modes as locomotion gaits over a specific time horizon and distance, allowing the MICP to solve for details such as contact locations and robot motion during execution. This shifts the focus to selecting locomotion gaits within a local environment while ensuring safety and completeness through synthesis-based approaches.

Unlike LGP that solves an integrated TAMP problem, synthesis-based approaches using Linear Temporal Logic (LTL) operate primarily at the task level, emphasizing safety and completeness guarantees within the task domain \citep{kress2018synthesis}. These methods typically synthesize a sequence of task-level actions, which are then executed by a motion planner. Recently, LTL has been applied to safe locomotion tasks, employing distinct task-level abstractions on topological maps of terrain \citep{kulgod2020temporal,zhou2022reactive,jiang2023abstraction} or locomotion keyframes for reduced-order models \citep{gu2022reactive,shamsah2023integrated,zhao2022reactive}.
Reactive synthesis \citep{pnueli1989synthesis}, widely applied to mobile robots \citep{kress2009temporal}, has been employed to enable prompt decision-making in response to more complex environments \citep{shamsah2023integrated} or external perturbations \citep{gu2022reactive}. Notably, \citep{zhao2022reactive} formulate the terrain-adaptive locomotion problem as a sequential decision-making task, selecting appropriate locomotion modes with predefined contact and locomotion keyframes to accommodate dynamically changing terrains. However, the locomotion mode selection is explicitly encoded in the LTL specification using expert knowledge, lacking a physically feasibility guarantee.
In this work, we aim to combine the strengths of reactive synthesis and MICP, utilizing the global certificate of MICP as a feasibility checker for terrain-adaptive locomotion.


\subsection{Physically-Feasible Reactive Synthesis}
While reactive synthesis can promptly and safely respond to dynamic environments, it typically does not assess physical feasibility when being deployed on complex robotic systems. To bridge the gap between discrete abstraction and continuous system dynamics, various strategies have been proposed. Studies in  \citep{decastro2014dynamics,fainekos2006translating} focus on automatically synthesizing controllers in the continuous domain based on high-level specifications. Another approach involves incorporating dynamics directly into the reactive synthesis process by abstracting physical systems, including nonlinear \citep{bhatia2010sampling}, switched \citep{liu2013synthesis}, and hybrid systems \citep{maly2013iterative} with manageable model complexity. However, for legged robots with multiple contacts navigating dynamic terrains, these physically feasible reactive synthesis approaches remain computationally intractable.

Recent efforts \citep{pacheck2020finding, pacheck2022physically} focus on symbolic repair for unrealizable task specifications, defining symbolic skills with preconditions and postconditions, similar to TAMP, to identify missing skills that make the specification realizable. These works introduce iterative feasibility checks based on symbolic repair suggestions, emphasizing alignment between symbolic task specifications and physical reality. Building upon this, Meng et al. \citep{meng2024automated} extend this type of approach to online symbolic repair.
Inspired by these works, Zhou et al. \citep{zhou2025physically} adopt a similar mechanism for terrain-adaptive locomotion. Specifically, they abstract the robot and terrain states in a local environment and define physically feasible skills by solving MICP. 
They leverage high-level manager and symbolic repair to efficiently identify missing but necessary skills, reducing the need for frequent calls to the computationally expensive gait-free MICP. Our work significantly completes \citep{zhou2025physically} through seamless integration with a low-level tracking control module, enabling systematic benchmarking, and demonstrating real-world hardware deployment.





\section{Preliminaries}
Consider a robot equipped with sensors navigating an environment toward a designated global goal $g_{\textrm{global}} \in \mathbb{R}^2$. This task can be broken into a sequence of local navigation subtasks, where the robot moves toward intermediate local request goals $\mathcal{G}_{\textrm{local}}$, determined by a global planner based on the nearby segmented terrain polygons $\mathcal{P}$. We illustrate an instance of such a local task setup.

\begin{figure}[ht]
    \centering
    \includegraphics[width=\linewidth]{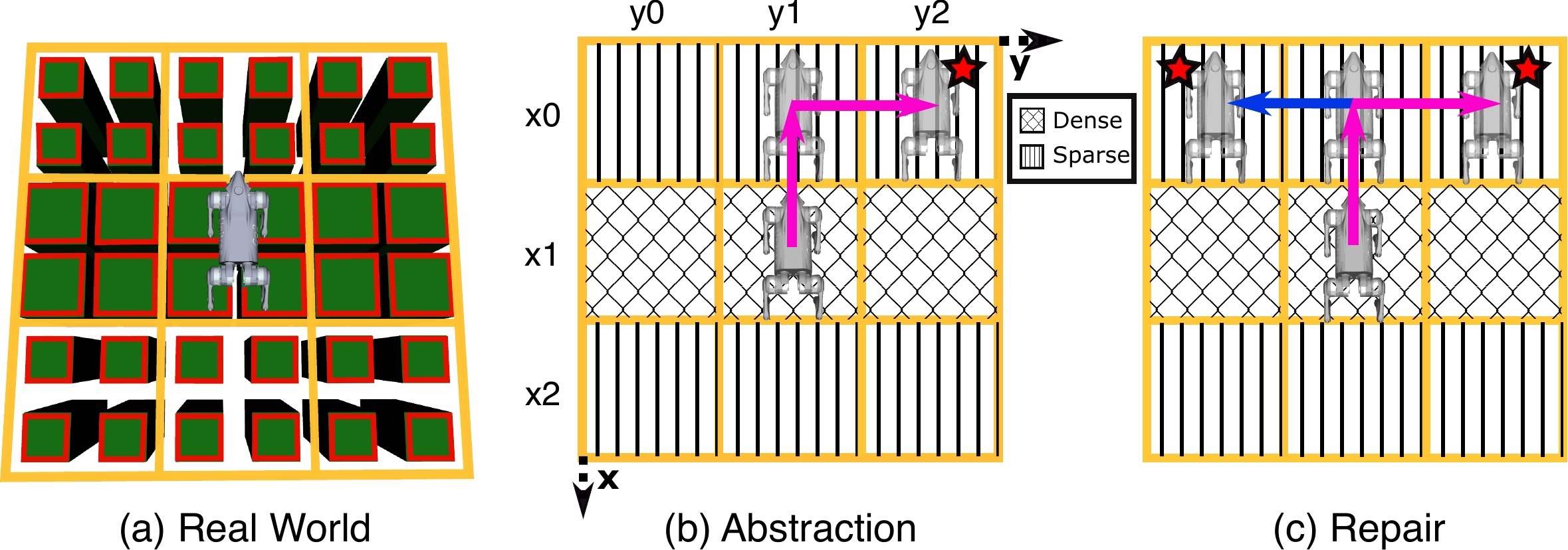}
    \caption{Terrain abstraction and skill definition. (a) Top-down view of the real-world terrain before abstraction. The red polygons denote the segmented terrain polygons. (b) Abstraction of the terrain and robot's skills (pink) moving from one location to another. (c) Repair process to find a new skill (blue).}
    \label{fig:preliminary_nine_squares}
\end{figure}

\noindent
\textbf{Example~\examplelabel{example:1}}.
\textit{
Consider a 2D grid world with nine cells (in yellow) in Fig.~\ref{fig:preliminary_nine_squares}. The robot is required to move from an initial cell (e.g. the center cell) 
to a desired goal cell indicated by the star. 
Each cell is assigned a terrain type, such as $\textit{Dense}$ or $\textit{Sparse}$ stepping stones.}

\subsection{Abstractions}\label{sec:preliminaries_abstraction}

Given a set of terrain polygons $\mathcal{P}$, such as those in Fig.~\ref{fig:preliminary_nine_squares}(a), we abstract them into a 2D grid of dimension $n \times m$. For each axis, we label every position with a symbol, producing two sets: $\xcells \coloneqq \{x_0, \dots, x_{n-1}\}$ and $\ycells \coloneqq \{y_0, \dots, y_{m-1}\}$.

We introduce a collection of atomic propositions $\ap$, partitioned into input variables $\mathcal{I}$ and output variables $\mathcal{O}$, which capture the symbolic world state and robot actions. The inputs $\mathcal{I}$ include robot position inputs $\inpcen$, request inputs $\inprequest$, and terrain inputs $\inpterrain$, with $\inp = \inpcen \cup \inprequest \cup \inpterrain$.
The robot inputs are defined as $\inpcen \coloneqq \{\var_x \mid x \in \xcells\} \cup \{\var_y \mid y \in \ycells\}$ to represent the robot’s location. Request inputs are given by $\inprequest \coloneqq \{\var_x^\textrm{req} \mid x \in \xcells\} \cup \{\var_y^\textrm{req} \mid y \in \ycells\}$ to represent the desired goal cell. The terrain inputs are defined as $\inpterrain \coloneqq \{\tvar{x}{y} \mid x \in \xcells, y \in \ycells\}$, which describe the terrain type associated with each grid cell.
We assume a finite set of terrain types of size $n_t$, defined according to their physical characteristics. For $\physicalstate \in \naturals$, we denote $\range{\physicalstate} \coloneqq \{0, \dots, \physicalstate - 1\}$. Accordingly, each terrain input $\tvar{x}{y} \in \inpterrain$ is an integer variable taking values from $\range{n_t}$, which we further translate into Boolean propositions following the approach of~\citep{ehlers2016slugs}. The robot state abstraction can also be extended to incorporate orientation, such as the yaw angle of the floating base (see Sec.~\ref{sec:extension}).

An input state $\inpstate$ is a mapping $\inpstate: \inp \to \{\true, \false\} \cup \range{n_t}$. Because the robot and its goal occupy only a single grid cell each, exactly one $\var_x$, one $\var_y$, one $\var_x^\textrm{req}$, and one $\var_y^\textrm{req}$ must evaluate to $\true$ for any valid input state. We denote projections of $\inpstate$ as robot states $\inpstatecen: \inpcen \to \{\true, \false\}$, request states $\inpstaterequest: \inprequest \to \{\true, \false\}$, and terrain states $\inpstateterrain: \inpterrain \to \range{n_t}$. The complete sets of such states are denoted by $\inpstates$, $\inpstatescen$, $\inpstatesrequest$, and $\inpstatesterrain$, respectively.

We introduce a grounding function to associate symbolic input states with their physical interpretations. Let the physical state space be $\ps \subseteq \reals^n$, which encodes the physical world, including the robot’s position, orientation, and terrain attributes. The grounding function $\grounding: \inpstates \to 2^\ps$ maps each input state $\inpstate$ to the corresponding set of physical states in $\ps$.

For robot states $\inpstatecen \in \inpstatescen$, we define 
$\grounding(\inpstatecen) \coloneqq \{
\physicalstate \in \ps \mid
\exists \var_x, \var_y \in \inpcen. 
\inpstatecen(\var_x) \land \inpstatecen(\var_y) \land
\textrm{robot\_pose}(\physicalstate) \in \textrm{cell}(x,y)
\}$.
Similarly, for request states $\inpstaterequest \in \inpstatesrequest$, we set
$\grounding(\inpstaterequest) \coloneqq \{
\physicalstate \in \ps \mid
\exists \var_x^\textrm{req}, \var_y^\textrm{req} \in \inprequest. 
\inpstaterequest(\var_x^\textrm{req}) \land \inpstaterequest(\var_y^\textrm{req}) \land
\textrm{request\_goal}(\physicalstate) \in \textrm{cell}(x,y)
\}$. 
For terrain states $\inpstateterrain \in \inpstatesterrain$, the grounding $\grounding(\inpstateterrain)$ is defined as the set of physical states whose terrain abstractions are consistent with $\inpstateterrain$.
Given an input state $\inpstate \in \inpstates$ with projections $\inpstatecen$, $\inpstaterequest$, and $\inpstateterrain$, we combine them as
$\grounding(\inpstate) \coloneqq \grounding(\inpstatecen) \cap \grounding(\inpstaterequest) \cap \grounding(\inpstateterrain)$.
Finally, we introduce an inverse grounding function $\inversegrounding: \ps \to \inpstates$ that maps each physical state $\physicalstate \in \ps$ back to its corresponding symbolic input state.


In Example~\ref{example:1}, the inputs are defined as
$\inpcen \coloneqq
\{
\var_{x_0}, \var_{x_1}, \var_{x_2},
\var_{y_0}, \var_{y_1}, \var_{y_2}
\}$,
$\inprequest \coloneqq \{\var^\textrm{req} \mid \var \in \inpcen\}$,
and
$\inpterrain \coloneqq \{\tvar{x_i}{y_j} \mid i,j\in {0,1,2}\}$,
where $\tvar{x_i}{y_j} = 1$ if the grid cell $(x_i, y_j)$ is classified as Dense, and $\tvar{x_i}{y_j} = 0$ if it is Sparse.
The input state shown in Fig.~\ref{fig:preliminary_nine_squares}(b) is
$\inpstate:
\var_{x_1} \mapsto \true,
\var_{y_1} \mapsto \true,
\var_{x_0}^\textrm{req} \mapsto \true,
\var_{y_2}^\textrm{req} \mapsto \true,
\tvar{x_1}{y_j} \mapsto 1 \ \text{for } j\in \{0,1,2\}$.
For brevity, throughout this paper we omit Boolean propositions set to $\false$ and integer propositions with value $0$ when describing input states.


The output propositions
$\out$
represent skills that enable the robot to move between grid cells under specific terrain conditions.
A skill $\skill \in \out$ is defined by its preconditions
$\skillpre \subseteq \inpstatescen \times \inpstatesterrain$,
which specify the robot and terrain states from which it can be executed,
and its postconditions
$\skillpost \subseteq \inpstatescen$,
which describe the resulting robot state after execution.
In Example~\ref{example:1}, the skill $\skillcustom{0}$ moves the robot from the middle cell to the upper cell (Fig.~\ref{fig:preliminary_nine_squares}b).
The precondition of $\skillcustom{0}$ is
$\skillprecustom{\skillcustom{0}} = \{ (\inpstatecen, \inpstateterrain): \var_{x_1}\mapsto \true, \var_{y_1}\mapsto \true, \tvar{x_1}{y_0}\mapsto 1, \tvar{x_1}{y_1}\mapsto 1, \tvar{x_1}{y_2}\mapsto 1 \}$,
while the postcondition is
$\skillpostcustom{\skillcustom{0}} = \{ \inpstatecen: \var_{x_0}\mapsto \true, \var_{y_1}\mapsto \true \}$.
Each symbolic skill is also paired with a continuous-level locomotion gait, such as a one-second trotting gait $L$, which physically realizes the symbolic transition (Sec.~\ref{sec:motion_primitive}).


\subsection{Specifications}

We adopt the Generalized Reactivity (1) (GR(1)) fragment of linear temporal logic (LTL)~\citep{bloem2012synthesis} to encode task specifications.
An LTL formula $\spec$ follows the grammar
$\spec \: := \var \:|\: \neg\: \spec \:|\: \spec \:\land\: \spec \:|\: \bigcirc\spec \:|\: \square \spec \:|\: \lozenge \spec$,
where $\var \in \ap$ denotes an atomic proposition,
$\neg$ and $\land$ are Boolean operators,
and $\bigcirc$ (``next''), $\square$ (``always''), and $\lozenge$ (``eventually'') are temporal operators.
For a comprehensive introduction to LTL, we refer the reader to~\citep{baier2008principles}.

A GR(1) specification is written in the form $\spec = \envspec \to \sysspec$,
where $\envspec = \envinit \land \envsafety \land \envlive$ encodes the assumptions about the (possibly adversarial) environment,
and $\sysspec = \sysinit \land \syssafety \land \syslive$ represents the guarantees that define the system’s behavior.
For $\alpha \in \{e, s\}$, the components $\alphainit, \alphasafety, \alphalive$ describe the initial conditions, safety requirements, and liveness objectives, respectively.
\revised{We refer below and throughout the rest of this article to \emph{symbolic repair}, the process of augmenting the symbolic transition set with new, dynamically feasible skills whenever the current specification becomes unrealizable. The full procedure is formalized in Sec.~\ref{sec:manager_and_repair}; for the purpose of the upcoming definitions, it suffices to treat symbolic repair as a black box that proposes candidate symbolic transitions whose physical feasibility is then verified by MICP.}
In the context of symbolic repair (Sec.~\ref{sec:manager_and_repair}), we further partition the safety constraints ($\envsafety, \syssafety$) into two parts: skill-related constraints ($\envsafetynothard, \syssafetynothard$) and hard constraints ($\envsafetyhard, \syssafetyhard$).
The skill constraints encode the preconditions and postconditions of skills (see Sec.~\ref{sec:spec_gen}) and are subject to modification by the repair procedure, whereas the hard constraints remain immutable.



In practice, GR(1) specifications often become quite large because they must capture detailed information about both terrain and task constraints. At runtime, however, the exact terrain and request states are already available. This allows us to apply a technique known as partial evaluation originally defined in \citep{zhou2025physically}, where known Boolean propositions are directly substituted with their truth values. By doing so, we reduce the overall state space, leading to a smaller and more computationally efficient specification for synthesis.

\noindent\textbf{Definition~\deflabel{def:partial_eval}.}
Given an LTL formula $\spec$
and
two subsets $S^\true, S^\false \subseteq \ap$, $S^\true \cap S^\false = \emptyset$,
we define
the \textit{partial evaluation} of $\spec$ over $S^\true, S^\false$,
as $\spec[S^\true, S^\false]$, 
where we substitute 
propositions $\pi\in AP$ in $\spec$ with $\true$ if $\pi \in S^\true$ and with $\false$ if $\pi \in S^\false$.

\section{Problem Statement and Approach Overview}\label{sec:problem_statement}

\subsection{Problem Statement}
\textbf{Problem~\problemlabel{problem:1}}.
Given
\begin{enumerate*}[label=(\roman*),ref=\roman*]
    \item a global goal $g_{\textrm{global}}$,
    \item a set of possible local request goals $\mathcal{G}_{\textrm{local}}$
    \item a set of predefined terrain polygons $\mathcal{P}_{\text{pre}}$ from user's prior knowledge,
    and
    \item a set of online terrain polygons $\mathcal{P}_{\text{on}}$ perceived during execution that may differ from the predefined ones,
    \item a set of predefined locomotion gaits $\mathcal{L}$;
\end{enumerate*}
generate controls that enable the robot to navigate the environment and reach the goal.

\subsection{Approach Overview}\label{sec:approach_overview}
To address Problem~\ref{problem:1} efficiently, we manage complexity on both symbolic and physical levels. At the symbolic level, we employ reactive synthesis to break down the local navigation task into smaller subproblems, which can be reused across different scenarios. Each subproblem involves planning a short-horizon symbolic transition and is accompanied by solving a MICP to certify its physical feasibility.

As shown in Fig.~\ref{fig:system_diagram}, our overall framework is composed of three main modules: offline synthesis (Sec.~\ref{sec:reactive_gait_synthesis}), online execution (Sec.~\ref{sec:online_execution}), and low-level tracking control (Sec.~\ref{sec:tracking_control}). During offline synthesis, we generate a diverse set of locomotion skills and their associated gaits for various predefined terrain-task combinations. Symbolic repair is applied to discover additional necessary transitions that may not be captured initially.

At runtime, we invoke a symbolic repair mechanism to synthesize new transitions when the robot encounters unexpected terrain conditions or request goals not seen during offline planning. This allows the framework to remain flexible and adaptive to evolving environments. \revised{Additionally, the system re-targets the desired pose during each transition to reflect current terrain constraints, and coordinates the strategy automaton with the tracking controller to handle MICP solve-time variability without compromising execution continuity.}

\section{Offline Synthesis}\label{sec:reactive_gait_synthesis}
\begin{figure*}[ht]
    \centering
    \includegraphics[width=\linewidth]{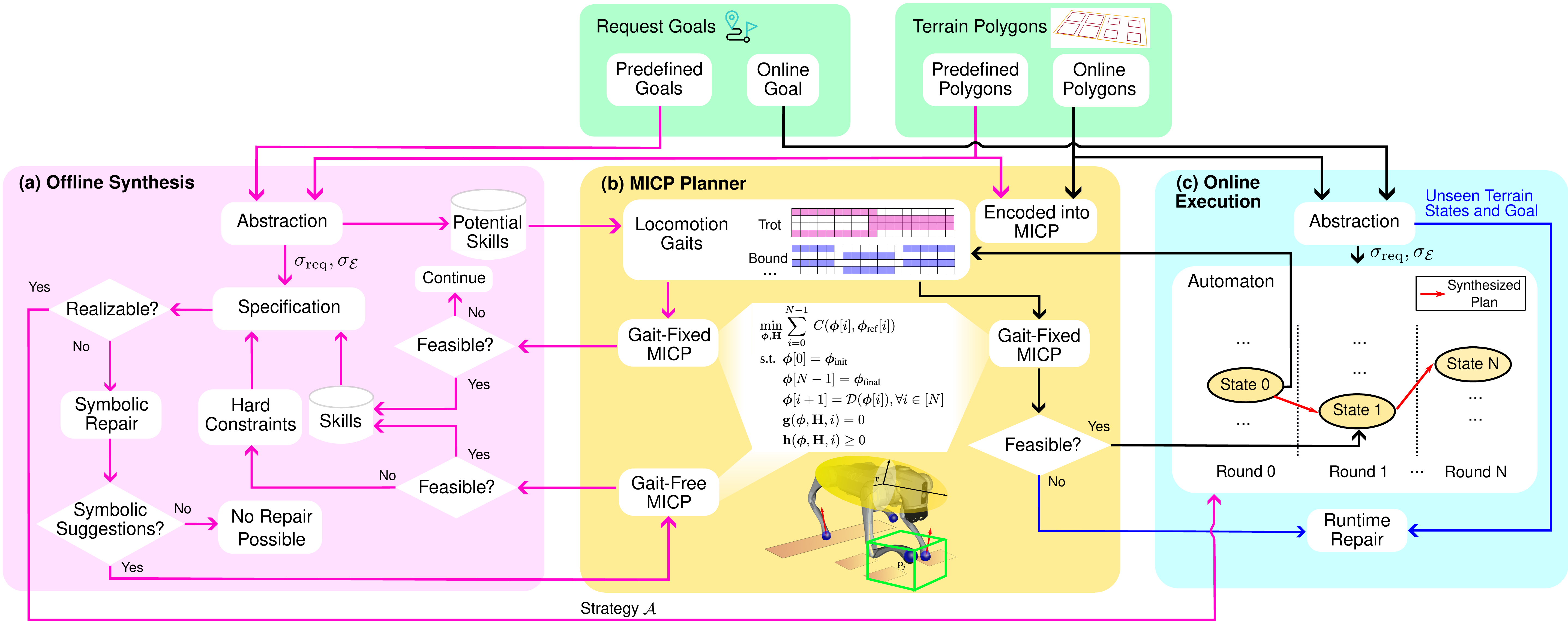}
    \caption{System overview.  
    During offline synthesis (pink arrows), an initial set of locomotion gaits is provided, and symbolic skills are iteratively generated by solving the MICP. When the task specifications are unrealizable, a symbolic repair is triggered to seek missing skills.
    During the online execution (black arrows), MICP is solved again taking online terrain segments and the symbolic state only advances when a solution is found.
    A runtime repair (blue arrows) is initiated if a solving failure occurs, 
    or an unseen terrain or request goal is encountered.
    }
    \label{fig:combined_offline_online}
    \vspace{-0.1in}
\end{figure*}

The offline synthesis module builds a strategy that allows the robot to reach local request goals under predefined terrain conditions. As shown in Fig.~\ref{fig:combined_offline_online}(a), the module takes in local request goals $\mathcal{G}_{\textrm{local}}$, terrain polygons $\mathcal{P}_{\text{pre}}$ from prior maps, and a set of locomotion gaits $\mathcal{L}$ specified by the user.
We first apply the inverse grounding function $\inversegrounding$ to discretize the local environment, deriving a set of candidate request states $\inpstatesrequestposs \subseteq \inpstatesrequest$ from the local request goals and mapping the predefined terrain polygons into possible terrain states $\inpstatesterrainexprposs \subseteq \inpstatesterrain$.
Afterward, we evaluate the feasibility of each candidate skill using a gait-fixed MICP formulated with the provided locomotion gaits (Sec.~\ref{sec:motion_primitive}), and record all feasible transitions as symbolic skills in the specification (Sec.~\ref{sec:spec_gen}).
To improve efficiency, a high-level manager then generates partial evaluations of the encoded specifications. Whenever one of these partial evaluations is deemed unrealizable, a repair mechanism is invoked: it proposes additional symbolic skills, whose feasibility is validated through gait-free MICP. This process ensures that the specification ultimately becomes realizable (Sec.~\ref{sec:manager_and_repair}).
\subsection{Locomotion Gait and Feasibility Checking via MICP}\label{sec:motion_primitive} 
We assume each locomotion gait $L$ is characterized by a contact sequence $\mathcal{G}$ and its associated time durations $T$, formally written as $L = \mathcal{M}(\mathcal{G}, T)$. Each skill is tied to a distinct locomotion gait, which dictates how the robot executes motion at the continuous level.
As illustrated in Fig.~\ref{fig:combined_offline_online}(a), since the offline phase has access to the complete set of request states $\inpstatesrequestposs$ and terrain states $\inpstatesterrainexprposs$, we can systematically enumerate all possible symbolic skills that enable navigation across the different local environments. The feasibility of each skill is then evaluated through gait-fixed MICP using the given locomotion gaits.
In most examples shown in this paper, we assume the robot is only allowed to move horizontally and vertically by one cell and we show diagonal movements and heading angle change in Sec.~\ref{sec:extension}.

\begin{figure}
    \centering
    \includegraphics[width=0.9\linewidth]{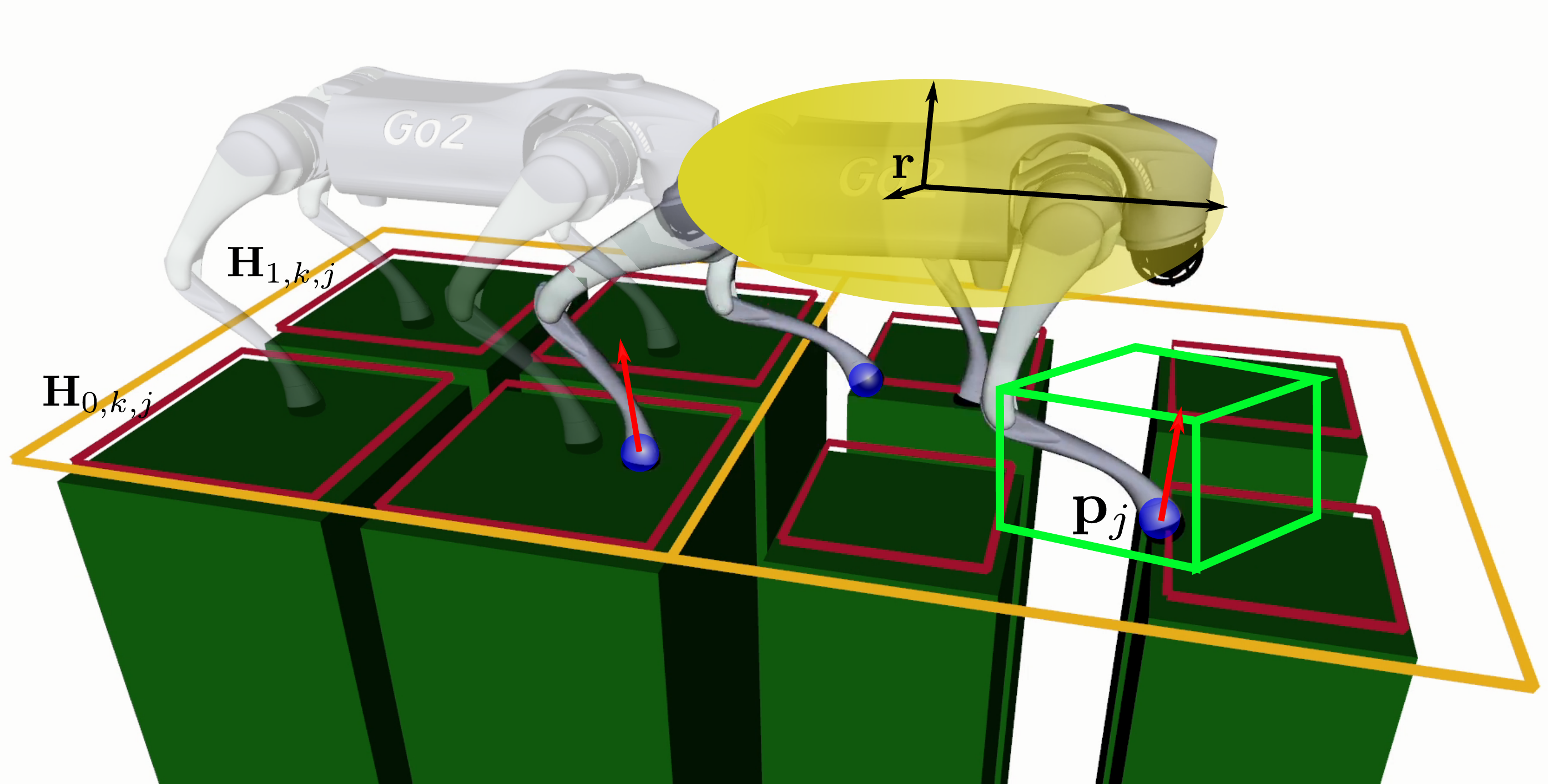}
    \caption{Demonstration of decision variables and polygons when solving MICP for skill $a_0$ in Example~\ref{example:1} using a trotting locomotion gait.}
    \label{fig:mip_go2}
\end{figure}

Then the next critical step is to find whether there exists a locomotion gait for each skill to be physically feasible.
We leverage mixed-integer convex programming (MICP) to check the physical feasibility given a set of predefined locomotion gaits, and only use the feasible one as robot skills to synthesize robot strategies. The MICP problem takes in
the centroidal states $\inpcen$ from the precondition and postcondition of the skill as its initial and final conditions, which are located at the center of each abstracted cell. A set of homogeneous, predefined terrain polygons are considered in the MICP as steppable regions corresponding to the terrain states $\inpterrain$ involved in the precondition. Lastly, the contact information $\mathcal{G}$ and $T$ is provided by the selected locomotion gait $L$. Fig.~\ref{fig:mip_go2} shows the maneuver of the quadruped robot Go2 after solving the MICP problem corresponding to skill $a_0$ in Example~\ref{example:1} with a one-second trotting locomotion gait. The generic MICP problem considered in this work can formulated as:
    \begin{subequations}\label{ocp:mip_general}
    \begin{alignat}{2}
        \nonumber &\underset{\vg \phi, \v H}{\min} && \sum_{i = 0}^{N-1} \hspace{2pt} C(\vg \phi[i], \vg \phi_{\text{ref}}[i])\\
        & \text{s.t. } && \vg \phi[0] = \vg \phi_{\text{init}} \\& && \vg \phi[N-1] = \vg \phi_{\text{final}} \\
     & &&  \vg \phi[i+1] = \mathcal{D}(\vg \phi[i]), \forall i \in \range{N} \\
        & && \v g(\vg \phi, \v H, i) = 0 \\
        & && \v h(\vg \phi, \v H, i) \geq 0
    \end{alignat}
\end{subequations}
where the decision variables $\vg \phi$ denote the continuous variables and $\v H$ indicate the binary variables across the entire $N$ timesteps. For any natural number $n \in \naturals$, we denote the set $\{0, \dots, n-1\}$ as $\range{n}$. \revised{The function $\mathcal{D}(\cdot)$ encodes the discrete-time system dynamics that propagate the continuous state $\vg \phi[i]$ to $\vg \phi[i+1]$; the specific dynamics used in this work are detailed in Eq.~(\ref{eq:dynamics}).} The general equality constraints $\v g(\cdot)$ and inequality constraints $\v h(\cdot)$ are incorporated and activated at certain time stamp $i$. The cost function depends on both the continuous variables $\vg \phi$ and a reference trajectory $\vg \phi_{\text{ref}}$.

The reference base position and angular trajectories are generated by linearly interpolating between the initial and final conditions. For scenarios with significant elevation changes, such as jumping onto higher terrain, an additional middle keyframe is introduced for the pitch angle to calculate the slope angle between the initial and final poses. The final reference pitch trajectory is then created by evenly interpolating between the initial, middle, and final poses. The reference EE trajectory relative to the base frame $^\mathcal{B} \v p_j^{\text{ref}}$ remains constant and moves in sync with the reference base trajectory. Given a specific locomotion gait, we first introduce the gait-fixed MICP formulation:

\subsubsection{Continuous Decision Variables}
The continuous decision variables $\vg \phi$ include base position $\v r$, velocity $\dot{\v r}$, acceleration $\ddot{\v r}$, orientation $\vg \theta$ parameterized by Euler angle and its first and second-order derivatives $\dot{\vg \theta},\ddot{\vg \theta}$, individual end-effector (EE) position $\v p_j$, velocity $\dot{\v p}_j$, and acceleration $\ddot{\v p}_j$, and individual contact force $\v f_j$ for the foot $j$ with $j\in\range{n_f}$, where $n_f=4$ denotes the index of feet. The compact form can be expressed as:
\begin{equation}
    \vg \phi = [\v r^\top, \dot{\v r}^\top, \ddot{\v r}^\top, \vg \theta^\top, \dot{\vg \theta}^\top, \ddot{\vg \theta}^\top, \v p_j^\top, \dot{\v p}_j^\top, \ddot{\v p}_j^\top, \v f_j^\top]^\top
\end{equation}

\subsubsection{System Dynamics}
We encode the system dynamics as a simplified single rigid body in Eq.~(\ref{eq:dynamics}),
using a double integrator to replace the original Euler equation to keep the dynamics constraint convex. The whole system is discretized through Backward Euler integration with $\Delta t$ between each time step.
\begin{equation}
        \label{eq:dynamics}
        \begin{bmatrix} \v r[i+1]\\\
        \dot{\v r}[i+1]\\\
        m \ddot{\v r}[i]\\\
        \vg \theta[i+1]\\
        \dot{\vg \theta}[i+1]
        \end{bmatrix}= \begin{bmatrix}
        \v r[i] + \Delta t \cdot \dot{\v r}[i+1]\\
        \dot {\v r}[i] + \Delta t \cdot \ddot{\v r}[i+1]\\
        \sum\nolimits_{j} \v f_j[i] + m \v g \\
            \vg \theta[i] + \Delta t \cdot \dot{\vg \theta}[i+1]\\
            \dot{\vg \theta}[i] + \Delta t \cdot \ddot{\vg \theta}[i+1]
	    \end{bmatrix}
\end{equation}
where $m$ is the robot mass and $g$ is the gravity term.

\revised{We note that the rotational dynamics adopted in Eq.~(\ref{eq:dynamics}) are intentionally simplified: the rotational equation of motion is replaced by a double integrator on the Euler angles, which keeps the constraint convex but \textit{neglects the true angular-momentum dynamics.} This makes the formulation tractable for online deployment, but it limits applicability to highly dynamic motions such as flips or aggressive yaw maneuvers at high speeds.
Future work will explore convex relaxations of the full angular-momentum dynamics to expand the range of feasible motions while maintaining online tractability.}

\subsubsection{Cost Function}
The cost function consists of tracking costs and regularization terms governed by diagonal matrices $\v Q$ and $\v R$. 
\begin{equation}
    \v C = \delta \vg \phi_Q[i]^T\mathbf{Q}\hspace{2pt}\delta \vg \phi_Q[i] + \vg \phi_R[i]^{T}\mathbf{R}\vg \phi_R[i]
\end{equation}
where $\delta \vg \phi_Q$ and $\vg \phi_R$ are defined as:
\begin{equation}
    \delta \vg \phi_Q = \begin{bmatrix}
        \v r - \v r_{\text{ref}}\\
        \vg \theta - \vg \theta_{\text{ref}}\\
        \v p_j - \v p_{j}^{\text{ref}}
    \end{bmatrix}, \vg \phi_R = \begin{bmatrix}
        \ddot{\v r}\\
        \ddot{\vg \theta}\\
        \ddot{\v p}_j\\
        \v f_j
    \end{bmatrix}
\end{equation}

Tracking costs include deviation from the desired base and foot EE trajectories. The regularization term includes minimizing the base acceleration, Euler angle acceleration, EE acceleration, and contact forces to encourage motion smoothness. The weights for the tracking terms are defined in Table~\ref{tab:cost_weights}.



\subsubsection{Safe Region Constraint}
To incorporate safe region constraints for selecting proper footholds, we introduce binary variables $\v H_{r,k,j}$, with $r$, $k$, and $j$ expressing the $r^{\rm th}$ convex region, $k^{\rm th}$ footstep, and $j^{\rm th}$ foot and $r\in\range{R}, k\in\range{n_s}$. One footstep is defined as a full swing phase for a foot. We use $n_s$ and $R$ to represent the number of footsteps specified by the gait configuration and the number of convex terrain polygons to be considered. For example,  Fig.~\ref{fig:mip_go2} shows a case with eight polygons.
Eqs.~(\ref{eq:safe_region}) - (\ref{eq:safe_region_eq}) restrict
the robot's EE to stay within one of the convex polygons when the corresponding binary variable $\v H_{r,k,j}$ is True. 
Each polygon is parameterized by an inequality constraint ($\v A_r$ and $\v b_r$) that defines multiple half-spaces, along with an equality constraint ($\v A_{\text{eq},r}$ and $\v b_{\text{eq},r}$) that ensures the foothold position lies on a 3D plane.
We use $\mathcal{C}_{k,j}$ to denote the set of time steps indicating stance after the $k^{\rm th}$ footstep for the $j^{\rm th}$ foot and the safe region constraint only applies to the first stance time step represented as $\mathcal{C}_{k,j}[0]$. Note that, during the offline phase, the terrain polygons are homogeneous and predefined.
\begin{align}
\nonumber
&\forall r \in [R], k \in [n_s], j \in [n_f]\\
    \label{eq:safe_region} &\v H_{r,k,j} \Rightarrow \v A_r \v p_j[i] \leq \v b_r, \forall i \in \mathcal{C}_{k,j}[0]
        \\\label{eq:safe_region_eq} & \hspace{1.5cm} \v A_{\text{eq},r} \v p_j[i] = \v b_{\text{eq},r}, \forall i \in \mathcal{C}_{k,j}[0]
        \\\label{eq:sum_contact_binary} &\sum_{r=0}^{R-1}\v H_{r,k,j} = 1
        \\&\v H_{r,k,j} \in \{0, 1\}
\end{align}

\subsubsection{Frictional and Contact Constraints}
Frictional constraints are defined in Eqs.~(\ref{eq:force_non_zero}) - (\ref{eq:force_friction_cone}), with $\v n_r$ and $\mathcal{F}_r$ as the normal vector and friction cone of the $r^{\rm th}$ convex region corresponding to the x and y dimensions of the EE position $\v p^{xy}_j$. Contact constraints in Eqs.~(\ref{eq:ee_velocity_zero}) - (\ref{eq:force_zero}) forces the EE velocity to be zero during stance phase and the contact force to be zero during the non-contact phase. $\mathcal{C}_{j}$ represents the set of all time steps indicating stance for the $j^{\rm th}$ foot.
\begin{align}
    \nonumber &\forall r \in [R], k \in [n_s], j \in [n_f]\\\label{eq:force_non_zero}
    &\v H_{r,k,j} \ \  \Rightarrow \ \ \v f_j[i] \cdot \v n_r(\v p_j^{xy}[i]) \ge 0, \ \forall i \in \mathcal{C}_{k,j}\\
    \label{eq:force_friction_cone}
        & \hspace{1.95cm} \v f_j[i] \in \mathcal{F}_r(\mu, \v n_r, \v p_j^{xy}[i]), \ \forall i \in \mathcal{C}_{k,j}
         \\\label{eq:ee_velocity_zero}
            &\dot{\v p}_j[i]=0, \ \forall i \in \mathcal{C}_j
        \\\label{eq:force_zero}
         &\v f_j[i] = 0, \ \forall i \notin \mathcal{C}_j
\end{align}
where $\mu$ denotes the friction coefficient.

\subsubsection{Actuation Constraint}
Since the joint angles are omitted in the single rigid body model, we use a fixed Jacobian $\v J_j(\v q_j^{\text{ref}})$ at a nominal joint pose $\v q_j^{\text{ref}}$ for each leg to approximately consider the torque limit constraint in Eq.~(\ref{eq:max_torque}). In addition, since the translational and angular motions are decoupled, another actuation constraint on the angular acceleration is also added in Eq.~(\ref{eq:max_acc}).
\begin{align}
\nonumber & \forall i \in [N], j \in [n_f] \\
    \label{eq:max_torque}
& \v J_j(\v q_j^{\text{ref}})^{\top} \v f_j[i] \leq \vg \tau_{\rm max}\\\label{eq:max_acc}
& \v I \ddot{\vg \theta}[i] \leq \vg \tau_{\rm max}^{\prime}
\end{align}
where $\vg \tau_{\rm max}$ is the joint torque limit, $\vg \tau_{\rm max}^{\prime}$ is the torque limit applied on the base, and $\v I$ is the moment of inertia for the approximated single rigid body.

\subsubsection{Kinematics Constraint}
Lastly, the kinematics constraint in Eq.~(\ref{eq:rom}) strictly limits the possible EE movements to assure safety.
Due to the convex nature of this MICP, we can determine the feasibility of a possible symbolic transition by checking if the optimal solution exists.
\begin{equation}
    \label{eq:rom}
    \v p_j[i] \in \mathcal{R}_j(^\mathcal{B} \v p_j^{\text{ref}}, \v r[i],\vg \theta_{\text{ref}}, \v p_j^{\text{max}}), \forall i \in [N], j \in [n_f]
\end{equation}
where $\mathcal{R}_j$ is defined as a 3D box constraint around a nominal foot EE position based on the base position and orientation, and constrained by a maximum deviation $\v p_j^{\text{max}}$. Note that the reference orientation $\vg \theta_{\text{ref}}$ is used to avoid nonlinear constraint keep the problem convex.

\begin{table}[t]
\centering
    \caption{Tracking Cost Weights}
    \setlength{\tabcolsep}{4pt}
     \begin{tabular}{ll}
         \hline
         \textbf{Cost Term} & \textbf{Weights}\\\hline
         $\v r - \v r_{\text{ref}}$ & (1000.0, 1000.0, 1000.0)\\
         $\vg \theta - \vg \theta_{\text{ref}}$ & (1000.0, 1000.0, 1000.0)\\
         $\v p_j - \v p_{j}^{\text{ref}}$ & (1000.0, 1000.0, 1000.0)\\
         $\ddot{\v r}$ & (10.0, 10.0, 10.0)\\
         $\ddot{\vg \theta}$ & (10.0, 10.0, 10.0)\\
         $\ddot{\v p}_j$ & (0.5, 0.5, 0.5)\\
         $\v f_j$ & (0.1, 0.1, 0.1)\\
         \hline
    \end{tabular}
    \label{tab:cost_weights}
\end{table}
\subsection{Task Specification Encoding}\label{sec:spec_gen}



After identifying a set of feasible robot skills, we encode them into the specification $\spec$ as part of the environment safety assumptions $\envsafety$ and the system safety guarantees $\syssafety$. \revised{Each skill is associated with a user-defined precondition set $\skillpre \subseteq 2^{\inpcen \cup \inpterrain}$ and postcondition set $\skillpost \subseteq 2^{\inpcen}$, expressed over the robot and terrain inputs introduced in Sec.~\ref{sec:preliminaries_abstraction}. The encoding rules below map these sets into the GR(1) components $\envsafetynothard$ and $\syssafetynothard$ deterministically.}

The environment-side skill assumptions $\envsafetynothard$ capture the postconditions of each skill. These assumptions enforce that whenever a skill’s precondition holds and the skill is executed, at least one of its associated postconditions must also hold:
\begin{equation}
\begin{aligned}
    \envsafetynothard \coloneqq \bigwedge_{\skill \in \mathcal{O}} 
\square
((\skill 
    \land
    \bigvee_{\statevar\in\skillprecustom{\skill}}
    \bigwedge_{\var\in \inpcen \cup \inpterrain} \var=\statevar(\var))\\
    \to
    \bigvee_{\statevar\in\skillpost} 
    \bigwedge_{\var \in \inpcen}
    {\bigcirc (\var=\statevar(\var))} 
    )
\end{aligned}
\end{equation}
%
In Example~\ref{example:1},
the postcondition of skill $\skillcustom{0}$
is defined as
\begin{equation}
\begin{aligned}
    \square (\skillcustom{0} 
     \land \var_{x_1} \land \var_{y_1} \land \tvar{x_1}{y_0} = 1 \land \tvar{x_1}{y_1} = 1 \land \tvar{x_1}{y_2} = 1 \land \dots \\
    \to \bigcirc \var_{x_0} \land \bigcirc \var_{y_1} \land \neg \dots)
\end{aligned}
\end{equation}

The system-side skill guarantees $\syssafetynothard$ encode the preconditions associated with each skill. These guarantees restrict when a skill may be executed by the robot. Specifically, a skill is permitted to run only if at least one of its preconditions is satisfied:
\begin{equation}
\begin{aligned}
    \syssafetynothard \coloneqq 
    \bigwedge_{\skill \in \mathcal{O}} 
    \square (
    \neg (
    \bigvee_{\statevar\in\skillprecustom{\skill}}
    \bigwedge_{\var\in \inpcen \cup \inpterrain}
    {\bigcirc(\var = \statevar(\var))}
    )
    \\
    \to \neg \bigcirc \skill)
\end{aligned}
\end{equation}
\revised{The disjunction $\bigvee_{\statevar \in \skillpre}$ in the equation above is taken over precondition pairs in $\skillpre$: a skill becomes eligible to execute when the conjunction of input assignments specified by at least one full pair $(\inpstatecen, \inpstateterrain) \in \skillpre$ holds, rather than when only some individual atomic propositions within a pair hold.}
%
In Example~\ref{example:1}, the precondition of skill $\skillcustom{0}$ is defined as
\begin{equation}
\begin{aligned}
\square \big( \neg 
     (
     \bigcirc \var_{x_1} \land 
     \bigcirc \var_{y_1} \land
     \bigcirc \tvar{x_1}{y_0}=1 \land 
     \bigcirc \tvar{x_1}{y_1}=1 \land 
     \\
     \bigcirc \tvar{x_1}{y_2}=1 \land 
     \dots)
     \to \neg \bigcirc \skillcustom{0} \big)
\end{aligned}
\end{equation}

The hard constraints $\envsafetyhard$ and $\syssafetyhard$ are constraints that a repair cannot modify 
(see Sec.~\ref{sec:manager_and_repair}).
Our system's hard constraints
$\syssafetyhard$
only allow the robot to execute one skill at a time:
    $\square\big(\neg (\bigcirc\skill \land \bigcirc\skill')\big)$, for any two different skills $\skill, \skill' \in \out$. 
Given that, 
we encode the following hard assumptions $\envsafetyhard$:
\begin{enumerate}[label=(\roman*),ref=\roman*]
    \item the uncontrollable terrain and request inputs 
    cannot change during execution:
    $\square (\var \leftrightarrow \bigcirc \var)$,
    $\forall \var \in \inpterrain \cup \inprequest$,
    and
    \item the robot inputs $\inpcen$ remain unchanged if no skill is executed: 
    $\square\big(\bigwedge_{\skill\in\out} \neg \skill \to \bigwedge_{\var\in\inpcen} (\var \leftrightarrow \bigcirc \var)\big)$.
\end{enumerate}

\subsection{High-level Manager and Symbolic Repair}\label{sec:manager_and_repair}
We design a high-level manager to efficiently synthesize controllers for encoded specifications under given sets of terrain and request states. The manager takes as input the specification $\spec$, a predefined collection of terrain states $\inpstatesterrainexprposs \subseteq \inpstatesterrain$, and a predefined set of request states $\inpstatesrequestposs \subseteq \inpstatesrequest$. For each pair $(\inpstateterrain, \inpstaterequest) \in \inpstatesterrainexprposs \times \inpstatesrequestposs$, the manager first converts the integer-valued terrain state $\inpstateterrain: \inpterrain \to \range{n_t}$ into a Boolean representation $\inpstateterrainbool: \inpterrainbool \to \{\true, \false\}$, following the approach in \citep{ehlers2016slugs}. We then overload $\inpstateterrainbool$ and $\inpstaterequest$ to denote the propositions that evaluate to $\true$. Using this, the manager generates a partial evaluation $\spec' \coloneqq \spec[\inpstateterrainbool \cup \inpstaterequest, \inpterrainbool \cup \inprequest \setminus \inpstateterrainbool \cup \inpstaterequest]$ (see Definition~\ref{def:partial_eval}). Skills whose preconditions are evaluated to $\false$ under the partial evaluation are discarded, which reduces the specification to robot inputs $\inpcen$ and a smaller skill set as outputs. This step prevents exponential growth of the synthesis procedure in the variable size. Finally, the manager synthesizes a strategy $\strategy_{\spec'}$ for each reduced specification.

If $\spec'$ is unrealizable, this indicates that gait-fixed MICP feasibility checks (Sec.~\ref{sec:motion_primitive}) failed to produce sufficient skills to reach the request under the given terrain. To address this, we rely on gait-free MICP (Sec.~\ref{sec:gait_free_micp}), which relaxes the fixed contact sequence and timing constraints while retaining the continuous dynamics. Since gait-free MICP is computationally demanding, we employ symbolic repair \citep{pacheck2022physically} to generate targeted symbolic suggestions that minimize unnecessary solves. Symbolic repair works by systematically modifying pre- and postconditions of existing skills to propose new candidates. These suggestions are then validated by gait-free MICP; if feasible, they are incorporated into the specification, making $\spec'$ realizable. An illustration of this process is shown in Fig.~\ref{fig:preliminary_nine_squares}(c), where repair alters the postcondition of a skill from reaching $(x_2, y_0)$ to instead target $(x_0, y_0)$, thereby enabling the robot to satisfy the request.
\subsection{Gait-Free MICP}\label{sec:gait_free_micp}
We aim to implement the suggested skills by reconfiguring the contact sequence and timing to generate new locomotion gaits. This can be achieved by solving a gait-free version of MICP in Eq.~(\ref{ocp:mip_general}). The gait-free MICP retains all continuous decision variables and constraints from the gait-fixed MICP but modifies the contact state-dependent constraints, as the contact sequence and timing are no longer fixed. Instead of using $\v H_{r,k,j}$, we introduce a different set of binary variables $\v H_{r,m,j}$ for defining the contact state for each foot at each time step explicitly. $r$ and $j$ still represent the $r^{\rm th}$ convex region and $j^{\rm th}$ foot, respectively, while $m \in \range{M}$ denotes the $m^{\rm th}$ time step, evenly discretized with $\Delta t_m$ over the entire $M$ time steps. We use $\mathcal{O}_{m,m+1}$ to denote the set of continuous time steps that span from the $m^{\rm th}$ to the ${m+1}^{\rm th}$ binary time step. As a result, the following constraints are modified:
\subsubsection{Safe Region Constraint}
The safe region constraint is defined in Eq.~(\ref{eq:safe_region_2}) - (\ref{eq:safe_region_2_eq}) similar to the gait-fixed case in Eq.~(\ref{eq:safe_region}) - (\ref{eq:safe_region_eq}) when a binary variable $\v H_{r,m,j}$ is activated. Different from Eq~(\ref{eq:sum_contact_binary}), the summation of the binary variables at $m^{\rm th}$ time step for the $j^{\rm th}$ foot is allowed to be zero to generate a swing phase as shown in Eq.~(\ref{eq:sum_contact_binary_2}).
\begin{align}
\nonumber &\forall r \in [R], m \in [M], j \in [n_f]\\    \label{eq:safe_region_2} &\v H_{r,m,j} \Rightarrow \v A_r \v p_j[i] \leq \v b_r, \forall i \in \mathcal{O}_{m,m+1}\\
    \label{eq:safe_region_2_eq} & \hspace{1.6cm} \v A_{\text{eq},r} \v p_j[i] = \v b_{\text{eq},r}, \forall i \in \mathcal{O}_{m,m+1}
        \\\label{eq:sum_contact_binary_2} &\sum_{r=0}^{R-1}\v H_{r,m,j} \leq 1
        \\&\v H_{r,m,j} \in \{0, 1\}
\end{align}

\subsubsection{Frictional and Contact Constraints}
Different from the gait-fixed case in Eq.~(\ref{eq:force_non_zero}) - (\ref{eq:force_zero}), the activation of frictional and contact constraints is purely decided by the binary variables.
\begin{align}
    \nonumber & \forall r \in [R], m \in [M], j \in [n_f]\\    \label{eq:force_non_zero_2} & \v H_{r,m,j} \  \Rightarrow \ \v f_j[i] \cdot \v n_r(\v p_j^{xy}[i]) \ge 0, \forall i \in \mathcal{O}_{m,m+1}\\
    \label{eq:force_friction_cone_2}
        & \hspace{1.8cm} \v f_j[i] \in \mathcal{F}_r(\mu, \v n_r, \v p_j^{xy}[i]), \forall i \in \mathcal{O}_{m,m+1}
         \\\label{eq:ee_velocity_zero_2} &
        \sum_{r=0}^{R-1}\v H_{r,m,j} = 1 \ \Rightarrow \  \dot{\v p}_j[i]=0, \forall i \in \mathcal{O}_{m,m+1}
        \\\label{eq:force_zero_2} &
         \sum_{r=0}^{R-1}\v H_{r,m,j} = 0 \ \Rightarrow \ \v f_j[i] = 0, \forall i \in \mathcal{O}_{m,m+1}
\end{align}

Compared to gait-fixed MICP, gait-free MICP introduces more binary variables to determine contact states, resulting in longer computation times. As such, it is only triggered after performing symbolic repair.

\section{Online Execution}\label{sec:online_execution}
Due to the inevitable disparity between offline synthesis and real-world online terrain conditions, such as variations in terrain shape and position, additional efforts are required to bridge this gap and prevent potential execution failures. The online execution module takes in terrain states $\inpstateterrain \subseteq \inpterrain$ after abstracting the terrain polygons online, a request state $\inpstaterequest \subseteq \inprequest$,
and a strategy $\strategy$
from the offline synthesis module (Sec.~\ref{sec:reactive_gait_synthesis}).
This module then executes the strategy automaton $\strategy$
by solving a MICP problem online again with the actual perceived terrain polygons, potentially different from the ones abstracted offline, 
to generate a reference trajectory for each transition (Sect.~\ref{sec:automaton_online_evaluation}). This reference trajectory 
along with the corresponding contact information will be further tracked by a tracking controller in real time
(Sec.~\ref{sec:tracking_control}).
As shown by the blue lines in Fig.~\ref{fig:combined_offline_online}(c),
if the robot encounters an unseen terrain state, 
we leverage online symbolic repair to create new skills that handle the unexpected terrain state and failure transition at runtime 
(Sec.~\ref{sec:online_repair}).
After reaching the request state $\inpstaterequest$,
the robot resets the strategy with a new terrain state, request goal states,
and repeats the execution
until reaching the final goal state.


\subsection{Strategy Automaton Execution and Online MICP Modifications}\label{sec:automaton_online_evaluation}
\begin{figure}
    \centering
    \includegraphics[width=\linewidth]{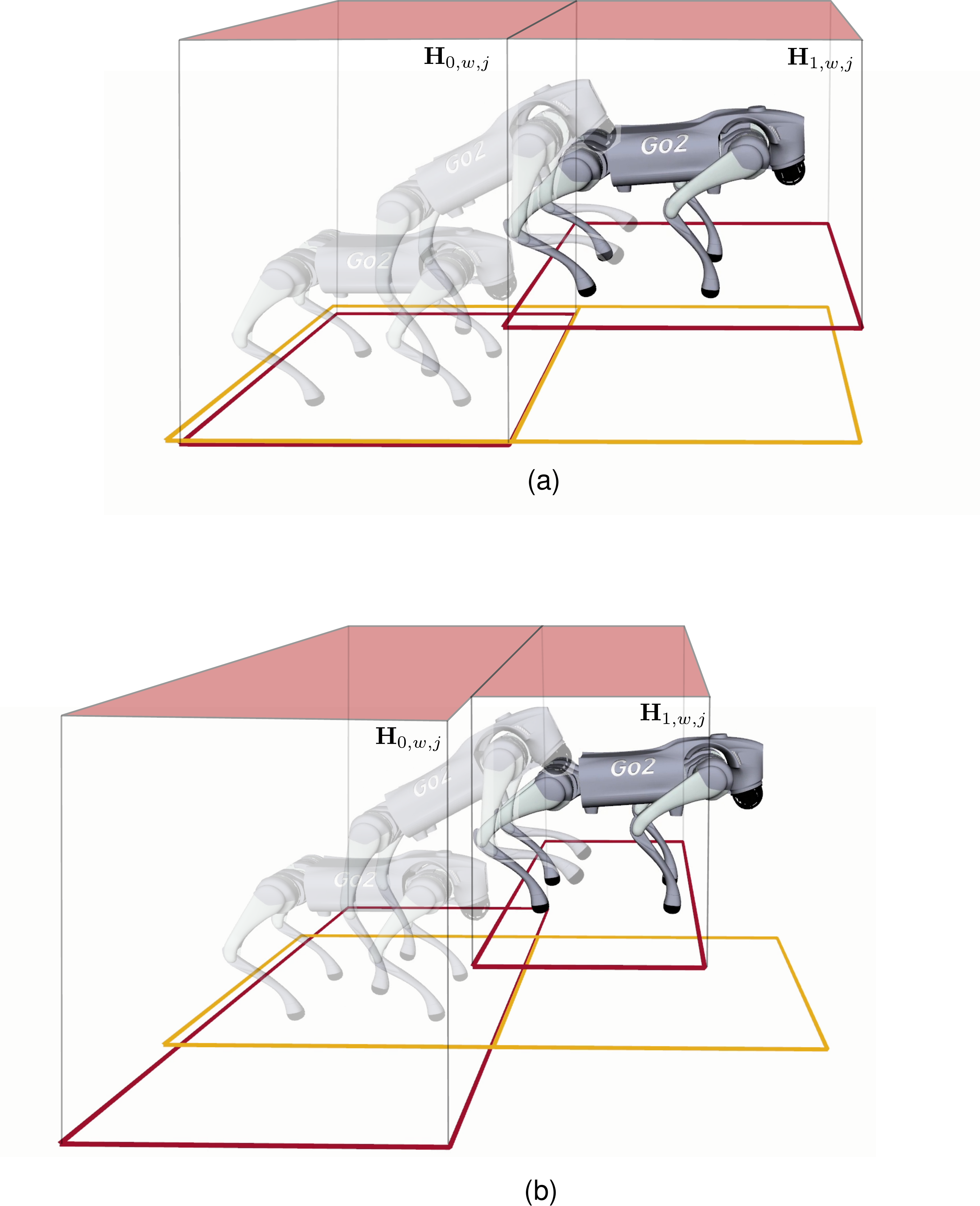}
    \caption{Demonstrations of (a) offline MICP for a skill transitioning from a flat terrain to a high terrain with predefined polygons; (b) online MICP for the same skill but with online terrain polygons.}
    \label{fig:online_micp}
\end{figure}

The red path in Fig.~\ref{fig:combined_offline_online}(c) represents the automaton online execution process. Before transitioning to the new symbolic state, an online gait-fixed MICP is solved according to the skill provided by the automaton. The automaton advances only if the MICP successfully finds a solution.
Slightly different from the gait-fixed MICP formulation during the offline phase, the online gait-fixed MICP is executed given terrain information from a terrain segmentation module that produces labeled convex polygons.
For example, a skill transitioning from flat terrain to high terrain, as shown in Fig.~\ref{fig:online_micp}(a),  uses  predefined, homogeneous terrain polygons for feasibility check during offline synthesis. Given that, Fig.~\ref{fig:online_micp}(b) illustrates how the  terrain polygons detected online for the same skill may differ from the offline ones in both position and geometry.
\revised{Note that, perception is intentionally treated as out of scope in this paper. We assume an upstream segmentation pipeline (e.g., elevation-mapping with plane segmentation~\citep{miki2022elevation,fankhauser2018robust}) provides labeled convex polygons; in our hardware experiments we use motion-capture-aided manual labeling of polygons, which is sufficient to validate the planning and control framework, although it does not represent a fully autonomous perception stack. Obtaining convex collision-free regions in unstructured environments is non-trivial, and we make no claim that our framework includes a solution to this perception problem. Robust perception integration is left for future work.}
In addition, the following modifications are highlighted when attempting to transition between two symbolic states in the automaton.

\subsubsection{Robot Pose Re-Targeting}
Since the final targeting condition significantly influences the reference trajectory and the feasibility of the MICP problem, we evaluate the final condition by solving a kinematic feasibility problem (a simplified MICP in Eq.~(\ref{ocp:mip_general})) before solving the online MICP. Compared with the gait-fixed MICP, the dynamics equation (\ref{eq:dynamics}) is replaced by enforcing the center of mass (CoM) to lie inside a convex support polygon with all feet in stance to ensure static stability. In addition, a hard constraint is added to ensure that the modified robot pose stays within a certain threshold compared with the original desired pose. The cost function is simplified to stay as close as possible to the original desired pose. The higher-order terms of body position, orientation, and end-effectors are excluded from the decision variables. All the other constraints not involving those higher-order terms remain the same. Once the kinematic feasibility problem is solved successfully, the online MICP is solved using the modified re-targeted final condition and reference trajectory.

\subsubsection{Collision Avoidance and Swing Foot Constraints}
\revised{Our framework handles obstacles at \emph{two distinct levels}, and we will explicitly elaborate which level is responsible in which scenario.
\begin{itemize}
    \item \emph{Symbolic-level obstacle avoidance.} When an obstacle is large enough that an entire abstracted cell becomes unsteppable, it is classified as the \textit{obstacle} terrain class in the local symbolic abstraction (Sec.~\ref{sec:preliminaries_abstraction}). The symbolic specification rules out transitions into such cells, and if an obstacle appears at runtime---as in the rebar-with-obstacle hardware experiment, where the obstacle is detected via motion capture and added to the local abstraction online---runtime symbolic re-synthesis (Sec.~\ref{sec:online_repair}) finds a detour around it. No collision-region variables need to be added to the online MICP in this case.
    \item \emph{MICP-level collision-free region selection (this subsection).} For smaller obstacles that the swing foot must clear within a single symbolic transition, we encode collision avoidance directly in the online MICP. Because the resulting trajectory is sent directly to the tracking controller, the quality of end-effector (EE) trajectories at this layer is critical for tracking performance. We introduce binary variables $\v H_{s,w,j}$ that constrain the swing-foot trajectory of foot $j$ to lie inside one of $S$ pre-labeled collision-free convex polytopes at each time step. These free-space polytopes come from the same labeled-polygon segmentation pipeline as the steppable terrain, with a separate label set distinguishing free-space polytopes from steppable terrain polygons. In simulation, this mechanism is exercised in the \textit{Unstructured--8} case with collision avoidance enabled (see Table~\ref{tab:micp_speed}), where the binary-variable count rises to 256.
\end{itemize}
We do not perform full-body, mesh-level collision checking against arbitrary geometry; the body footprint is implicitly handled by steppable-region selection keeping each foot inside a labeled terrain polygon. The formulation here trades fidelity for online tractability.}

Specifically, $s \in \range{S}$ represents the $s^{\rm th}$ convex collison-free region and $w \in \range{W}$ denotes the $w^{\rm th}$ time step, evenly discretized with $\Delta t_w$ over the entire $W$ time steps. Fig.~\ref{fig:online_micp}(b) demonstrates two collision-free regions.
\begin{align}
    \nonumber & \forall s \in [S], w \in [W], j \in [n_f]\\\label{eq:collision_avoidance} &\v H_{s,w,j} \Rightarrow \v A_s \v p_j[i] \leq \v b_s, \forall i \in \mathcal{C}_{w,w+1}
        \\&\sum_{s=0}^{S-1}\v H_{s,w,j} = 1
        \\&\v H_{s,w,j} \in \{0, 1\}
\end{align}
where $\mathcal{C}_{w,w+1}$ denotes the set of continuous time steps that span from the $w^{\rm th}$ to the ${w+1}^{\rm th}$ binary time step. \revised{We emphasize that the collision-free region selection in Eq.~(\ref{eq:collision_avoidance}) and the steppable-region selection in Eq.~(\ref{eq:safe_region}) handle different geometric primitives: the steppable-region constraint forces \emph{stance-foot} placement onto a labeled steppable terrain polygon at the first stance time step, while the collision-free region constraint here keeps the \emph{swing-foot} trajectory inside a free-space polytope across all time steps within a swing phase. The two are complementary and not redundant.}

To enable larger swing foot clearance from the terrain, we also add constraints to force the swing foot height above a user-defined threshold $h_{\text{swing}}$, given the fixed contact timing. \revised{This constraint targets cleanly clearing the terrain during a transition---for example, when jumping onto a higher platform. While such clearance could in principle be enforced through the collision-free region selection alone, doing so would require a fine time-step discretization of the binary variables and would tend to yield trajectories that only marginally clear the terrain. The height threshold $h_{\text{swing}}$ instead provides a direct, continuously enforced clearance margin that is straightforward to tune.}

\revised{Note that the collision-avoidance and swing-foot constraints are also incorporated in the offline MICP as part of the feasibility-checking rules for the relevant terrain transitions. We highlight them here because their impact on online tracking performance is substantially more critical.}
\subsection{Runtime Repair}\label{sec:online_repair}
The runtime repair procedure takes as input a terrain state $\inpstateterrain \in \inpstatesterrain$, a request state $\inpstaterequest \in \inpstatesrequest$, and a Boolean formula $\disallowedtrans$ that encodes disallowed transitions identified at runtime. We first update the system’s hard constraint $\syssafety$ in the specification $\spec$ (see Sec.~\ref{sec:spec_gen}) to $\syssafety \land \square \disallowedtrans$, ensuring that the specification reflects these newly discovered restrictions. Next, the high-level manager constructs a partial evaluation of $\spec$ over $\inpstateterrain$ and $\inpstaterequest$. Symbolic repair is then applied to generate additional robot skills and synthesize a new strategy that accommodates the current terrain and request states, following the approach in Sec.~\ref{sec:manager_and_repair}.


\section{Tracking Control}\label{sec:tracking_control}
The tracking control module adopts an MPC-Whole-Body-Control (WBC) hierarchy, ensuring precise end-effector (EE) and centroidal momentum tracking. To smoothly and efficiently execute the strategy considering the potential computational delay in solving MICP, a harmonic coordination module between the strategy automaton roll-out and the tracking module is designed (Sec.~\ref{sec:coordination}).

\subsection{Tracking Control Module}
\subsubsection{Nonlinear Model Predictive Control (NMPC)}\label{subsec:nmpc}
The NMPC  operates independently from the symbolic planning module, tracking reference trajectories generated by the online MICP using a more accurate dynamics model. In this work, the NMPC accounts for the robot's centroidal dynamics and kinematics, similar to approaches in \citep{dai2014whole,chignoli2021humanoid,sleiman2021unified}, presenting a more accurate model than the simplified angular dynamics in the MICP. An analytical inverse kinematics solution based on the MICP output provides the full joint state reference trajectory for the NMPC.
In addition to standard frictional and contact constraints and base tracking costs, an additional cost function for EE reference tracking from the MICP is included to enhance precision. The NMPC optimization problem is formulated as a Sequential Quadratic Program (SQP) and solved using the OCS2 library \citep{OCS2}, with a time horizon of one second and 100 knot points.

\subsubsection{Whole-Body Control (WBC)}
While NMPC operates at the centroidal level to generate dynamically feasible trajectories, WBC ensures precise execution by resolving full-body state control, including joint torques, accelerations, and contact forces.
The whole-body controller tracks the NMPC trajectory by solving a weighted quadratic program (QP) \citep{kuindersma2014efficiently} at 500 Hz. To improve the performance of agile motions such as leaping and jumping, centroidal momentum tracking \citep{wensing2013generation} is included. This MPC-WBC hierarchy allows the system to handle both centroidal-dynamics-level and full-body-dynamics-level control in a coordinated manner, ensuring robustness in real-world deployment especially for highly dynamic motions such as jumping.

\subsubsection{State Estimation}
The state estimator employs a linear Kalman filter similar to that in \citep{bledt2018cheetah}, with the addition of OptiTrack motion capture (mocap) measurements fused alongside IMU data and joint encoder readings to achieve accurate body position estimation. For agile motions such as jumping, we observed improved base-height estimation by assigning higher noise values to the mocap feedback during aerial phases, due to its limited 120 Hz update rate.
\subsection{Coordination between Strategy Execution and Tracking Module}\label{sec:coordination}
\begin{figure}
    \centering \includegraphics[width=\linewidth]{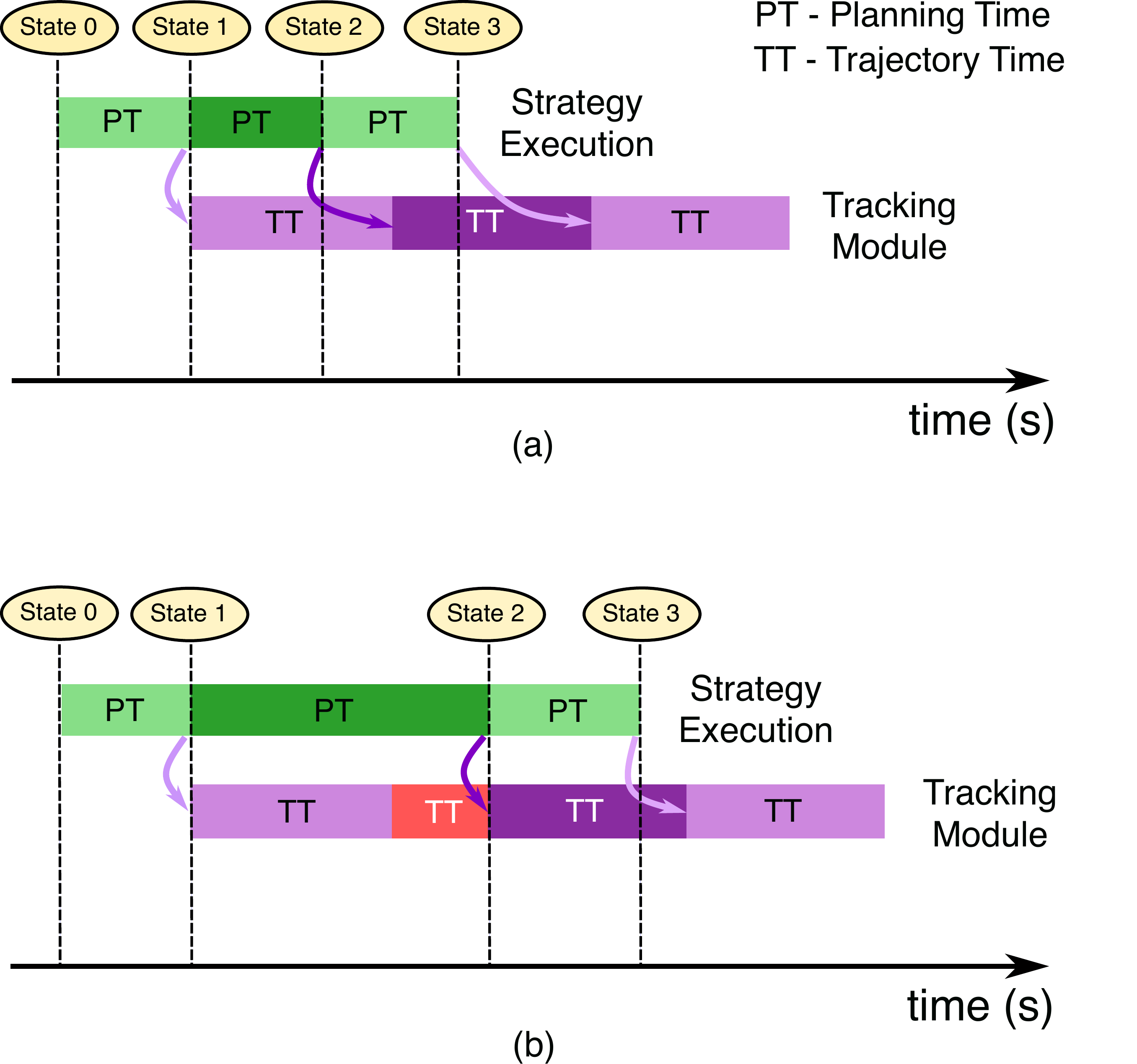}
    \caption{Coordination between strategy execution and tracking control module.
    (a) The MICP solving time (denoted by PT) for transitioning from State 1 to State 2 is shorter than the time horizon of the previous reference trajectory (denoted by TT), allowing the seamless appending for the new trajectory.
    (b) The MICP solving time exceeds the time horizon of the previous reference trajectory, requiring the robot to come to a stop (red) and wait for the new trajectory whenever available.}
    \label{fig:micp_mpc_automaton}
\end{figure}
Due to the potential delays in solving MICP during the strategy automaton roll-out, a harmonic coordination between the strategy execution and tracking control module is essential. The tracking module's reference trajectory is updated dynamically alongside the strategy execution process, as shown in Fig.~\ref{fig:micp_mpc_automaton}. The overall process can be summarized as follows:

\subsubsection{Strategy Execution} The strategy automaton execution thread begins by solving the online gait-fixed MICP, starting from State 0 in the automaton. Once the MICP solution is obtained, it immediately sends the reference trajectories and corresponding contact information to the tracking control module. The automaton then progresses to the next transition, using the final condition in the previous trajectory segment as the new initial condition. In Fig.~\ref{fig:micp_mpc_automaton}, the colored blocks in the strategy execution timeline indicate the time taken for each MICP solve during the automaton roll-out. PT and TT denote the planning time and trajectory time, respectively.

\subsubsection{Tracking Control Module}
The MPC-WBC tracking control module receives time-indexed reference trajectories, which can be updated in real-time. After completing the tracking of the current reference trajectory, the robot returns to a stance phase, ready to accept new reference trajectories. As depicted in Fig.~\ref{fig:micp_mpc_automaton}, the colored blocks in the tracking control module represent the time horizon of each reference trajectory sent by the strategy execution thread. Once the initial reference trajectory is generated, the MPC-WBC thread begins tracking it using more accurate dynamics models, as described in Sec.\ref{sec:tracking_control}.

\subsubsection{Delay Handling}
For more complex scenarios involving a larger number of terrain polygons, the MICP solving time may increase and exceed the time horizon of the generated reference trajectory. Fig.~\ref{fig:micp_mpc_automaton}(a) and (b) illustrate two cases of MICP solving times:
Case (a): The solving time for transitioning from automaton State 1 to State 2 is shorter than the time horizon of the previous reference trajectory. In this case, the new reference trajectory is seamlessly appended to the existing one, and the robot continues tracking the previous trajectory.
Case (b): The solving time exceeds the time horizon of the previous reference trajectory, and the robot has already entered a stance phase (red block) when the MICP solution is obtained. In this case, the robot waits until the newly generated reference trajectory is sent based on the current system time (red block in Fig.~\ref{fig:micp_mpc_automaton} (b)).



\section{Simulation}\label{sec:experiments}
We present examples of maneuvering across diverse environments in a Gazebo simulation to demonstrate the framework's efficacy, scalability, and generalizability.
In addition, benchmarking experiments are conducted to further compare our proposed framework with pure-MIP approaches.

\revised{\subsection{Implementation Details}\label{sec:impl_details}
All MICP problems---including the offline gait-free MICP, the online gait-fixed MICP, the kinematic-feasibility re-targeting MICP, and the pure-MIP benchmark---are solved using Gurobi~9~\citep{gurobi}. The NMPC tracker is implemented with the OCS2 library~\citep{OCS2}. All offline-synthesis, simulation, and online-planning experiments are run on a desktop workstation equipped with an Intel Core i9-13900K (24 cores) and 64\,GB of RAM. Reported solve times reflect this desktop-class CPU and should be interpreted as such; onboard, low-power deployment is left as future work.}

\subsection{Robot Setup}
\begin{figure}[t]
    \centering
    \includegraphics[width=0.85\linewidth]{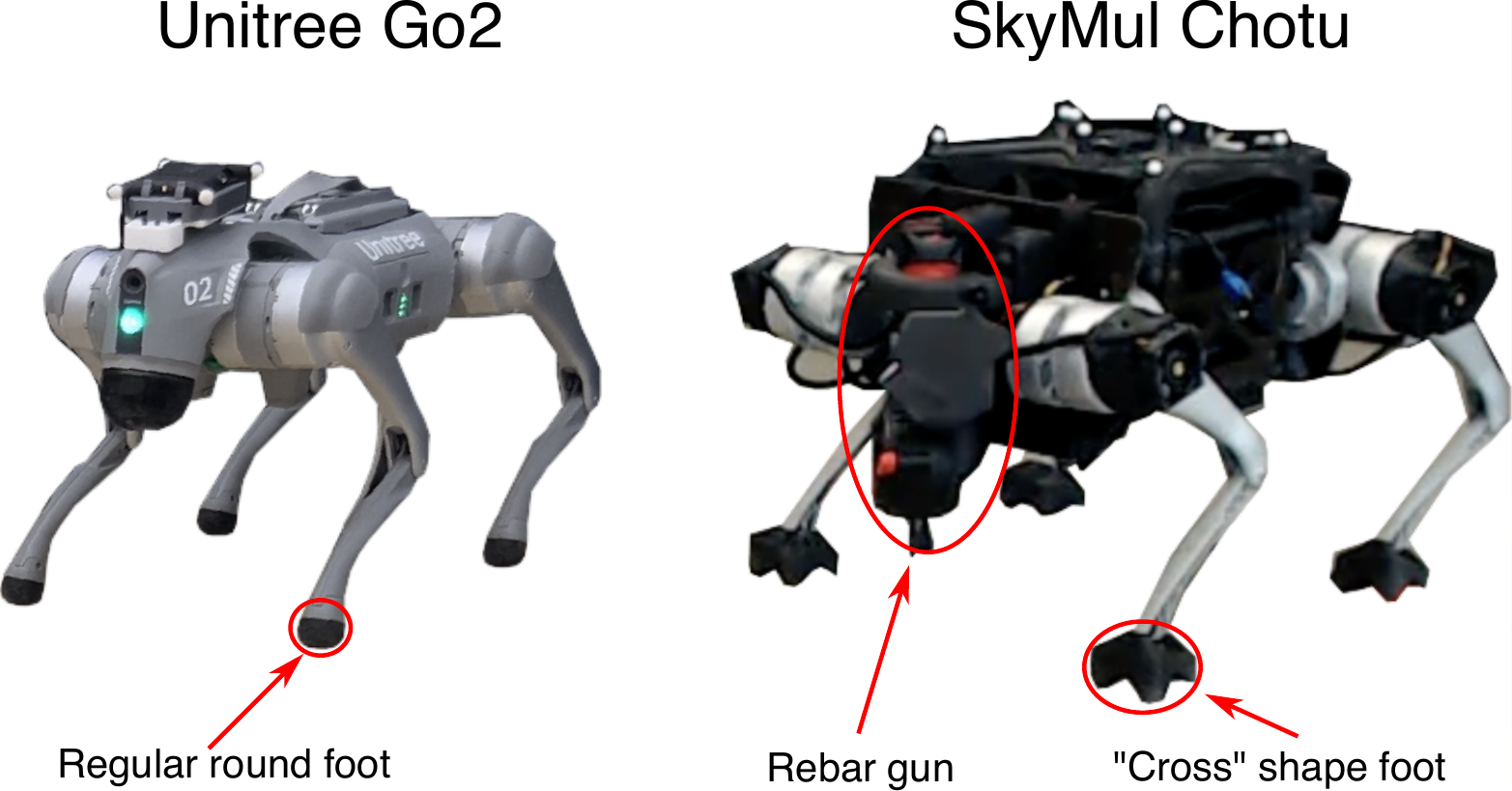}
    \caption{Hardware platforms used for experiments: Unitree Go2 and SkyMul Chotu (equipped with a rebar gun and ``cross"-shaped feet).}
    \label{fig:hardware_info}
\end{figure}

As shown in Fig.~\ref{fig:hardware_info}, we verify our algorithms on two separate robot platforms, Unitree Go2 and SkyMul Chotu (a modified Unitree Go1 robot dedicated for rebar tying tasks). The SkyMul Chotu is equipped with a rebar gun mounted on its head and customized ``cross"-shaped feet designed for reliable standing on rebars. Note that in simulation, this foot shape is simplified to a regular round foot to avoid the complexity of simulating contact with irregular surfaces. In our planner, we model both cases as point contact.
The parameters for each robot are defined in Table~\ref{tab:robot_parameter} and used in our implementation, including the robot mass $m$, torque limit $\vg \tau_{\text{max}}$ and the reference joint position $\v q_j^{\text{ref}}$ for a single leg with three joints, the reference foot position relative to the base frame $^\mathcal{B} \v p_j^{\text{ref}}$ and the maximum deviation of the foot position $\v p_j^{\text{max}}$ for the front left leg ($j=0$), and the moment of inertia of the approximated single rigid body $\v I$.
\begin{table}[t]
\centering
\footnotesize
    \caption{Robot Parameters}
    \setlength{\tabcolsep}{4pt}
     \begin{tabular*}{\linewidth}{@{\extracolsep{\fill}}lll}
         \hline
         \textbf{Parameter} & \textbf{Unitree Go2} & \textbf{SkyMul Chotu}\\\hline
         $m$ (kg) & 15 & 20\\
         $\vg \tau_{\text{max}} $ (N $\cdot$ m) & (23.5, 23.5, 45.4) & (23.5, 23.5, 33.5)\\
         $\v q_j^{\text{ref}}$ (rad) & (0.0, 0.72, -1.44) & (0.0, 0.72, -1.44)\\
         $^\mathcal{B} \v p_j^{\text{ref}}$ (m) & (0.1805, 0.1308, -0.29) & (0.2118, 0.210, -0.30)\\
         $\v p_j^{\text{max}}$ (m) & (0.15, 0.1, 0.15) & (0.15, 0.1, 0.15)\\
         $\v I$ (kg $\cdot$ m$^2$) & diag(0.152, 0.369, 0.388) & diag(0.396, 0.915, 1.107)\\
         \hline
    \end{tabular*}
    \label{tab:robot_parameter}
\end{table}


\begin{figure*}[t!]
    \centering
    \includegraphics[width=\linewidth]{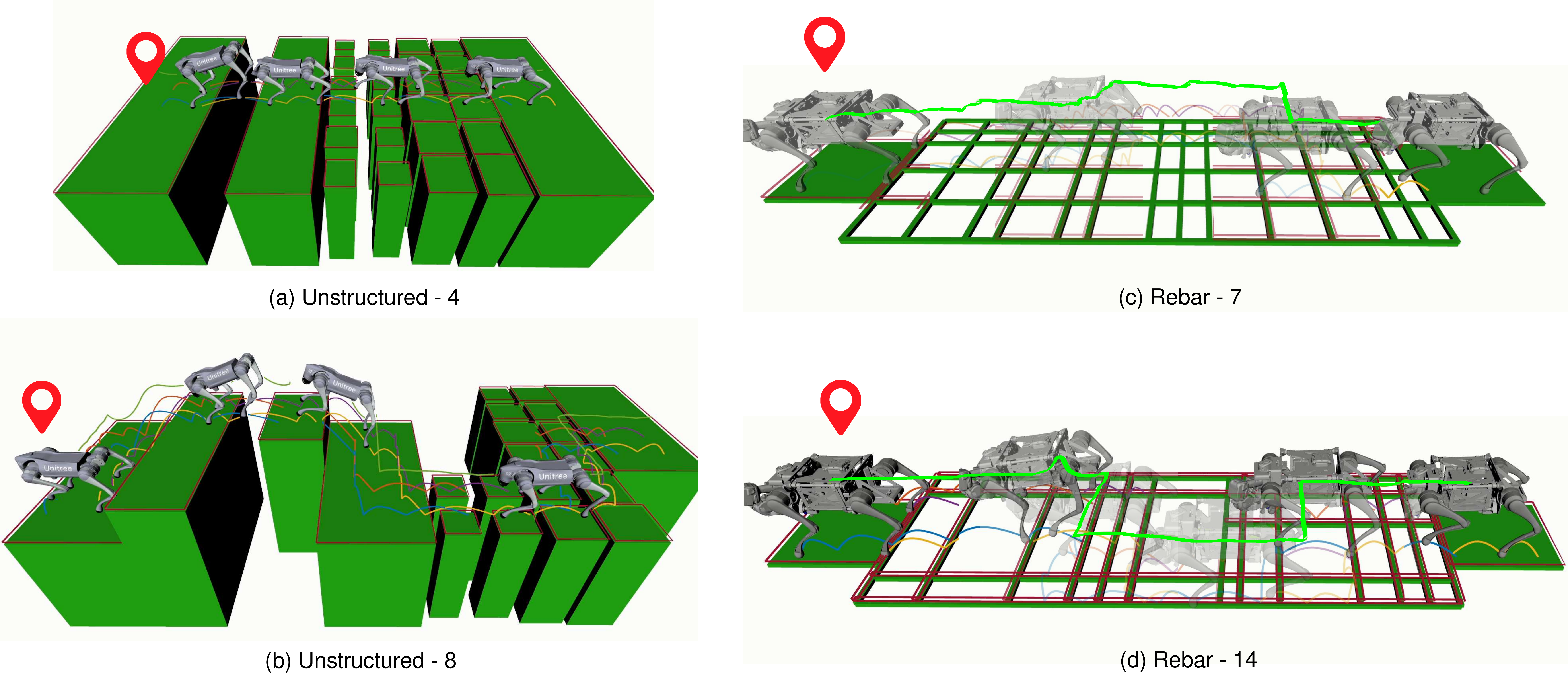}
    \caption{Snapshots from online execution across the four simulation scenarios: (a) unstructured terrain with four terrain types, (b) unstructured terrain with the full eight-type abstraction, (c) rebar terrain with the 7-type abstraction, and (d) rebar terrain with the 14-type abstraction. Green trajectories in (c) and (d) are detours produced by runtime re-synthesis after online MICP solving failures.}
    \label{fig:online_terrains}
\end{figure*}

\begin{table*}[t]
\centering
\footnotesize
    \caption{Offline Synthesis and Repair Time}
    \setlength{\tabcolsep}{4pt}
     \begin{tabular*}{\linewidth}{@{\extracolsep{\fill}}cccccc}
         \hline
         \textbf{Scenario} & \textbf{Grid Size} &
         \begin{tabular}{@{}c@{}}\textbf{Terrain and Request State Pairs}\\ \textbf{(Success/Total)}\end{tabular}
         & \begin{tabular}{@{}c@{}}\textbf{Number of Skills} \\ \textbf{(Original/New/Total)}\end{tabular} & \textbf{Symbolic Repair Time (s)} & \begin{tabular}{@{}c@{}}\textbf{Feasibility Check Time (s)}\\ \textbf{(Gait-Fixed/Gait-Free)}\end{tabular}\\\hline
         Unstructured - 4 & $3 \times 3$ & 169/169 & 18/\textcolor{red}{13}/64 & 8.07 &  18.97/449.94 \\
         & $5 \times 5$ & 129/158 & 19/\textcolor{red}{9}/64 & 77.16 &  20.26/241.42 \\\hline
         Unstructured - 8 & $3 \times 3$ & 250/264 & 19/\textcolor{red}{20}/256 & 14.90 & 22.92/382.63 \\
         & $5 \times 5$ & 179/246 & 20/\textcolor{red}{16}/256 & 93.21 & 18.59/196.88\\\hline
         Rebar - 7& $3 \times 3$ & 
         196/196 & 18/\textcolor{red}{11}/196 & 7.44 &  15.63/417.55
         \\
         & $5 \times 5$ & 196/196 & 18/\textcolor{red}{10}/196 & 21.63 &  14.83/374.89 \\\hline
         Rebar - 14 & $3 \times 3$ & 237/237 & 20/\textcolor{red}{29}/784 & 10.34 & 21.92/701.38\\
         & $5 \times 5$ & 196/196 & 20/\textcolor{red}{18}/784 & 24.31 & 23.06/453.94 \\\hline
    \end{tabular*}
    \label{tab:offline_synthesis_time}
\end{table*}

\subsection{Simulation Environment Setup}
To demonstrate the generalizability of our proposed framework, we evaluate it across two scenarios involving different terrain types, safe stepping regions, and robotic platforms—Unitree Go2 and SkyMul Chotu. Obstacles are modeled as a distinct terrain class, and obstacle avoidance is encoded as hard constraints within $\syssafety$. The requested waypoints are assumed to be free of obstacles.
We test both $3 \times 3$ and $5 \times 5$ grid abstractions. The $5 \times 5$ grid enables a longer planning horizon but introduces higher decision complexity. For discretization, we use a cell size of 0.8 m in unstructured terrains and 0.6 m for the rebar case. These values can be adjusted based on the requirements of specific deployment environments.


\subsubsection{Unstructured Terrain}
We abstract 8 terrain types—\textit{flat terrain}, \textit{high terrain}, \textit{low terrain}, \textit{dense stone}, \textit{sparse stone}, \textit{gap}, \textit{high gap}, and \textit{low gap}—based on classification criteria involving polygon heights, counts, and areas relative to the abstracted grid cells. A terrain is categorized as a \textit{gap} when the ratio of its overlapping area with a grid cell falls below a specified threshold.
We define two terrain configurations: one consisting of four selected terrain types, and another comprising all eight. These are illustrated in Figs.~\ref{fig:online_terrains}(a) and \ref{fig:online_terrains}(b), respectively.

\subsubsection{Rebar Terrain}
Inspired by efforts to automate labor-intensive rebar tying tasks on construction sites \citep{asselmeier2024hierarchical}, we investigate a second case study in which a quadrupedal robot navigates a rebar mat. Each rebar is modeled as a rectangular polygon with a 3 cm width. Unlike the stepping stone scenario, this setup presents a denser distribution of potential footholds due to the higher rebar density. Notably, the robot’s traversability depends on the directional sparsity of the rebars—whether aligned with or orthogonal to the robot’s facing direction—within an acceptable tolerance.

We abstract and classify rebar types based on horizontal sparsity (perpendicular to the robot’s facing direction) and vertical sparsity (parallel to it). Within each abstracted cell, the number and spacing of rebars are evaluated to classify the sparsity in each direction as \textit{dense} (0.05–0.15 m), \textit{sparse} (0.15–0.35 m), \textit{extreme sparse} (above 0.35 m), \textit{single}, or \textit{none}. To reduce complexity, combinations deemed too challenging—such as \textit{none} or \textit{single} in both directions—are treated as \textit{obstacle}.
Based on this abstraction, we define two example configurations comprising 7 and 14 rebar terrain types. The 7-type configuration in Fig.~\ref{fig:online_terrains}(c) samples rebar sparsity randomly between 0.15 and 0.35 m, omitting \textit{extreme sparse} regions. In contrast, the 14-type configuration in Fig.~\ref{fig:online_terrains}(d) includes sparsity ranging from 0.15 to 0.6 m, encompassing more complex and difficult combinations including \textit{extreme sparse} types.
\begin{figure*}[t]
    \centering
    \includegraphics[width=\linewidth]{online_repair_new_gap.pdf}
    \caption{Runtime symbolic repair triggered by an unforeseen \textit{gap} state in the \textit{Unstructured-4} / $3\times 3$ scenario, in which gap states are deliberately excluded offline. Repair proposes the missing transition and gait-free MICP synthesizes a leaping skill, taking $12.36$\,s total ($0.18$\,s symbolic repair + $12.18$\,s gait-free MICP).}
    \label{fig:online_repair}
\end{figure*}

\begin{figure}[t]
    \centering
    \includegraphics[width=\linewidth]{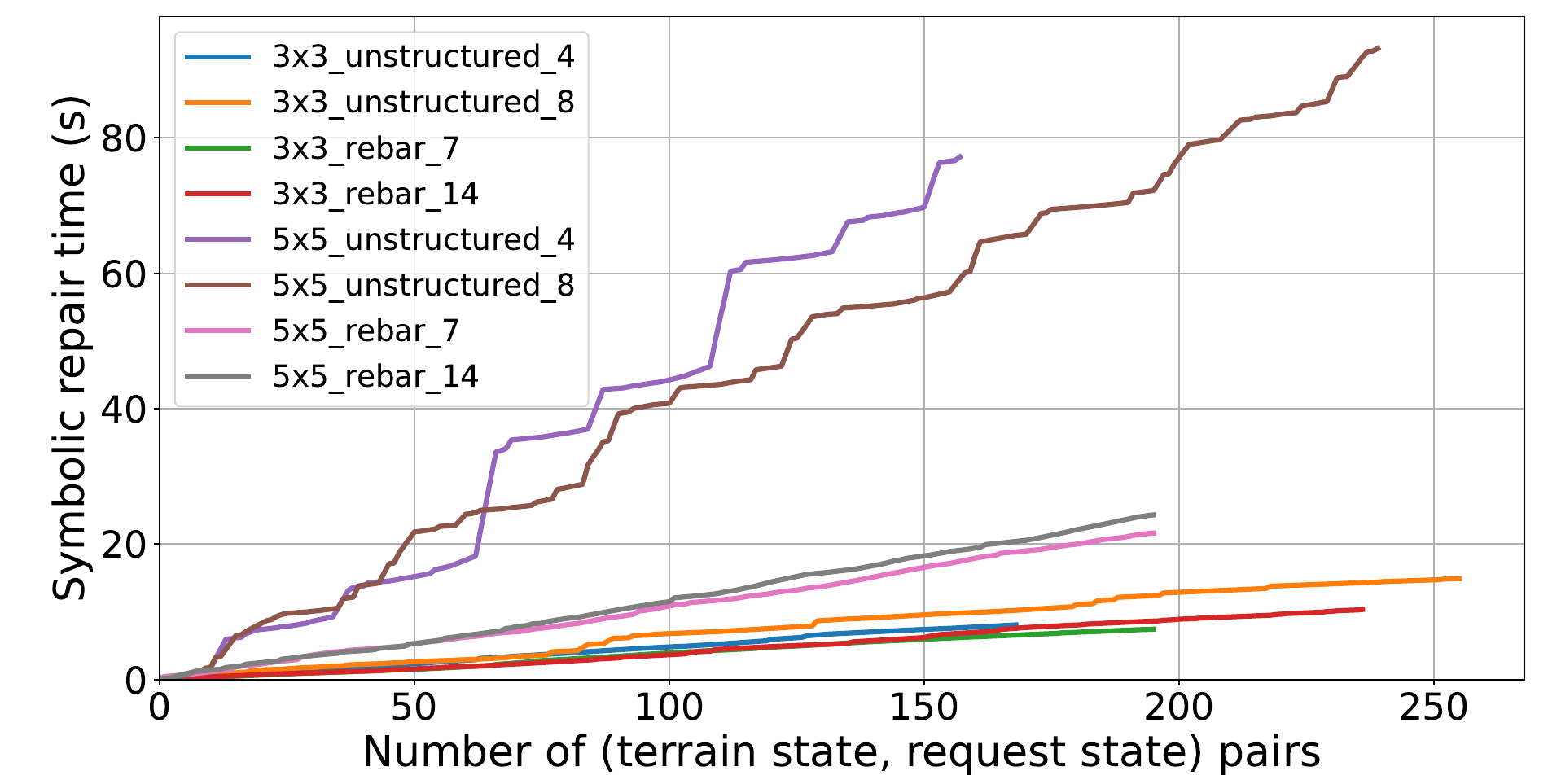}
    \caption{Offline symbolic-repair time versus the number of terrain--request state pairs for the $3\times 3$ and $5\times 5$ abstractions. Repair time grows approximately linearly in both cases; per state pair, the $5\times 5$ grid takes $478.1\%$ more time than the $3\times 3$ grid.}
    \label{fig:repair_vs_num_states}
\end{figure}

\begin{table*}
\centering
\small
    \caption{Online Gait-Fixed MICP Solving Time for Simulation}
    \setlength{\tabcolsep}{4pt}
     \begin{tabular*}{\linewidth}{@{\extracolsep{\fill}}cccccc}
         \hline
         \textbf{Scenario} & \textbf{Collision Avoidance} &
         \textbf{Binary Variables} &
         \textbf{Continuous Variables} &
         \textbf{Number of Polygons} &
         \textbf{Solve Time (s)}\\\hline
         Unstructured - 4 & No & 52 & 6583.5 & 6.5 & 0.37\\\hline
         Unstructured - 8 & Yes & \textcolor{red}{256} & 5166 & 2 & \textcolor{red}{2.65}\\
         & No & 69 & 7938 & 7.5 & 0.49\\\hline
         Rebar - 7& No & 70 & 7938 & 7.875 & 1.11\\
         \hline
         Rebar - 14 & No & 58 & 8142.75 & 6.25 & 1.09\\\hline
    \end{tabular*}
    \label{tab:micp_speed}
\end{table*}

\subsection{Offline Synthesis Results}
We evaluate the offline synthesis module for both \textit{Unstructured} and \textit{Rebar Terrain} scenarios by initializing the skill set with two trotting gaits ($2$ s and $3$ s). The inclusion of the longer-horizon trotting gait provides additional safer walking options from a practical standpoint. Prior to synthesis, we construct the set of possible terrain states $\inpstatesterrainexprposs$—assumed to be provided by the user—by systematically sweeping a local grid over the entire terrain map, based on the best available prior knowledge. The set of request states $\inpstatesrequestposs$ is defined as the top row of the local grid in the robot's facing direction, with the two corner cells also included to enable movement in diagonal directions.


Table~\ref{tab:offline_synthesis_time} reports 
\begin{enumerate*}[label=(\roman*),ref=\roman*]
    \item the number of terrain and request states pairs $(\inpstateterrain, \inpstaterequest) \in \inpstatesterrainexprposs \times \inpstatesrequestposs$, including both successful and total ones,
    \item the number of skills, 
    including the original,
    the newly discovered,
    and the total possible ones,
    and
    \item offline repair and feasibility checking times, including both from gait-fixed and gait-free MICP.
\end{enumerate*}
We first observe that the repair process successfully resolves $93.4\%$ of the terrain and request state pairs. The remaining cases are deemed unrepairable due to either unreachable request states caused by obstacles or inherent physical infeasibility. As highlighted in red, the number of newly discovered skills is reduced by $71.6$–$97.6\%$ compared to exhaustively checking all possible skills—each of which requires solving a computationally expensive gait-free MICP. On average, generating a new locomotion gait via gait-free MICP takes $27.23$ s, with a maximum of $145.66$ s. This demonstrates the effectiveness of offline repair in substantially reducing the number of costly gait-free MICP solves.

To investigate the scalability of offline repair,
we perform repair across various sizes of the terrain and request states.
As shown in
Fig.~\ref{fig:repair_vs_num_states},
the repair runtime grows linearly in the size of the terrain and request states for both $3\times 3$ and $5 \times 5$ grids.
While fewer terrain and request states handled offline reduce checked time, they may lead to more frequent unforeseen states during online execution.
Moreover, we note that
the slope of $5 \times 5$ cases in Fig.~\ref{fig:repair_vs_num_states}
are steeper than those of $3\times 3$
as a result of the increased state space.
On average,
repair takes 478.1$\%$ more time for $5\times 5$ grids than $3\times 3$ for one terrain and request states pair.
%
In practice, 
the perception range and the robot size limit the grid size, 
so we do not test beyond $5\times 5$ grids,
though the user can adjust the resolution.

\begin{figure*}[t]
    \centering
    \includegraphics[width=\linewidth]{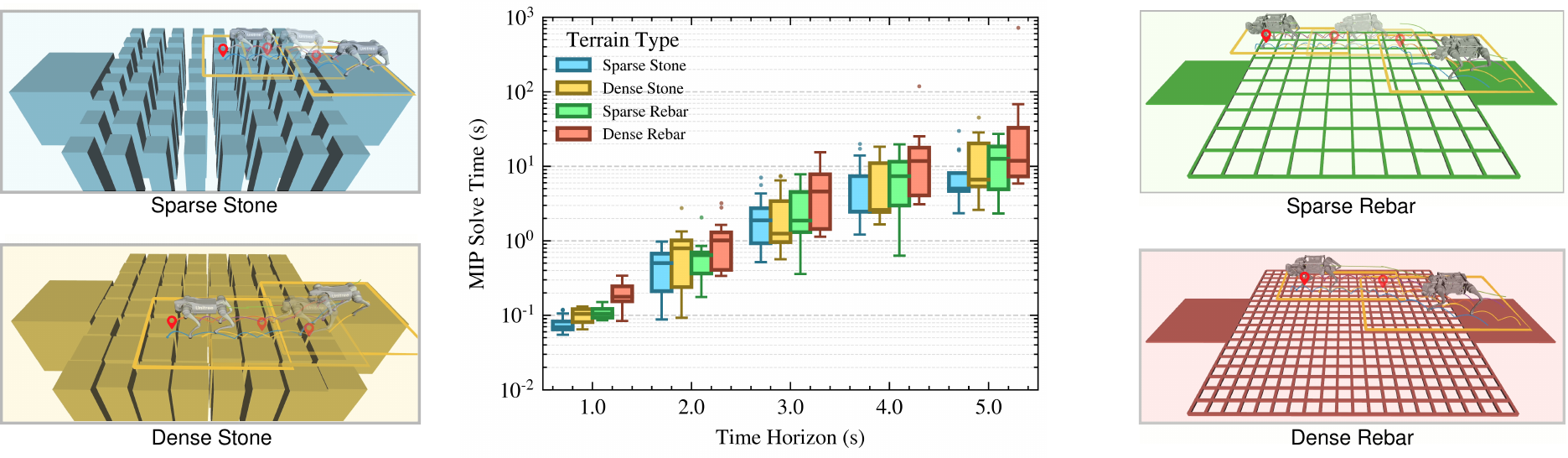}
    \caption{\revised{MIP solve time across four terrains---sparse stone (blue), dense stone (yellow), sparse rebar (green), dense rebar (red)---and horizons of $1$--$5$\,s, as box plots over $10$ seeded waypoints per scenario. In each terrain subfigure, selected traversal snapshots at planning horizons of $1$\,s (blue), $2$\,s (yellow), $4$\,s (green), and $5$\,s (red) are overlaid for visualization.}}
    \label{fig:pure_mip_benchmark}
\end{figure*}

\begin{figure}
    \centering
    \includegraphics[width=\linewidth]{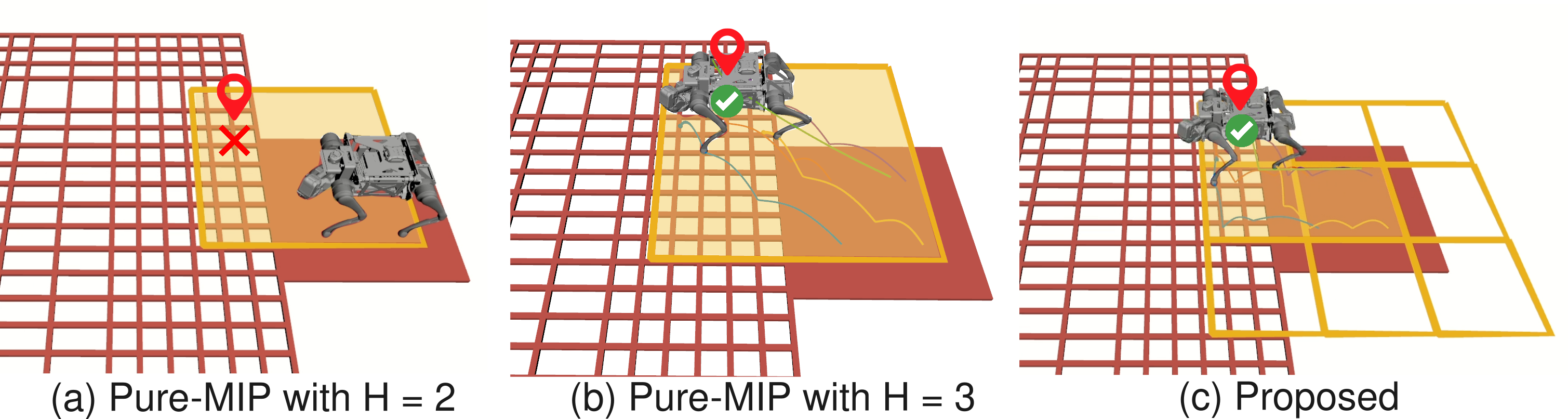}
    \caption{\revised{Representative dense rebar trial after randomly removing terrains for benchmarking different baselines: (a) Pure-MIP with $H{=}2$ s; (b) Pure-MIP with $H{=}3$ s; (c) Proposed approach with $3\times 3$ grid.}}
    \label{fig:dense_rebar_comparison}
\end{figure}

\subsection{Online Execution Results}
Fig.~\ref{fig:online_terrains} presents snapshots of Go2 and Chotu navigating various terrains during online execution. The robot is tasked with reaching a global waypoint using a naive global shortest-path planner. At each step, the local waypoint is selected as the closest non-obstacle cell in the local grid map pointing toward the global target.

In Fig.~\ref{fig:online_terrains}(a), the robot traverses a sequence of terrain types including \textit{flat terrain}, \textit{dense stepping stone}, \textit{sparse stepping stone}, and \textit{gap}. Fig.~\ref{fig:online_terrains}(b) showcases a more complex maneuver involving all eight terrain types, including elevation changes, where the robot adaptively selects suitable locomotion gaits. Similarly, Figs.~\ref{fig:online_terrains}(c) and (d) illustrate Chotu traversing rebar terrains with 7 and 14 defined rebar types, respectively.
Notably, the robot utilizes newly discovered leaping gaits with short aerial phases to execute challenging transitions—such as moving from lower to higher elevations or crossing over gaps and \textit{extreme sparse} rebar segments, as demonstrated in Figs.~\ref{fig:online_terrains}(b) and(d).

Table \ref{tab:micp_speed} summarizes the average number of decision variables, terrain polygons, and solve time for each scenario, computed across all step transitions in a full locomotion run with the online gait-fixed MICP.
On average, the online gait-fixed MICP takes $430$ ms to solve the unstructured terrain cases and $1.1$ s for rebar cases when the collision avoidance is disabled. 
The rebar scenarios exhibit $155.8\%$ longer solve times due to their overlapping, skewed rebar arrangement, and a larger number of terrain polygons. 
Additionally, 
enabling 
collision avoidance (necessary for \textit{Unstructured - 8} case to handle the elevation change) 
increases the number of binary variables
by $271.0\%$, 
increasing the solve time 
by $440.8\%$
(highlighted in red).

\revised{\subsection{Runtime Repair Case Study}\label{sec:runtime_repair_case_study}
As a case study to demonstrate the framework's capability to handle unforeseen terrains and online failures, we highlight the following run-time scenarios that trigger runtime repair or re-synthesis.

\subsubsection{Unforeseen Terrain States}
With limited prior knowledge of the global terrain map, the terrain states provided during offline synthesis may be insufficient when encountering new terrain configurations. Fig.~\ref{fig:online_repair} illustrates this in an \textit{Unstructured - 4 types} scenario with a $3 \times 3$ grid size, where terrain states involving \textit{gap} terrain are deliberately excluded from the offline phase. When the robot reaches a new terrain state that contains a \textit{gap}, runtime repair is triggered for the current terrain and request states. A new skill with a leaping gait is generated to successfully cross the gap. The runtime repair takes $12.36$\,s in total, of which the symbolic-repair stage accounts for $0.18$\,s and the gait-free MICP for the new leaping gait accounts for $12.18$\,s---confirming that the dominant cost lies in the dynamic-feasibility check, not the symbolic search.

\subsubsection{Solving Failure}
Another common runtime failure arises from MICP solving failures. Due to discrepancies between the predefined polygons used during offline synthesis and the actual online terrains---such as variations in shape and size---a skill deemed feasible offline may become infeasible when executed online, even if classified under the same symbolic transition. The detours shown in Fig.~\ref{fig:online_terrains}(c) and (d) illustrate the re-synthesis process triggered by solving failures in the 7-type and 14-type rebar cases. Solving failures occur more frequently in the rebar scenarios than in the unstructured ones, due to the higher uncertainty in the shapes and sizes of online rebar polygons. Re-synthesis takes $0.11$\,s and $0.07$\,s respectively for the two cases. Neither case triggers full runtime repair, because the existing skill set suffices to navigate to the goal under the failed transition.}

\subsection{Benchmark: Pure-MIP Planner}
To evaluate how the symbolic-continuous planner separation benefits the overall planning framework, we benchmark our method against \textit{pure-MIP} approaches, where we implement a planner to evaluate how computation time scales with the planning horizon and terrain complexity. \revised{The pure-MIP approach can also be treated as an ablation study for the symbolic planning.} Specifically, we examine five planning horizons ranging from $1$ s to $5$ s. In addition, we vary the terrain types by considering sparse stepping stones (0.15 m spacing), dense stepping stones (0.05 m spacing), sparse rebar (0.3 m spacing), and dense rebar (0.15 m spacing), as illustrated in Fig.~\ref{fig:pure_mip_benchmark} with specific colors.

\revised{To make the comparison precise: the pure-MIP baseline uses the \emph{same} convex relaxations of the dynamics (Eq.~(\ref{eq:dynamics})), the \emph{same} linearized friction-cone and fixed-Jacobian kinematic constraints, and the \emph{same} cost weights as our online gait-fixed MICP. The only differences are (i) the planner solves a single long-horizon gait-fixed MICP without symbolic-level decomposition, and (ii) to evaluate performance across different planning horizons, the local planning region is also expanded proportionally to the horizon to encompass all relevant terrain polygons, rather than being limited to the cells adjacent to a symbolic transition. The same Gurobi solver and settings are used in both cases.}

\revised{The pure-MIP planner uses an \emph{adaptive} local region whose maximum size scales with the planning horizon $H$ (in seconds): along each axis the region spans up to $\textrm{gs} + (\textrm{gs}/2)\cdot H$, where $\textrm{gs}$ is the symbolic-cell grid spacing, with a floor of $1.5\,\textrm{gs}$ corresponding to the $H{=}1$\,s setting. So the $H{=}2$\,s column reads as a like-for-like comparison for an equivalent two-cell transition footprint, while each additional second of horizon adds half a cell of reach. As shown in Sparse and Dense Rebar cases of Fig.~\ref{fig:pure_mip_benchmark}, the region further shrinks when the goal lies within the maximum range, avoiding redundant incorporation of terrain polygons in each MIP solve. 
In contrast, our proposed planner always plans against a fixed $3{\times}3$ symbolic grid centered on the robot, decoupling the per-MIP problem size from the total traversal distance. We omit the gait-free MICP from the benchmark because its solve time becomes prohibitively long once $H>2$\,s; the locomotion gait is repeated cyclically on a $1$\,s trotting pattern, and the planner receives a global waypoint and incrementally computes local waypoints as in the proposed framework.}

\revised{Fig.~\ref{fig:pure_mip_benchmark} also reports the per-MIP solve time as box plots aggregated over $10$ randomly sampled waypoints per scenario. Within the same terrain, the solve time grows almost exponentially with the horizon, and the dense variants consistently take longer than their sparse counterparts because more terrain polygons enter the MIP. The $H{=}1$\,s setting is too myopic to commit to scenarios that require multi-step planning, while $H\ge4$\,s pushes the median solve into multiple seconds even for the easier scenarios. The $H{=}2$\,s and $H{=}3$\,s horizons sit on the knee of this trade-off curve, balancing optimality and per-step compute, and we therefore adopt them as the two reference pure-MIP configurations for the end-to-end benchmark below.}

\revised{To assess closed-loop behavior beyond a per-MIP solve time, we run an end-to-end comparison on the dense rebar scenario, where the pure-MIP planner is most stressed. For each method, we dispatch $10$ seeded random waypoints; between trials the robot is reset back to the launch pose, and we randomly remove a fraction of the rebar polygons as shown in Fig.~\ref{fig:dense_rebar_comparison} to add randomness and make the scenario more challenging. Identical random seeds across methods ensure all baselines see the byte-for-byte same waypoints and terrain, so any difference in outcome reflects the planner alone. Fig.~\ref{fig:dense_rebar_comparison} also shows one representative instance on which the pure-MIP planner with short horizons (i.e., 1 or 2 s) stalls---the planning region is too limited for the solver to discover a feasible routing path
, whereas the proposed planner routes through a sequence of cells.}

\revised{Table~\ref{tab:pure_mip_benchmark} reports the success rate, mean traversal time of successful trials, and the mean per-module active computation per trial---broken down into MIP transition solving, the pre-flight MIP feasibility check (FC), reactive synthesis (strategy query plus strategy reload), and runtime repair (online specification regeneration). In this scenario, the proposed planner attains $90\%$ success rate with $18.5$\,s mean traversal time, matching the $H{=}3$\,s pure-MIP success rate ($90\%$) at roughly three quarters of the traversal time and at less than half the MIP transition cost ($5.26$\,s vs.\ $11.54$\,s per trial). The shorter $H{=}2$\,s pure-MIP setting reaches only $60\%$, indicating that its horizon is too myopic to recover from the gaps. The two ablation rows confirm the value of the online additions introduced in this work: disabling the FC drops the success rate substantially in both planners ($60\%\rightarrow20\%$ for pure-MIP at $H{=}2$\,s and $90\%\rightarrow60\%$ for the proposed planner); disabling the delay-aware coordination (DAC) preserves the $90\%$ success rate of the proposed planner but inflates the mean traversal time from $18.5$\,s to $23.4$\,s, because each symbolic transition now must wait for the previous trajectory to be fully tracked before the next MIP solve can start.}

\begin{table*}[t]
\centering
\caption{\revised{End-to-end benchmark on the dense rebar scenario ($10$ seeded waypoints, identical seed and terrain across methods, with randomly removed rebars per trial to add randomness). Traversal time and per-module times are mean values computed over successful trials only. Reactive synthesis combines the per-step strategy query and the per-spec-switch strategy reload. Dashes denote modules that do not exist in the corresponding method or are disabled by the ablation.}}
\label{tab:pure_mip_benchmark}
\setlength{\tabcolsep}{4pt}
{\color{black}
\begin{tabular}{lcccccc}
\hline
\textbf{Method} & \textbf{Success} & \textbf{Traversal} & \textbf{MIP trans.} & \textbf{MIP feas.} & \textbf{Reactive synth.} & \textbf{Runtime repair} \\
 & \textbf{rate} & \textbf{time (s)} & \textbf{solving (s)} & \textbf{check (s)} & \textbf{(query + load, s)} & \textbf{(s)} \\
\hline
Pure MIP ($H{=}2$\,s)                              & 60\% (6/10) & 14.76 &  3.04 & 0.21 & --   & --   \\
Pure MIP ($H{=}3$\,s)                              & 90\% (9/10) & 24.73 & 11.54 & 0.19 & --   & --   \\
Pure MIP ($H{=}2$\,s), no FC        & 20\% (2/10) &  9.00 &  1.38 & --   & --   & --   \\
Pure MIP ($H{=}3$\,s), no FC        & 40\% (4/10) & 14.23 &  5.17 & --   & --   & --   \\
\hline
Proposed                                           & 90\% (9/10) & 18.48 &  5.26 & 0.19 & 0.29 & 0.03 \\
Proposed, no FC                     & 60\% (6/10) & 20.08 &  5.85 & --   & 0.33 & 0.05 \\
Proposed, no DAC & 90\% (9/10) & 23.43 &  5.11 & 0.27 & 0.28 & 0.04 \\
\hline
\end{tabular}
}
\end{table*}

\subsection{Preliminary Study: Yaw Command Extension}\label{sec:extension}
\begin{figure}
    \centering
    \includegraphics[width=\linewidth]{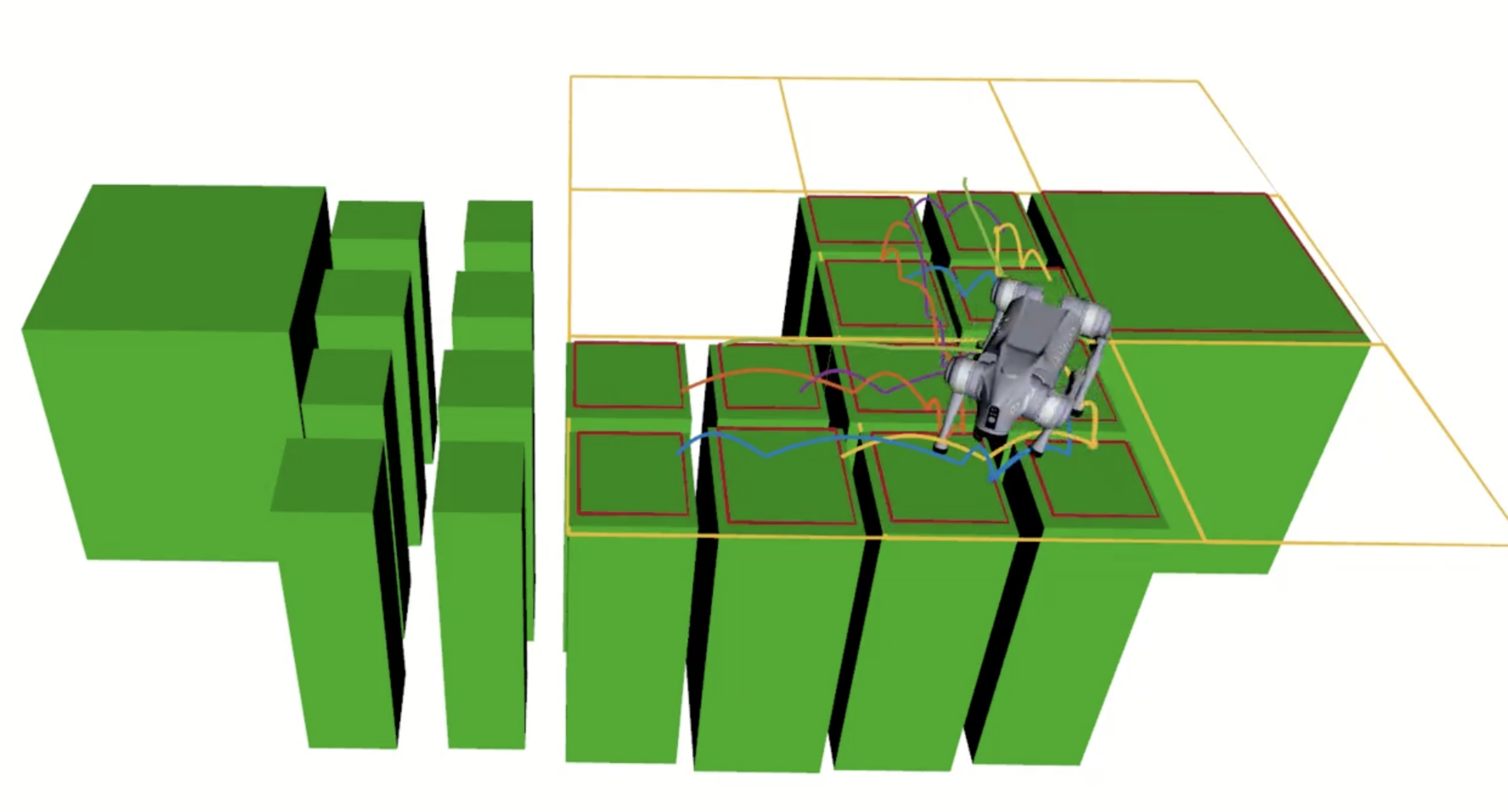}
    \caption{Demonstration of the robot executing multiple turning behaviors to continuously progress toward the global waypoint on terrain arranged in a zig-zag pattern.}
    \label{fig:yaw_extension}
\end{figure}
\revised{This subsection demonstrates that the proposed framework can be extended to non-zero yaw waypoints in principle, but we report the result as a \emph{preliminary feasibility study} only. We do not claim a full demonstration of generalizability of any angular movements. Specifically: (i) the symbolic abstraction in Sec.~\ref{sec:preliminaries_abstraction} grows combinatorially when yaw is added as a discretized input, and we have not characterized how the manager and symbolic-repair scaling numbers reported in Sec.~\ref{sec:experiments} change with yaw discretization; (ii) the angular-momentum coupling neglected by the convex-dynamics simplification in Eq.~(\ref{eq:dynamics}) becomes more relevant once yaw is treated as a planned variable, especially for fast turns; (iii) we report simulation results only---no hardware deployment of the yaw extension.}

In the previous examples, we assume that the yaw angle of the robot base is fixed to simplify the problem. However, the robot's physical capability to traverse terrains largely depends on the base orientation. For instance, performing a forward jump onto a high terrain is different from a sideway jump, in terms of both leg kinematics and dynamics.
Therefore, to demonstrate the framework's modularity, we provide a preliminary extension to the existing abstraction of the robot state to allow yaw variation.

Specifically, the robot inputs are now defined as
$\inpcen \coloneqq 
\{\var_x \mid x \in \xcells\} \cup 
\{\var_y \mid y \in \ycells\} \cup \{\var_{yaw} \mid yaw \in \mathcal{W}\}$, where we define a new set of yaw angles $\mathcal{W} \coloneqq \{-180 \degree, -90 \degree, 0 \degree, 90 \degree, 180 \degree\}$. Note that, a finer discretization of the yaw angles can be defined as needed.
Similarly, the request inputs are defined as 
$\inprequest \coloneqq
\{\var_x^\textrm{req} \mid x \in \xcells\} \cup 
\{\var_y^\textrm{req} \mid y \in \ycells\} \cup \{\var_{yaw}^\textrm{req} \mid yaw \in \mathcal{W}\}$. The robot and the request input can be further divided into translational 
$\inpcentrans, \inprequesttrans$ and rotational parts $\inpcenorien, \inprequestorien$, where the translational ones contain the $x$ and $y$ states and the orientational ones contain the $yaw$ state.

The skill is further divided into translational and rotational skills. A translational skill $\skill_{\textrm{trans}} \in \out_{\textrm{trans}}$
consists of 
sets of preconditions $\skillpretrans \subseteq \inpstatescen \times \inpstatesterrain$ that include the full robot state, and 
translational-only postconditions $\skillposttrans \subseteq \inpstatescen^{\textrm{trans}}$. A rotational skill in place $\skill_{\textrm{orien}} \in \out_{\textrm{orien}}$ consists of sets of rotational-only preconditions $\skillpreorien \subseteq \inpstatescen^{\textrm{orien}}$ and postconditions $\skillpostorien \subseteq \inpstatescen^{\textrm{orien}}$. The terrain state can also be incorporated when generating rotational skills, particularly for terrains that pose challenges for turning. However, we omit this in our current implementation for simplicity. The specification encoding then follows the same procedure as described in Sec.~\ref{sec:spec_gen}, based on each skill's precondition and postcondition. The hard constraints remain unchanged, except that translational and rotational ones are defined separately to ensure the corresponding states remain unchanged when no skill is selected.

To demonstrate the resulting maneuver, we construct a scenario with multiple dense and sparse stepping stones arranged in an overall zig-zag pattern, requiring several turning behaviors. Although sideways movement skills are feasible, we manually exclude them before synthesis to highlight yaw adjustments controlled by the rotational skills. As shown in Fig.~\ref{fig:yaw_extension}, the robot executes multiple turns to continuously progress towards the global waypoint. More complex coupled translational and rotational behaviors can also be defined by modifying the preconditions and postconditions associated with each skill.


\section{Hardware Demonstrations}\label{sec:hardware}
\begin{figure*}[t]
    \centering
    \includegraphics[width=\linewidth]{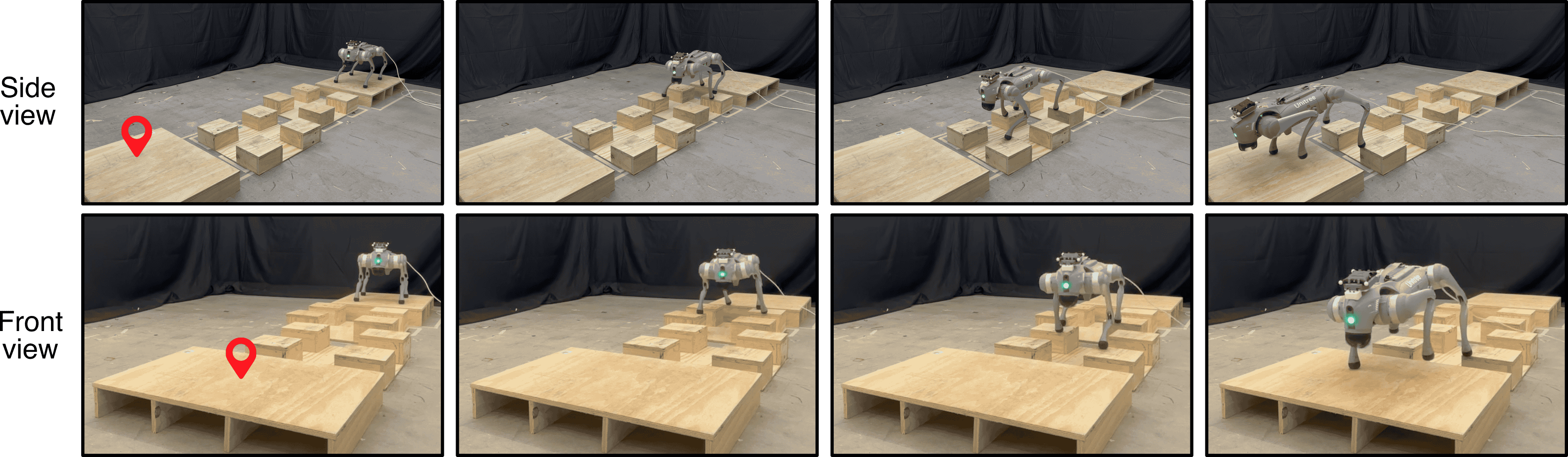}
    \caption{Hardware experiment on the first unstructured-terrain scenario: sparse stepping stones ($0.18$--$0.23$\,m apart) interleaved with flat regions. Side and front views highlight the lateral misalignment of the stones; the online MICP produces a sideways trajectory and a $3$\,s trotting gait is used throughout.}
    \label{fig:hardware_withoutgap}
\end{figure*}
We demonstrate that our entire framework can be successfully deployed on hardware, highlighting two key aspects. First, we showcase the framework’s capability to generate adaptive locomotion gaits across diverse terrain types, particularly focusing on agile leaping motions. Second, we illustrate the framework’s assured safety through obstacle avoidance and selection of dynamically feasible gaits, benchmarked against a heuristic-based planner.
\subsection{Hardware Experiment Setup}
For hardware experiments, we set up similar real-world scenarios for both unstructured and rebar terrains. For the unstructured terrain scenario, we construct wooden blocks to form stepping stones, gaps, and flat regions, with each stepping stone elevated approximately $0.12$ m above the ground. For the rebar scenario, we assemble a realistic rebar mat using fourteen $1/2$-inch $\times$ $4$-foot rebars and six $1/2$-inch × $10$-foot rebars.
In the synthesis setup, obstacles are considered a distinct terrain type, consistent with the simulation. Both scenarios assume the maximum potential terrain types—8 for unstructured and 14 for rebar—though the real setups include fewer types. As demonstrated in the simulation, this does not increase complexity, as it only scales linearly with terrain-request pairs due to our high-level manager.
For safety, we set abstraction cell sizes to $0.8$ m for the unstructured terrain and $0.45$ m for the rebar terrain.

\subsection{Unstructured Terrain}
As shown in Fig.~\ref{fig:hardware_withoutgap}, we first create a scenario with sparse stepping stones and flat terrain regions. The stepping stones are spaced between $0.18 - 0.23$ m apart. Both side and front views are provided to highlight the misalignment of stepping stones along the robot walking direction. 
The MICP planner successfully generates a reference trajectory involving non-trivial sideways movement, enabling the robot to safely land on the stepping stones. All transitions utilize a $3$ s trotting gait. 

\begin{figure}
    \centering
    \includegraphics[width=\linewidth]{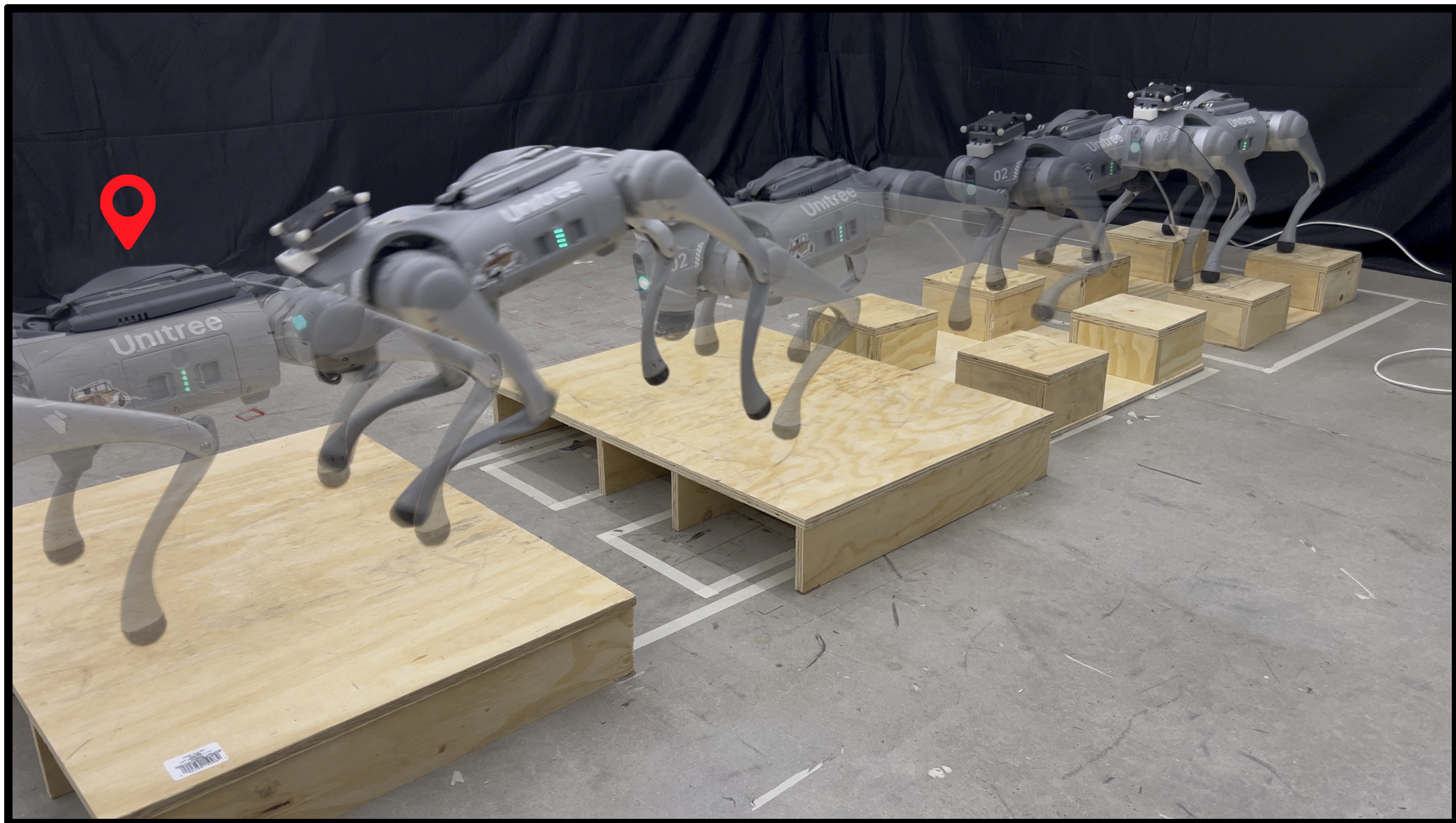}
    \caption{Hardware experiment demonstration for the second unstructured terrain scenario with sparse stepping stones, flat, and gap terrains.}
    \label{fig:hardware_withgap}
\end{figure}

\begin{figure}
    \centering    \includegraphics[width=\linewidth]{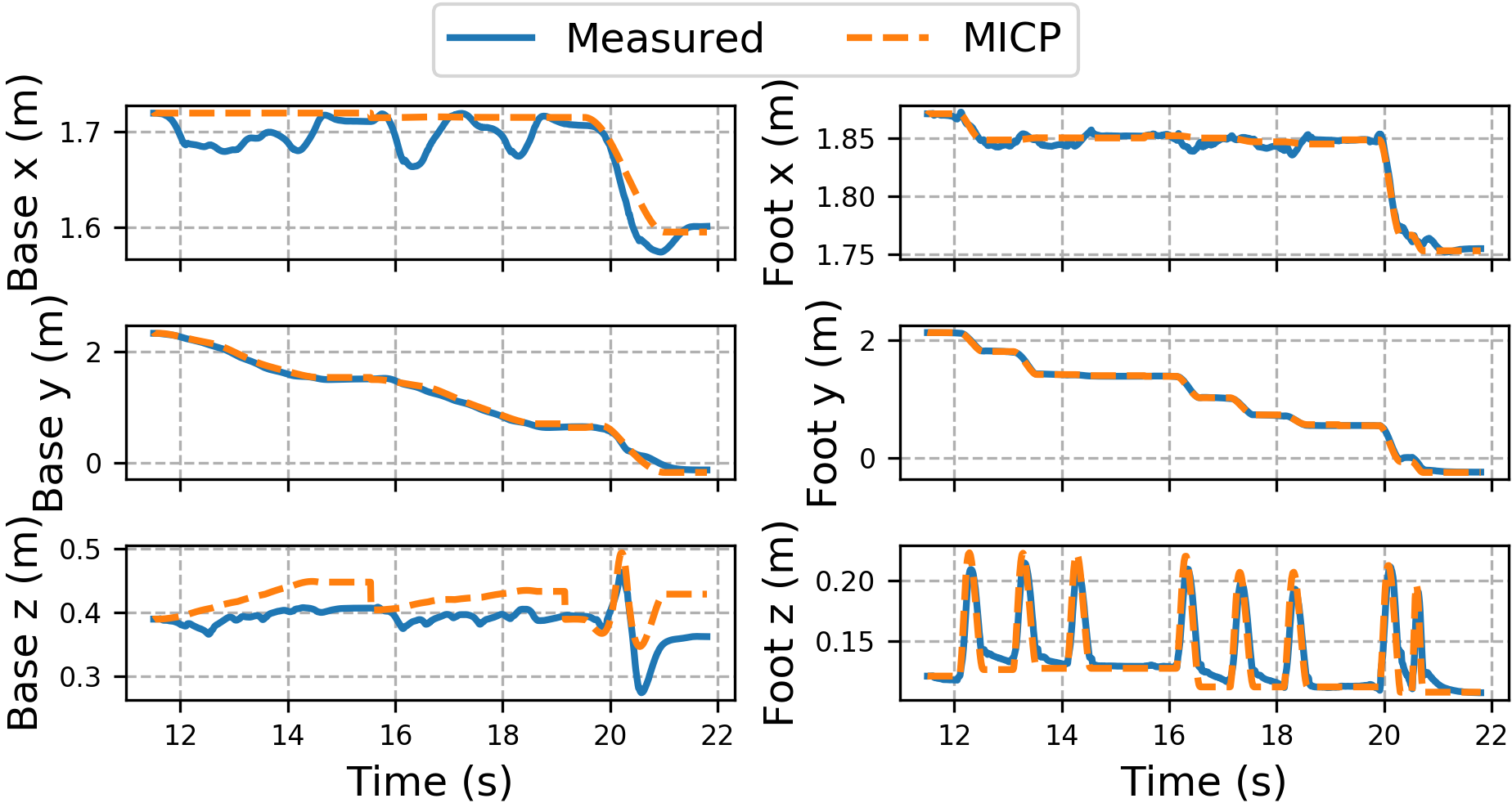}
    \caption{Tracking performance of the base and front-left foot positions: comparison between measured states and MICP-generated reference trajectories.}
    \label{fig:tracking_performance}
\end{figure}

In the second scenario, shown in Fig.~\ref{fig:hardware_withgap}, we further include a gap terrain segment. The gap, approximately 30 cm wide, necessitates the use of a leaping gait. Since the gap terrain was incorporated during offline synthesis, our framework can directly select an optimized $1.5$ s leaping gait generated by gait-free MICP upon encountering this gap terrain. The task-space tracking performance between the measured state and the reference trajectory generated by the online MICP is shown in Fig.~\ref{fig:tracking_performance}. This includes the floating base position and the end-effector (EE) position for the front-left foot. The EE tracking maintains errors within $0.01$ m in the x and y axes, ensuring precise foot placement. Meanwhile, the NMPC effectively adjusts the base position to compensate for model simplifications at the MICP level, enabling successful execution of both trotting and leaping motions.
\subsection{Rebar Terrain}\label{subsec:rebar_terrain}
\begin{figure*}[t]
    \centering
    \includegraphics[width=\linewidth]{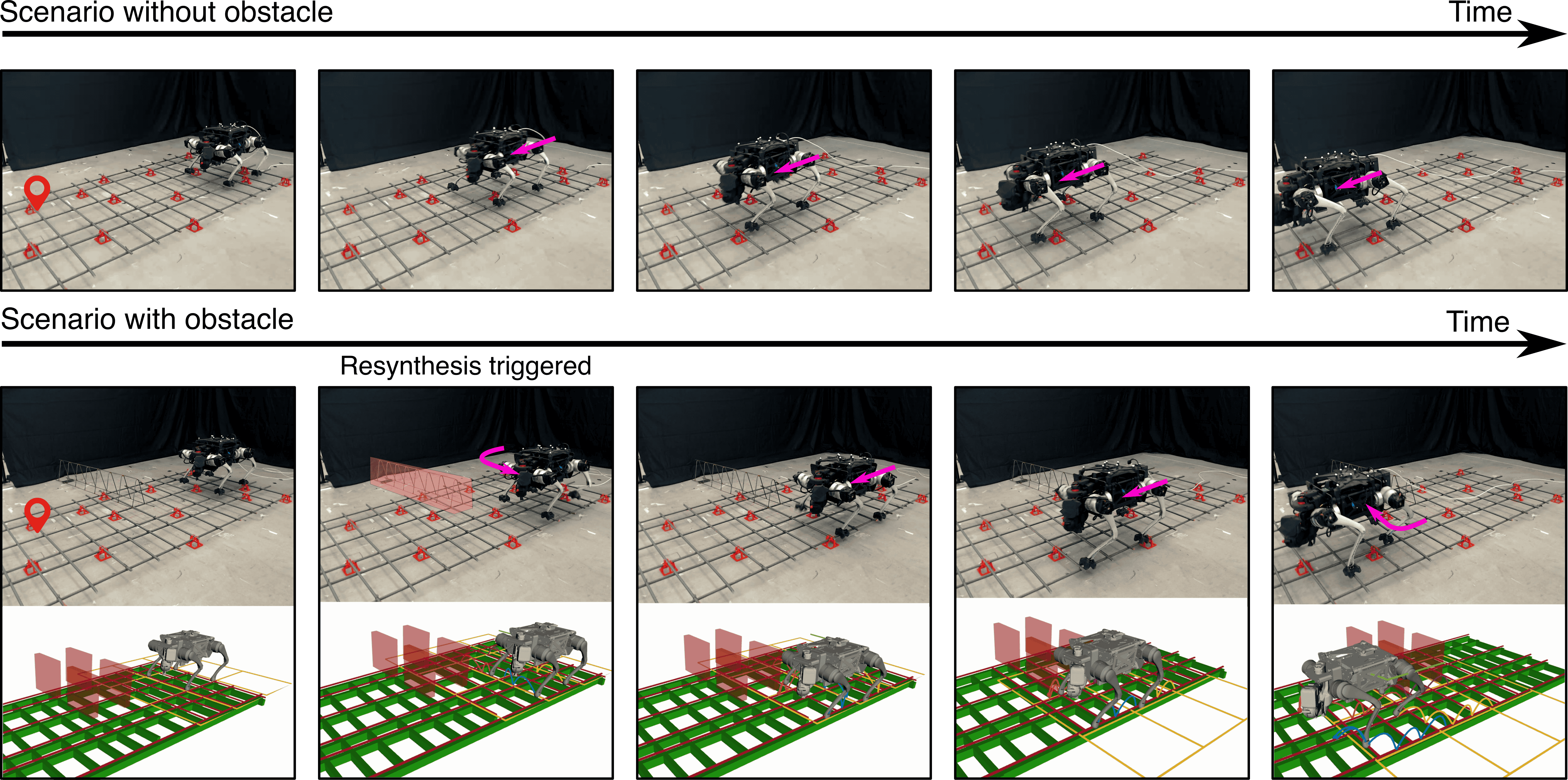}
    \caption{Execution results in rebar terrain without (top row) and with obstacle (bottom row). In the obstacle case, runtime resynthesis is triggered to generate a detour strategy and ensure successful traversal.}
    \label{fig:hardware_rebar}
\end{figure*}
For the rebar terrain scenarios shown in Fig.~\ref{fig:hardware_rebar}, we test two cases: one without any obstacles and one with an obstacle introduced during execution. The rebar spacing varies between $0.1$ m and $0.3$ m. Offline synthesis is conducted without prior knowledge of obstacles. In the first case (top row), the robot Chotu successfully performs the entire maneuver without encountering any runtime failures, consistently moving forward until reaching the goal waypoint.
In the second case (bottom row), an obstacle is added to the environment but only detected during the online execution phase. \revised{In our hardware setup, the obstacle's location is not estimated by an autonomous perception stack but rather provided through motion-capture-aided manual annotation: when the obstacle enters the robot's local field of view, its position is added to the online terrain set and the affected cell is classified as the \textit{obstacle} terrain class at the symbolic level---i.e., obstacle avoidance in this scenario is handled entirely at the symbolic level (Sec.~\ref{sec:preliminaries_abstraction}), not through the MICP-level collision-region variables $\v H_{s,w,j}$ in Eq.~(\ref{eq:collision_avoidance}). This setup is explicit about what is and is not validated---the planning and runtime-repair pipeline is exercised end-to-end, while the terrain-segmentation problem itself is not.} Upon encountering a new terrain/request pair caused by the obstacle, the planner promptly triggers a runtime resynthesis. Leveraging previously generated skills, it computes a new strategy within $0.05$ s to detour around the obstacle and continue toward the goal. All transitions in both scenarios use a $2.25$ s static walking gait, where only one foot is lifted at a time to ensure safety. \revised{We note that the gait timings adopted on hardware---$3$\,s trotting, $1.5$\,s leaping, $2.25$\,s static walking---are conservative compared to the most agile motions reported in the legged-locomotion literature; however, this locomotion gait setup still verifies the framework's gait-switching, online MICP, and runtime-repair mechanisms, which is the validation target of this article. Aggressive, high-speed gaits would require revisiting the convex-dynamics simplification (Eq.~(\ref{eq:dynamics})) discussed in Sec.~\ref{sec:online_execution}.}

\subsection{Benchmark: Heuristics-Based Planner}
\begin{figure*}
    \centering
    \includegraphics[width=\linewidth]{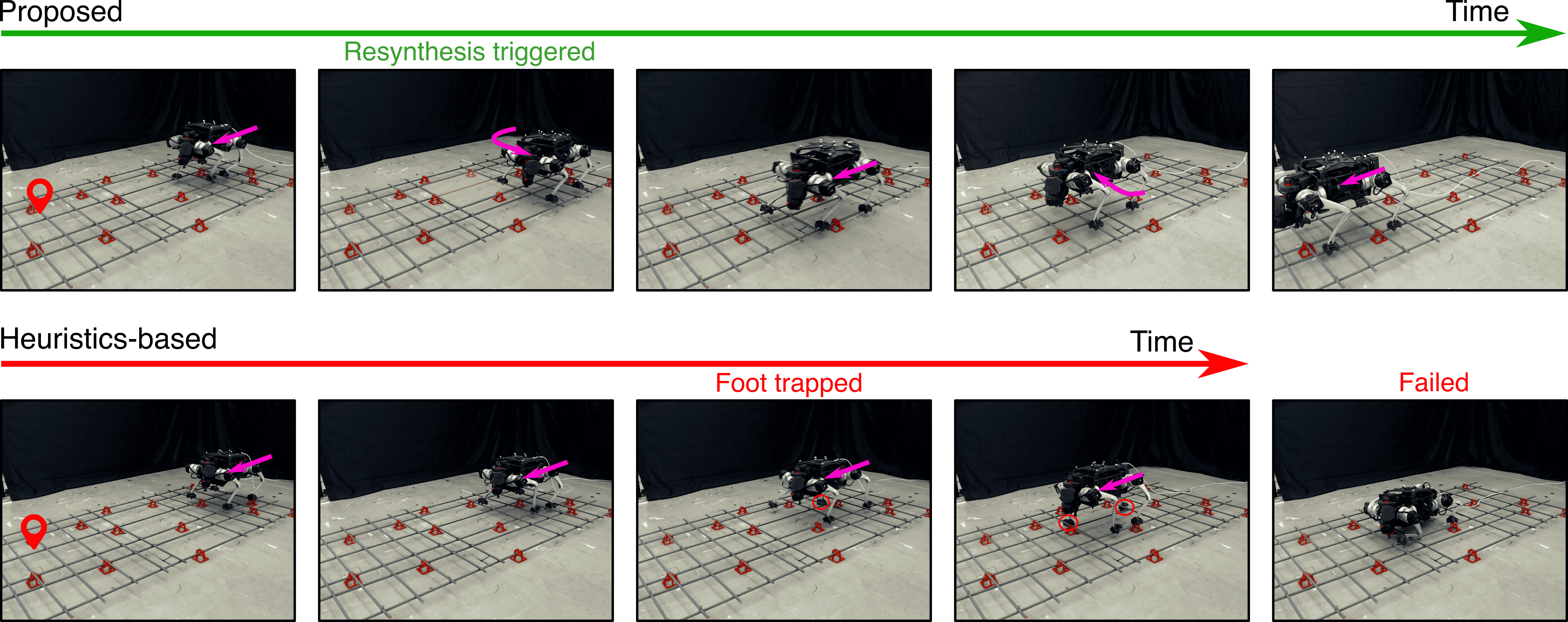}
    \caption{Comparison between the proposed method and a heuristics-based planner in an extreme sparse rebar scenario. The proposed method performs resynthesis and succeeds, while the heuristics-based planner fails due to foot trapping.}
    \label{fig:hardware_rebar_benchmark}
\end{figure*}
We create a third rebar scenario, shown in Fig.~\ref{fig:hardware_rebar_benchmark}, by removing two rebars to further increase the maximum spacing to $0.48$ m. This highlights the safety advantage of our planner compared to a heuristics-based footstep planner. The heuristic baseline follows a similar strategy to \citep{fankhauser2018robust, kim2020vision}, adapted for the rebar scenario by selecting the nearest rebar polygon and the closest feasible foothold to a nominal target. The resulting footstep plan and linearly interpolated base trajectory are sent to the same tracking control module. For fairness, all parameters are tuned to achieve successful traversal at a speed of $0.1$ m/s (as shown in Extension 1).

\revised{We chose this nearest-feasible-foothold heuristic because it isolates the value of feasibility-aware, longer-horizon planning over a single-step heuristic, and because such heuristics remain widely used in deployed perceptive-locomotion stacks due to its computational efficiency. Comparisons against more sophisticated baselines are discussed as future work in Sec.~\ref{sec:discussion}.}

Without re-performing the offline synthesis, our framework directly uses the results from Sec.~\ref{subsec:rebar_terrain}. It classifies the extreme sparse region as an obstacle and performs runtime resynthesis, generating a detour to safely reach the goal. In contrast, we task the heuristics-based planner with the same goal at $0.2$ m/s to match our planner's effective gait speed (i.e., 0.45 m per 2.25 s transition). Without reasoning about the compatibility between terrain geometry and gait feasibility, the heuristic planner suffers from poor EE tracking after a few steps, leading to foot slippage and eventual failure as the robot is entangled by the sparse rebar layout.

\subsection{Computational Time}
We report the hardware computational time in Table~\ref{tab:micp_speed_hardware}. Except for the second scenario in the unstructured terrain, we observe a notable increase in solve time compared to simulation. This is mainly due to two factors: (1) In the first unstructured terrain scenario, the two sets of stepping stones are laterally misaligned from the robot’s viewpoint, requiring non-trivial deviation from the nominal foot location and sideways movement; 
(2) In the rebar scenarios, the real-world rebar polygons are not perfectly aligned with the x or y-axis as in simulation, introducing additional numerical complexity for branch-and-bound solvers. Further acceleration of MIP solving through tighter convex relaxations \citep{marcucci2024shortest, lin2024accelerate} remains an important direction for future work.
\begin{table}
\centering
\scriptsize
    \caption{Online Gait-Fixed MICP Solving Time for Hardware}
    \setlength{\tabcolsep}{4pt}
     \begin{tabular}{lccccc}
         \hline
         \textbf{Scenario} &
         \textbf{Polygons} &
         \textbf{Min (s)} &
         \textbf{Mean (s)} &
         \revised{\textbf{Std (s)}} &
         \textbf{Max (s)}\\\hline
         Unstructured - Scene 1 & 6 & 0.81 & 4.04 & \revised{2.50} & 7.86\\
         Unstructured - Scene 2 & 5 & 0.16 & 0.53 & \revised{0.25} & 0.96\\\hline
         Rebar - Scene 1 & 8 & 3.88 & 6.54 & \revised{2.20} & 9.26\\
         Rebar - Scene 2 & 7.5 & 3.60 & 9.13 & \revised{5.50} & 19.23\\
         Rebar - Scene 3 & 6.33 & 2.45 & 7.37 & \revised{4.41} & 14.7\\\hline
    \end{tabular}
    \label{tab:micp_speed_hardware}
\end{table}

\section{Discussion}\label{sec:discussion}
\subsection{Generalization to More Types of Terrain}
The current framework relies on manual terrain abstraction and classification, which requires expert knowledge to define terrain types and assign them to discrete symbolic categories. This process typically involves analyzing physical features such as terrain shape, height variation, and support area to determine whether a terrain segment should be labeled as flat, sparse, dense, high, gap, etc. While effective, this expert-driven approach may limit scalability and generalization to novel environments or terrain conditions. Future extensions could incorporate data-driven terrain classification and generation methods, such as supervised learning or clustering based on 3D point cloud statistics to automate the terrain abstraction process and reduce reliance on human heuristics.

\subsection{Optimality}
The synthesis process focuses on guaranteeing feasibility and correctness of transitions but does not account for the relative cost or quality of different feasible gaits. When multiple locomotion gaits (e.g., trotting, bounding, leaping) are viable for a given symbolic transition, the current method does not evaluate their optimality in terms of energy efficiency, robustness, execution time, or other performance metrics. Instead, it selects the first gait that satisfies the physical feasibility conditions via MICP. Integrating cost-aware synthesis would enable the framework to prioritize higher-quality behaviors and make more informed decisions under trade-offs.

\subsection{Perception Module}
The proposed framework is modular and can be directly integrated with a wide range of perception pipelines that provide terrain geometry in the form of segmented polygonal regions \citep{miki2022elevation}. Existing perception modules that perform terrain segmentation using LiDAR, RGB-D sensors, or stereo vision can be adapted to supply the required input to the MICP planner and symbolic abstraction layers. \revised{In this work we deliberately treated the perception layer as out of scope and used motion-capture-aided manual labeling on hardware (Sec.~\ref{sec:hardware}); a fully autonomous deployment will need to address how perception uncertainty---segmentation noise, polygon misalignment, and per-frame perception delay---propagates through the MICP feasibility certificates and the symbolic-level realizability guarantees. In the current framework, runtime symbolic repair already provides a partial fallback: when perceived terrain causes the symbolic specification to become unrealizable, repair triggers and synthesizes new skills. Quantifying the resilience of this fallback under realistic perception noise is an important next step.}

\revised{\subsection{Reactivity and Update Frequencies}
Across the experiments reported in this article, the symbolic strategy automaton evolves at the timescale of one symbolic transition, which corresponds to one online MICP solve---typically hundreds of milliseconds to a few seconds (Tables~\ref{tab:micp_speed} and~\ref{tab:micp_speed_hardware}). Two complementary directions exist for tightening this loop in future work.

First, more frequent MICP updates---e.g., re-solving the online gait-fixed MICP at every NMPC tick rather than once per symbolic transition---are easily adaptable in the proposed framework: the delay-aware coordination scheme already supports asynchronous MICP solves of arbitrary duration, so any speed-up of the underlying mixed-integer solver, or any per-transition simplification, translates directly into a higher MICP update rate. The current limit is therefore the MICP solve time itself rather than any architectural constraint.

Second, more frequent symbolic-level planning is also achievable but requires careful design of the symbolic abstraction. Higher-frequency symbolic decisions would generally call for a finer abstraction---more atomic propositions per cell, finer terrain or robot-state granularity, or a larger skill repertoire. Reactive synthesis, however, scales exponentially in the number of variables in the specification~\citep{bloem2012synthesis}, so any finer abstraction must be balanced against synthesis cost. The high-level manager and symbolic repair adopted in this article already mitigate this scalability barrier; pushing toward truly real-time symbolic planning is an interesting future direction at the abstraction level rather than a limit of the framework itself.}

\revised{\subsection{Stronger Baseline Comparisons as Future Work}
The experimental validation in this article compares against a pure-MIP planner and a nearest-feasible-foothold heuristic. Two stronger comparison points that we did not implement here remain valuable future targets: (i) optimization-based planners such as TOWR-style phase-based end-effector parameterization~\citep{winkler2018gait}, which jointly optimize gait timing and footholds in continuous time; and (ii) end-to-end perceptive reinforcement-learning controllers, in particular parkour-style policies such as~\citep{zhuang2023robot,hoeller2023anymal,cheng2024extreme,luo2024pie}. A fair comparison against these two methods requires either equipping them with an additional discrete terrain selection mechanism or a comparable high-level navigation layer, which we did not pursue in the present validation.}

\revised{\subsection{Extension to Bipedal and Other Platforms}
This article focuses on quadrupedal locomotion. The framework's symbolic-level machinery---terrain abstraction, GR(1) specifications, symbolic manager, symbolic repair, runtime repair---is platform-agnostic and applicable directly to other systems such as bipedal robots. The MICP layer requires more careful adaptation: angular-momentum modulation becomes critical even at moderate speeds---a regime where the convex-dynamics simplification adopted here (Eq.~(\ref{eq:dynamics})) becomes a stronger limitation. Generalizing to humanoids would also require revisiting the kinematic-feasibility re-targeting MICP to account for arm-swing and balance constraints. We see this as a future direction.}

\section{Conclusion}
In this work, we presented an integrated planning framework for terrain-adaptive locomotion that combines reactive synthesis with MICP, enabling safe and reactive responses in dynamically changing environments. 
To address unrealizable specifications arising from limited motion primitives, we introduced a symbolic repair approach that incorporates dynamic feasibility checks, automatically identifying missing transitions necessary for navigating adversarial terrains. In the online execution phase, we tackled the disparity between offline synthesis and real-world conditions by using an online MICP solver and a runtime repair process based on real-world terrain data, including unforeseen terrain conditions. 
Our approach not only enhances motion feasibility and safety but also provides a scalable solution for legged robots maneuvering in complex and safety-critical environments. 
%


\begin{acks}
We thank Eohan George, Suphakorn Lertruchtkul, and Jane Ivanova for their hardware support on Chotu and contributions to rebar scenario design, and Pengyuan Shu for constructing the real-world terrains. The authors would also like to thank Hemanth Surya for the early exploration of the terrain segmentation.
\end{acks}

\begin{aucon}
All authors contributed to the study. Ziyi Zhou: Conceptualization,
Methodology, Formal analysis, Investigation, Software, Project
administration, Writing - original draft. Qian Meng: Methodology, Formal analysis, Investigation Software, Writing - original draft. Hadas Kress-Gazit: Methodology, Supervision, Writing – review and editing. Ye Zhao: Funding acquisition, Resources, Methodology, Supervision, Writing – review and editing. All authors reviewed and approved the final version
of the manuscript.
\end{aucon}

\begin{ethical}
Not applicable.
\end{ethical}

\begin{conpart}
Not applicable.
\end{conpart}

\begin{conpub}
Not applicable.
\end{conpub}

\begin{dci}
The author(s) declared no potential conflicts of interest with respect
to the research, authorship, and/or publication of this article.
\end{dci}

\begin{funding}
The author(s) disclosed receipt of the following financial support
for the research, authorship, and/or publication of this article:
This work was supported
through a sponsored research grant from SkyMul Inc. and a Georgia Tech
IRIM/IPaT Aware Home Seed Grant.
This work was also supported in part by the National
Science Foundation (NSF) under Grant \#CMMI-2144309, and Grant \#FRR-2328254, Office of Naval Research (ONR)
under Grant N000142312223, and in part by the United States Department of Agriculture (USDA) under Grant 2023-67021-41397.
\end{funding}

\bibliographystyle{SageH}
\bibliography{references}

@inproceedings{agarwal2023legged,
  title={Legged locomotion in challenging terrains using egocentric vision},
  author={Agarwal, Ananye and Kumar, Ashish and Malik, Jitendra and Pathak, Deepak},
  booktitle={Conference on robot learning},
  pages={403--415},
  year={2023},
  organization={PMLR}
}

@article{luo2024pie,
  title={Pie: Parkour with implicit-explicit learning framework for legged robots},
  author={Luo, Shixin and Li, Songbo and Yu, Ruiqi and Wang, Zhicheng and Wu, Jun and Zhu, Qiuguo},
  journal={IEEE Robotics and Automation Letters},
  volume={9},
  number={11},
  pages={9986--9993},
  year={2024},
  publisher={IEEE}
}

@article{miki2022learning,
  title={Learning robust perceptive locomotion for quadrupedal robots in the wild},
  author={Miki, Takahiro and Lee, Joonho and Hwangbo, Jemin and Wellhausen, Lorenz and Koltun, Vladlen and Hutter, Marco},
  journal={Science robotics},
  volume={7},
  number={62},
  pages={eabk2822},
  year={2022},
  publisher={American Association for the Advancement of Science}
}

@inproceedings{cheng2024extreme,
  title={Extreme parkour with legged robots},
  author={Cheng, Xuxin and Shi, Kexin and Agarwal, Ananye and Pathak, Deepak},
  booktitle={2024 IEEE International Conference on Robotics and Automation (ICRA)},
  pages={11443--11450},
  year={2024},
  organization={IEEE}
}

@article{zhuang2023robot,
  title={Robot parkour learning},
  author={Zhuang, Ziwen and Fu, Zipeng and Wang, Jianren and Atkeson, Christopher and Schwertfeger, Soeren and Finn, Chelsea and Zhao, Hang},
  journal={arXiv preprint arXiv:2309.05665},
  year={2023}
}

@inproceedings{zhou2025physically,
  title={Physically-Feasible Reactive Synthesis for Terrain-Adaptive Locomotion via Trajectory Optimization and Symbolic Repair},
  author={Zhou, Ziyi and Meng, Qian and Kress-Gazit, Hadas and Zhao, Ye},
  booktitle={2025 IEEE/RSJ International Conference on Intelligent Robots and Systems (IROS)},
  year={2025},
  organization={IEEE}
}

@inproceedings{wensing2013generation,
  title={Generation of dynamic humanoid behaviors through task-space control with conic optimization},
  author={Wensing, Patrick M and Orin, David E},
  booktitle={2013 IEEE International Conference on Robotics and Automation},
  pages={3103--3109},
  year={2013},
  organization={IEEE}
}

@article{zhao2024survey,
  title={A survey of optimization-based task and motion planning: From classical to learning approaches},
  author={Zhao, Zhigen and Cheng, Shuo and Ding, Yan and Zhou, Ziyi and Zhang, Shiqi and Xu, Danfei and Zhao, Ye},
  journal={IEEE/ASME Transactions on Mechatronics},
  year={2024},
  publisher={IEEE}
}

@article{drnach2021robust,
  title={Robust trajectory optimization over uncertain terrain with stochastic complementarity},
  author={Drnach, Luke and Zhao, Ye},
  journal={IEEE Robotics and Automation Letters},
  volume={6},
  number={2},
  pages={1168--1175},
  year={2021},
  publisher={IEEE}
}

@article{drnach2022mediating,
  title={Mediating between contact feasibility and robustness of trajectory optimization through chance complementarity constraints},
  author={Drnach, Luke and Zhang, John Z and Zhao, Ye},
  journal={Frontiers in Robotics and AI},
  volume={8},
  pages={785925},
  year={2022},
  publisher={Frontiers Media SA}
}

@article{wu2025learn,
  title={Learn to Teach: Sample-Efficient Privileged Learning for Humanoid Locomotion over Real-World Uneven Terrain},
  author={Wu, Feiyang and Nal, Xavier and Jang, Jaehwi and Zhu, Wei and Gu, Zhaoyuan and Wu, Anqi and Zhao, Ye},
  journal={IEEE Robotics and Automation Letters},
  year={2025},
  publisher={IEEE}
}

@article{Kamohara2025RLMPC,
  title={RL-augmented Adaptive Model Predictive Control for Bipedal Locomotion over Challenging Terrain},
  author={Kamohara, Junnosuke and Wu, Feiyang and Wamorkar, Chinmayee and Hutchinson, Seth and Zhao, Ye},
  year={2025}
}

@article{lin2024accelerate,
  title={Accelerate Hybrid Model Predictive Control using Generalized Benders Decomposition},
  author={Lin, Xuan},
  journal={arXiv preprint arXiv:2406.00780},
  year={2024}
}

@article{marcucci2024shortest,
  title={Shortest paths in graphs of convex sets},
  author={Marcucci, Tobia and Umenberger, Jack and Parrilo, Pablo and Tedrake, Russ},
  journal={SIAM Journal on Optimization},
  volume={34},
  number={1},
  pages={507--532},
  year={2024},
  publisher={SIAM}
}

@inproceedings{bledt2018cheetah,
  title={Mit cheetah 3: Design and control of a robust, dynamic quadruped robot},
  author={Bledt, Gerardo and Powell, Matthew J and Katz, Benjamin and Di Carlo, Jared and Wensing, Patrick M and Kim, Sangbae},
  booktitle={2018 IEEE/RSJ International Conference on Intelligent Robots and Systems (IROS)},
  pages={2245--2252},
  year={2018},
  organization={IEEE}
}

@article{garrett2021integrated,
  title={Integrated task and motion planning},
  author={Garrett, Caelan Reed and Chitnis, Rohan and Holladay, Rachel and Kim, Beomjoon and Silver, Tom and Kaelbling, Leslie Pack and Lozano-P{\'e}rez, Tom{\'a}s},
  journal={Annual review of control, robotics, and autonomous systems},
  volume={4},
  number={1},
  pages={265--293},
  year={2021},
  publisher={Annual Reviews}
}

@inproceedings{zhu2023efficient,
  title={Efficient object manipulation planning with monte carlo tree search},
  author={Zhu, Huaijiang and Meduri, Avadesh and Righetti, Ludovic},
  booktitle={2023 IEEE/RSJ International Conference on Intelligent Robots and Systems (IROS)},
  pages={10628--10635},
  year={2023},
  organization={IEEE}
}

@article{migimatsu2020object,
  title={Object-centric task and motion planning in dynamic environments},
  author={Migimatsu, Toki and Bohg, Jeannette},
  journal={IEEE Robotics and Automation Letters},
  volume={5},
  number={2},
  pages={844--851},
  year={2020},
  publisher={IEEE}
}

@article{sleiman2023versatile,
  title={Versatile multicontact planning and control for legged loco-manipulation},
  author={Sleiman, Jean-Pierre and Farshidian, Farbod and Hutter, Marco},
  journal={Science Robotics},
  volume={8},
  number={81},
  pages={eadg5014},
  year={2023},
  publisher={American Association for the Advancement of Science}
}

@article{toussaint2018differentiable,
  title={Differentiable physics and stable modes for tool-use and manipulation planning},
  author={Toussaint, Marc A and Allen, Kelsey Rebecca and Smith, Kevin A and Tenenbaum, Joshua B},
  year={2018},
  publisher={Robotics: Science and systems foundation}
}

@inproceedings{aydinoglu2022real,
  title={Real-time multi-contact model predictive control via admm},
  author={Aydinoglu, Alp and Posa, Michael},
  booktitle={2022 International Conference on Robotics and Automation (ICRA)},
  pages={3414--3421},
  year={2022},
  organization={IEEE}
}

@article{le2024fast,
  title={Fast contact-implicit model predictive control},
  author={Le Cleac'h, Simon and Howell, Taylor A and Yang, Shuo and Lee, Chi-Yen and Zhang, John and Bishop, Arun and Schwager, Mac and Manchester, Zachary},
  journal={IEEE Transactions on Robotics},
  year={2024},
  publisher={IEEE}
}

@inproceedings{Park2015OnlinePF,
  title={Online Planning for Autonomous Running Jumps Over Obstacles in High-Speed Quadrupeds},
  author={Hae-Won Park and Patrick M. Wensing and Sangbae Kim},
  booktitle={Robotics: Science and Systems},
  year={2015},
}

@article{yu2024learning,
  title={Learning Generic and Dynamic Locomotion of Humanoids Across Discrete Terrains},
  author={Yu, Shangqun and Perera, Nisal and Marew, Daniel and Kim, Donghyun},
  journal={arXiv preprint arXiv:2405.17227},
  year={2024}
}

@article{dai2019global,
  title={Global inverse kinematics via mixed-integer convex optimization},
  author={Dai, Hongkai and Izatt, Gregory and Tedrake, Russ},
  journal={The International Journal of Robotics Research},
  volume={38},
  number={12-13},
  pages={1420--1441},
  year={2019},
  publisher={SAGE Publications Sage UK: London, England}
}

@article{mordatch2012discovery,
  title={Discovery of complex behaviors through contact-invariant optimization},
  author={Mordatch, Igor and Todorov, Emanuel and Popovi{\'c}, Zoran},
  journal={ACM Transactions on Graphics (ToG)},
  volume={31},
  number={4},
  pages={1--8},
  year={2012},
  publisher={ACM New York, NY, USA}
}

@inproceedings{liao2023walking,
  title={Walking in narrow spaces: Safety-critical locomotion control for quadrupedal robots with duality-based optimization},
  author={Liao, Qiayuan and Li, Zhongyu and Thirugnanam, Akshay and Zeng, Jun and Sreenath, Koushil},
  booktitle={2023 IEEE/RSJ International Conference on Intelligent Robots and Systems (IROS)},
  pages={2723--2730},
  year={2023},
  organization={IEEE}
}

@inproceedings{norby2020fast,
  title={Fast global motion planning for dynamic legged robots},
  author={Norby, Joseph and Johnson, Aaron M},
  booktitle={2020 IEEE/RSJ International Conference on Intelligent Robots and Systems},
  pages={3829--3836},
  year={2020},
  organization={IEEE}
}

@article{winkler2018gait,
  title={Gait and trajectory optimization for legged systems through phase-based end-effector parameterization},
  author={Winkler, Alexander W and Bellicoso, C Dario and Hutter, Marco and Buchli, Jonas},
  journal={IEEE Robotics and Automation Letters},
  volume={3},
  number={3},
  pages={1560--1567},
  year={2018},
  publisher={IEEE}
}

@misc{gurobi,
  author = {{Gurobi Optimization, LLC}},
  title = {{Gurobi Optimizer Reference Manual}},
  year = {2025},
  howpublished = {\url{https://www.gurobi.com}}
}

@inproceedings{miki2022elevation,
  title={Elevation mapping for locomotion and navigation using gpu},
  author={Miki, Takahiro and Wellhausen, Lorenz and Grandia, Ruben and Jenelten, Fabian and Homberger, Timon and Hutter, Marco},
  booktitle={2022 IEEE/RSJ International Conference on Intelligent Robots and Systems (IROS)},
  pages={2273--2280},
  year={2022},
  organization={IEEE}
}

@inproceedings{meng2024automated,
	title = {Automated {Robot} {Recovery} from {Assumption} {Violations} of {High}-{Level} {Specifications}},
	
	language = {en},
	urldate = {2024-10-29},
	booktitle = {2024 {IEEE} 20th {International} {Conference} on {Automation} {Science} and {Engineering} ({CASE})},
	publisher = {IEEE},
	author = {Meng, Qian and Kress-Gazit, Hadas},
	month = aug,
	year = {2024},
	pages = {4154--4161},
}

@misc{OCS2,
   title = {{OCS2}:  An  open  source  library  for  Optimal  Control  of  Switched Systems},
   note = {[Online]. Available: \url{https://github.com/leggedrobotics/ocs2}},
   year = {2020},
   author = {Farbod Farshidian and others}
}

@article{sleiman2021unified,
  title={A unified mpc framework for whole-body dynamic locomotion and manipulation},
  author={Sleiman, Jean-Pierre and Farshidian, Farbod and Minniti, Maria Vittoria and Hutter, Marco},
  journal={IEEE Robotics and Automation Letters},
  volume={6},
  number={3},
  pages={4688--4695},
  year={2021},
  publisher={IEEE}
}

@inproceedings{chignoli2021humanoid,
  title={The MIT humanoid robot: Design, motion planning, and control for acrobatic behaviors},
  author={Chignoli, Matthew and Kim, Donghyun and Stanger-Jones, Elijah and Kim, Sangbae},
  booktitle={2020 IEEE-RAS 20th International Conference on Humanoid Robots (Humanoids)},
  pages={1--8},
  year={2021},
  organization={IEEE}
}

@inproceedings{dai2014whole,
  title={Whole-body motion planning with centroidal dynamics and full kinematics},
  author={Dai, Hongkai and Valenzuela, Andr{\'e}s and Tedrake, Russ},
  booktitle={2014 IEEE-RAS International Conference on Humanoid Robots},
  pages={295--302},
  year={2014},
  organization={IEEE}
}

@inproceedings{zhou2022reactive,
  title={Reactive task allocation and planning for quadrupedal and wheeled robot teaming},
  author={Zhou, Ziyi and Lee, Dong Jae and Yoshinaga, Yuki and Balakirsky, Stephen and Guo, Dejun and Zhao, Ye},
  booktitle={IEEE International Conference on Automation Science and Engineering},
  pages={2110--2117},
  year={2022},
  organization={IEEE}
}

@article{shamsah2023integrated,
  title={Integrated task and motion planning for safe legged navigation in partially observable environments},
  author={Shamsah, Abdulaziz and Gu, Zhaoyuan and Warnke, Jonas and Hutchinson, Seth and Zhao, Ye},
  journal={IEEE Transactions on Robotics},
  year={2023},
  publisher={IEEE}
}

@inproceedings{gu2022reactive,
  title={Reactive locomotion decision-making and robust motion planning for real-time perturbation recovery},
  author={Gu, Zhaoyuan and Boyd, Nathan and Zhao, Ye},
  booktitle={2022 International Conference on Robotics and Automation (ICRA)},
  pages={1896--1902},
  year={2022},
  organization={IEEE}
}

@inproceedings{pnueli1989synthesis,
  title={On the synthesis of a reactive module},
  author={Pnueli, Amir and Rosner, Roni},
  booktitle={Proceedings of the 16th ACM SIGPLAN-SIGACT symposium on Principles of programming languages},
  pages={179--190},
  year={1989}
}

@inproceedings{piterman2006synthesis,
  title={Synthesis of reactive (1) designs},
  author={Piterman, Nir and Pnueli, Amir and Sa’ar, Yaniv},
  booktitle={Verification, Model Checking, and Abstract Interpretation: 7th International Conference, VMCAI 2006, Charleston, SC, USA, January 8-10, 2006. Proceedings 7},
  pages={364--380},
  year={2006},
  organization={Springer}
}

@inproceedings{fainekos2006translating,
  title={Translating temporal logic to controller specifications},
  author={Fainekos, Georgios E and Loizou, Savvas G and Pappas, George J},
  booktitle={Proceedings of the 45th IEEE Conference on Decision and Control},
  pages={899--904},
  year={2006},
  organization={IEEE}
}

@inproceedings{bhatia2010sampling,
  title={Sampling-based motion planning with temporal goals},
  author={Bhatia, Amit and Kavraki, Lydia E and Vardi, Moshe Y},
  booktitle={2010 IEEE International Conference on Robotics and Automation},
  pages={2689--2696},
  year={2010},
  organization={IEEE}
}

@inproceedings{maly2013iterative,
  title={Iterative temporal motion planning for hybrid systems in partially unknown environments},
  author={Maly, Matthew R and Lahijanian, Morteza and Kavraki, Lydia E and Kress-Gazit, Hadas and Vardi, Moshe Y},
  booktitle={Proceedings of the 16th international conference on Hybrid systems: computation and control},
  pages={353--362},
  year={2013}
}

@article{liu2013synthesis,
  title={Synthesis of reactive switching protocols from temporal logic specifications},
  author={Liu, Jun and Ozay, Necmiye and Topcu, Ufuk and Murray, Richard M},
  journal={IEEE Transactions on Automatic Control},
  volume={58},
  number={7},
  pages={1771--1785},
  year={2013},
  publisher={IEEE}
}

@inproceedings{acosta2023bipedal,
  title={Bipedal walking on constrained footholds with mpc footstep control},
  author={Acosta, Brian and Posa, Michael},
  booktitle={2023 IEEE-RAS 22nd International Conference on Humanoid Robots},
  pages={1--8},
  year={2023},
  organization={IEEE}
}

@article{bloem2012synthesis,
  title={Synthesis of reactive (1) designs},
  author={Bloem, Roderick and Jobstmann, Barbara and Piterman, Nir and Pnueli, Amir and Saar, Yaniv},
  journal={Journal of Computer and System Sciences},
  volume={78},
  number={3},
  pages={911--938},
  year={2012},
  publisher={Elsevier}
}

@article{posa2014direct,
  title={A direct method for trajectory optimization of rigid bodies through contact},
  author={Posa, Michael and Cantu, Cecilia and Tedrake, Russ},
  journal={The International Journal of Robotics Research},
  volume={33},
  number={1},
  pages={69--81},
  year={2014},
  publisher={Sage Publications Sage UK: London, England}
}

@inproceedings{chen2021trajectotree,
  title={Trajectotree: Trajectory optimization meets tree search for planning multi-contact dexterous manipulation},
  author={Chen, Claire and Culbertson, Preston and Lepert, Marion and Schwager, Mac and Bohg, Jeannette},
  booktitle={2021 IEEE/RSJ International Conference on Intelligent Robots and Systems (IROS)},
  pages={8262--8268},
  year={2021},
  organization={IEEE}
}

@inproceedings{pacheck2020finding,
  title={Finding missing skills for high-level behaviors},
  author={Pacheck, Adam and Moarref, Salar and Kress-Gazit, Hadas},
  booktitle={2020 IEEE International Conference on Robotics and Automation (ICRA)},
  pages={10335--10341},
  year={2020},
  organization={IEEE}
}

@article{zhao2021sydebo,
  title={Sydebo: Symbolic-decision-embedded bilevel optimization for long-horizon manipulation in dynamic environments},
  author={Zhao, Zhigen and Zhou, Ziyi and Park, Michael and Zhao, Ye},
  journal={IEEE Access},
  volume={9},
  pages={128817--128826},
  year={2021},
  publisher={IEEE}
}

@inproceedings{feng2023gpf,
  title={GPF-BG: A Hierarchical Vision-Based Planning Framework for Safe Quadrupedal Navigation},
  author={Feng, Shiyu and Zhou, Ziyi and Smith, Justin S and Asselmeier, Max and Zhao, Ye and Vela, Patricio A},
  booktitle={2023 IEEE International Conference on Robotics and Automation (ICRA)},
  pages={1968--1975},
  year={2023},
  organization={IEEE}
}

@article{tonneau2018efficient,
  title={An efficient acyclic contact planner for multiped robots},
  author={Tonneau, Steve and Del Prete, Andrea and Pettr{\'e}, Julien and Park, Chonhyon and Manocha, Dinesh and Mansard, Nicolas},
  journal={IEEE Transactions on Robotics},
  volume={34},
  number={3},
  pages={586--601},
  year={2018},
  publisher={IEEE}
}

@inproceedings{fernbach2017kinodynamic,
  title={A kinodynamic steering-method for legged multi-contact locomotion},
  author={Fernbach, Pierre and Tonneau, Steve and Del Prete, Andrea and Ta{\"\i}x, Michel},
  booktitle={2017 IEEE/RSJ International Conference on Intelligent Robots and Systems (IROS)},
  pages={3701--3707},
  year={2017},
  organization={IEEE}
}

@inproceedings{wermelinger2016navigation,
  title={Navigation planning for legged robots in challenging terrain},
  author={Wermelinger, Martin and Fankhauser, P{\'e}ter and Diethelm, Remo and Kr{\"u}si, Philipp and Siegwart, Roland and Hutter, Marco},
  booktitle={2016 IEEE/RSJ International Conference on Intelligent Robots and Systems (IROS)},
  pages={1184--1189},
  year={2016},
  organization={IEEE}
}

@article{decastro2014dynamics,
  title={Dynamics-based reactive synthesis and automated revisions for high-level robot control},
  author={DeCastro, Jonathan A and Ehlers, R{\"u}diger and Rungger, Matthias and Balkan, Ayca and Tabuada, Paulo and Kress-Gazit, Hadas},
  journal={arXiv preprint arXiv:1410.6375},
  year={2014}
}

@article{zhang2023simultaneous,
  title={Simultaneous Trajectory Optimization and Contact Selection for Multi-Modal Manipulation Planning},
  author={Zhang, Mengchao and Jha, Devesh K and Raghunathan, Arvind U and Hauser, Kris},
  journal={arXiv preprint arXiv:2306.06465},
  year={2023}
}

@inproceedings{jelavic2021combined,
  title={Combined sampling and optimization based planning for legged-wheeled robots},
  author={Jelavic, Edo and Farshidian, Farbod and Hutter, Marco},
  booktitle={2021 IEEE International Conference on Robotics and Automation (ICRA)},
  pages={8366--8372},
  year={2021},
  organization={IEEE}
}

@inproceedings{chignoli2022rapid,
  title={Rapid and reliable quadruped motion planning with omnidirectional jumping},
  author={Chignoli, Matthew and Morozov, Savva and Kim, Sangbae},
  booktitle={International Conference on Robotics and Automation},
  pages={6621--6627},
  year={2022},
  organization={IEEE}
}

@inproceedings{ding2021hybrid,
  title={Hybrid sampling/optimization-based planning for agile jumping robots on challenging terrains},
  author={Ding, Yanran and Zhang, Mengchao and Li, Chuanzheng and Park, Hae-Won and Hauser, Kris},
  booktitle={2021 IEEE International Conference on Robotics and Automation (ICRA)},
  pages={2839--2845},
  year={2021},
  organization={IEEE}
}

@article{jiang2023locomotion,
  title={Locomotion generation for quadruped robots on challenging terrains via quadratic programming},
  author={Jiang, Xinyang and Chi, Wanchao and Zheng, Yu and Zhang, Shenghao and Ling, Yonggen and Xu, Jiafeng and Zhang, Zhengyou},
  journal={Autonomous Robots},
  volume={47},
  number={1},
  pages={51--76},
  year={2023},
  publisher={Springer}
}

@inproceedings{shirai2022simultaneous,
  title={Simultaneous contact-rich grasping and locomotion via distributed optimization enabling free-climbing for multi-limbed robots},
  author={Shirai, Yuki and Lin, Xuan and Schperberg, Alexander and Tanaka, Yusuke and Kato, Hayato and Vichathorn, Varit and Hong, Dennis},
  booktitle={2022 IEEE/RSJ International Conference on Intelligent Robots and Systems (IROS)},
  pages={13563--13570},
  year={2022},
  organization={IEEE}
}

@inproceedings{risbourg2022real,
  title={Real-time Footstep Planning and Control of the Solo Quadruped Robot in 3D Environments},
  author={Risbourg, Fanny and Corb{\`e}res, Thomas and L{\'e}ziart, Pierre-Alexandre and Flayols, Thomas and Mansard, Nicolas and Tonneau, Steve},
  booktitle={IEEE/RSJ International Conference on Intelligent Robots and Systems},
  pages={12950--12956},
  year={2022}
}

@inproceedings{tonneau2020sl1m,
  title={SL1M: Sparse L1-norm Minimization for contact planning on uneven terrain},
  author={Tonneau, Steve and Song, Daeun and Fernbach, Pierre and Mansard, Nicolas and Ta{\"\i}x, Michel and Del Prete, Andrea},
  booktitle={2020 IEEE International Conference on Robotics and Automation (ICRA)},
  pages={6604--6610},
  year={2020},
  organization={IEEE}
}

@inproceedings{ding2020kinodynamic,
  title={Kinodynamic motion planning for multi-legged robot jumping via mixed-integer convex program},
  author={Ding, Yanran and Li, Chuanzheng and Park, Hae-Won},
  booktitle={2020 IEEE/RSJ International Conference on Intelligent Robots and Systems (IROS)},
  pages={3998--4005},
  year={2020},
  organization={IEEE}
}

@article{aceituno2017simultaneous,
  title={Simultaneous contact, gait, and motion planning for robust multilegged locomotion via mixed-integer convex optimization},
  author={Aceituno-Cabezas, Bernardo and Mastalli, Carlos and Dai, Hongkai and Focchi, Michele and Radulescu, Andreea and Caldwell, Darwin G and Cappelletto, Jos{\'e} and Grieco, Juan C and Fern{\'a}ndez-L{\'o}pez, Gerardo and Semini, Claudio},
  journal={IEEE Robotics and Automation Letters},
  volume={3},
  number={3},
  pages={2531--2538},
  year={2017},
  publisher={IEEE}
}

@inproceedings{ponton2016convex,
  title={A convex model of humanoid momentum dynamics for multi-contact motion generation},
  author={Ponton, Brahayam and Herzog, Alexander and Schaal, Stefan and Righetti, Ludovic},
  booktitle={2016 IEEE-RAS 16th International Conference on Humanoid Robots (Humanoids)},
  pages={842--849},
  year={2016},
  organization={IEEE}
}

@inproceedings{deits2014footstep,
  title={Footstep planning on uneven terrain with mixed-integer convex optimization},
  author={Deits, Robin and Tedrake, Russ},
  booktitle={2014 IEEE-RAS international conference on humanoid robots},
  pages={279--286},
  year={2014},
  organization={IEEE}
}

@phdthesis{valenzuela2016mixed,
  title={Mixed-integer convex optimization for planning aggressive motions of legged robots over rough terrain},
  author={Valenzuela, Andr{\'e}s Klee},
  year={2016},
  school={Massachusetts Institute of Technology}
}

@article{hoeller2023anymal,
  title={ANYmal Parkour: Learning Agile Navigation for Quadrupedal Robots},
  author={Hoeller, David and Rudin, Nikita and Sako, Dhionis and Hutter, Marco},
  journal={arXiv preprint arXiv:2306.14874},
  year={2023}
}

@article{sun2023free,
  title={Free Gait Generation of Quadruped Robots via Impulse-Based Feasibility Analysis},
  author={Sun, Hao and Yang, Junjie and Jia, Yinghao and Wang, Changhong},
  journal={IEEE/ASME Transactions on Mechatronics},
  year={2023},
  publisher={IEEE}
}

@article{boussema2019online,
  title={Online gait transitions and disturbance recovery for legged robots via the feasible impulse set},
  author={Boussema, Chiheb and Powell, Matthew J and Bledt, Gerardo and Ijspeert, Auke J and Wensing, Patrick M and Kim, Sangbae},
  journal={IEEE Robotics and automation letters},
  volume={4},
  number={2},
  pages={1611--1618},
  year={2019},
  publisher={IEEE}
}

@article{wang2023and,
  title={When and Where to Step: Terrain-Aware Real-Time Footstep Location and Timing Optimization for Bipedal Robots},
  author={Wang, Ke and Hu, Zhaoyang Jacopo and Tisnikar, Peter and Helander, Oskar and Chappell, Digby and Kormushev, Petar},
  journal={arXiv preprint arXiv:2302.07345},
  year={2023}
}

@inproceedings{yang2022fast,
  title={Fast and efficient locomotion via learned gait transitions},
  author={Yang, Yuxiang and Zhang, Tingnan and Coumans, Erwin and Tan, Jie and Boots, Byron},
  booktitle={Conference on Robot Learning},
  pages={773--783},
  year={2022},
  organization={PMLR}
}

@inproceedings{xie2022glide,
  title={Glide: Generalizable quadrupedal locomotion in diverse environments with a centroidal model},
  author={Xie, Zhaoming and Da, Xingye and Babich, Buck and Garg, Animesh and de Panne, Michiel van},
  booktitle={International Workshop on the Algorithmic Foundations of Robotics},
  pages={523--539},
  year={2022},
  organization={Springer}
}

@article{ren2025accelerating,
  title={Accelerating Signal-Temporal-Logic-Based Task and Motion Planning of Bipedal Navigation using Benders Decomposition},
  author={Ren, Jiming and Lin, Xuan and Mineyev, Roman and Feigh, Karen M and Coogan, Samuel and Zhao, Ye},
  journal={arXiv preprint arXiv:2508.13407},
  year={2025}
}

@article{jiang2023abstraction,
  title={Abstraction-based planning for uncertainty-aware legged navigation},
  author={Jiang, Jesse and Coogan, Samuel and Zhao, Ye},
  journal={IEEE Open Journal of Control Systems},
  year={2023},
  publisher={IEEE}
}

@inproceedings{muenprasitivej2024bipedal,
  title={Bipedal safe navigation over uncertain rough terrain: Unifying terrain mapping and locomotion stability},
  author={Muenprasitivej, Kasidit and Jiang, Jesse and Shamsah, Abdulaziz and Coogan, Samuel and Zhao, Ye},
  booktitle={2024 IEEE/RSJ International Conference on Intelligent Robots and Systems (IROS)},
  pages={11264--11271},
  year={2024},
  organization={IEEE}
}

@inproceedings{muenprasitivej2026bipedal,
  title={Probabilistically-Safe Bipedal Navigation over Uncertain Terrain via Conformal Prediction and Contraction Analysis},
  author={Muenprasitivej, Kasidit  and Zhao, Ye and Chou, Glen},
  year={2025}
}

@inproceedings{shamsah2025terrain,
  title={Terrain-aware model predictive control of heterogeneous bipedal and aerial robot coordination for search and rescue tasks},
  author={Shamsah, Abdulaziz and Jiang, Jesse and Yoon, Ziwon and Coogan, Samuel and Zhao, Ye},
  booktitle={2025 IEEE International Conference on Robotics and Automation (ICRA)},
  pages={12352--12358},
  year={2025},
  organization={IEEE}
}

@inproceedings{da2021learning,
  title={Learning a contact-adaptive controller for robust, efficient legged locomotion},
  author={Da, Xingye and Xie, Zhaoming and Hoeller, David and Boots, Byron and Anandkumar, Anima and Zhu, Yuke and Babich, Buck and Garg, Animesh},
  booktitle={Conference on Robot Learning},
  pages={883--894},
  year={2021},
  organization={PMLR}
}

@article{magana2019fast,
  title={Fast and continuous foothold adaptation for dynamic locomotion through cnns},
  author={Magana, Octavio Antonio Villarreal and Barasuol, Victor and Camurri, Marco and Franceschi, Luca and Focchi, Michele and Pontil, Massimiliano and Caldwell, Darwin G and Semini, Claudio},
  journal={IEEE Robotics and Automation Letters},
  volume={4},
  number={2},
  pages={2140--2147},
  year={2019},
  publisher={IEEE}
}

@InProceedings{pmlr-v164-margolis22a,
  title={Learning to Jump from Pixels},
  author={Margolis, Gabriel B and Chen, Tao and Paigwar, Kartik and Fu, Xiang and Kim, Donghyun and Kim, Sang bae and Agrawal, Pulkit},
  booktitle={Proceedings of the 5th Conference on Robot Learning},
  pages={1025--1034},
  year={2022},
  publisher={PMLR},
}

@inproceedings{griffin2019footstep,
  title={Footstep planning for autonomous walking over rough terrain},
  author={Griffin, Robert J and Wiedebach, Georg and McCrory, Stephen and Bertrand, Sylvain and Lee, Inho and Pratt, Jerry},
  booktitle={2019 IEEE-RAS 19th international conference on humanoid robots (humanoids)},
  pages={9--16},
  year={2019},
  organization={IEEE}
}

@inproceedings{mastalli2015line,
  title={On-line and on-board planning and perception for quadrupedal locomotion},
  author={Mastalli, Carlos and Havoutis, Ioannis and Winkler, Alexander W and Caldwell, Darwin G and Semini, Claudio},
  booktitle={2015 IEEE International Conference on Technologies for Practical Robot Applications (TePRA)},
  pages={1--7},
  year={2015},
  organization={IEEE}
}

@inproceedings{winkler2014path,
  title={Path planning with force-based foothold adaptation and virtual model control for torque controlled quadruped robots},
  author={Winkler, Alexander and Havoutis, Ioannis and Bazeille, Stephane and Ortiz, Jesus and Focchi, Michele and Dillmann, R{\"u}diger and Caldwell, Darwin and Semini, Claudio},
  booktitle={2014 IEEE International Conference on Robotics and Automation (ICRA)},
  pages={6476--6482},
  year={2014},
  organization={IEEE}
}

@inproceedings{kalakrishnan2010fast,
  title={Fast, robust quadruped locomotion over challenging terrain},
  author={Kalakrishnan, Mrinal and Buchli, Jonas and Pastor, Peter and Mistry, Michael and Schaal, Stefan},
  booktitle={2010 IEEE International Conference on Robotics and Automation},
  pages={2665--2670},
  year={2010},
  organization={IEEE}
}

@inproceedings{kolter2008control,
  title={A control architecture for quadruped locomotion over rough terrain},
  author={Kolter, J Zico and Rodgers, Mike P and Ng, Andrew Y},
  booktitle={2008 IEEE International Conference on Robotics and Automation},
  pages={811--818},
  year={2008},
  organization={IEEE}
}

@article{pacheck2022physically,
  title={Physically feasible repair of reactive, linear temporal logic-based, high-level tasks},
  author={Pacheck, Adam and Kress-Gazit, Hadas},
  journal={IEEE Transactions on Robotics},
  year={2023},
  publisher={IEEE}
}

@article{kress2018synthesis,
  title={Synthesis for robots: Guarantees and feedback for robot behavior},
  author={Kress-Gazit, Hadas and Lahijanian, Morteza and Raman, Vasumathi},
  journal={Annual Review of Control, Robotics, and Autonomous Systems},
  volume={1},
  pages={211--236},
  year={2018},
  publisher={Annual Reviews}
}

@article{bretl2006motion,
  title={Motion planning of multi-limbed robots subject to equilibrium constraints: The free-climbing robot problem},
  author={Bretl, Timothy},
  journal={The International Journal of Robotics Research},
  volume={25},
  number={4},
  pages={317--342},
  year={2006},
  publisher={SAGE Publications}
}

@inproceedings{hauser2005non,
  title={Non-gaited humanoid locomotion planning},
  author={Hauser, Kris and Bretl, Timothy and Latombe, J-C},
  booktitle={5th IEEE-RAS International Conference on Humanoid Robots, 2005.},
  pages={7--12},
  year={2005},
  organization={IEEE}
}

@inproceedings{kuindersma2014efficiently,
  title={An efficiently solvable quadratic program for stabilizing dynamic locomotion},
  author={Kuindersma, Scott and Permenter, Frank and Tedrake, Russ},
  booktitle={2014 IEEE International Conference on Robotics and Automation (ICRA)},
  pages={2589--2594},
  year={2014},
  organization={IEEE}
}

@article{fey20243d,
  title={3d hopping in discontinuous terrain using impulse planning with mixed-integer strategies},
  author={Fey, Nolan and Frei, Robert J and Wensing, Patrick M},
  journal={IEEE Robotics and Automation Letters},
  year={2024},
  publisher={IEEE}
}

@article{zhao2022reactive,
  title={Reactive task and motion planning for robust whole-body dynamic locomotion in constrained environments},
  author={Zhao, Ye and Li, Yinan and Sentis, Luis and Topcu, Ufuk and Liu, Jun},
  journal={The International Journal of Robotics Research},
  pages={02783649221077714},
  year={2022},
  publisher={SAGE Publications Sage UK: London, England}
}

@inproceedings{asselmeier2024hierarchical,
  title={Hierarchical experience-informed navigation for multi-modal quadrupedal rebar grid traversal},
  author={Asselmeier, Max and Ivanova, Jane and Zhou, Ziyi and Vela, Patricio A and Zhao, Ye},
  booktitle={2024 IEEE International Conference on Robotics and Automation (ICRA)},
  pages={8065--8072},
  year={2024},
  organization={IEEE}
}

@inproceedings{di2018dynamic,
  title={Dynamic locomotion in the mit cheetah 3 through convex model-predictive control},
  author={Di Carlo, Jared and Wensing, Patrick M and Katz, Benjamin and Bledt, Gerardo and Kim, Sangbae},
  booktitle={2018 IEEE/RSJ international conference on intelligent robots and systems (IROS)},
  pages={1--9},
  year={2018},
  organization={IEEE}
}

@inproceedings{kulgod2020temporal,
  title={Temporal logic guided locomotion planning and control in cluttered environments},
  author={Kulgod, Sutej and Chen, Wentao and Huang, Junda and Zhao, Ye and Atanasov, Nikolay},
  booktitle={American Control Conference},
  pages={5425--5432},
  year={2020},
  organization={IEEE}
}

@article{kress2009temporal,
  title={Temporal-logic-based reactive mission and motion planning},
  author={Kress-Gazit, Hadas and Fainekos, Georgios E and Pappas, George J},
  journal={IEEE transactions on robotics},
  volume={25},
  number={6},
  pages={1370--1381},
  year={2009},
  publisher={IEEE}
}

@article{zhou2022momentum,
  title={Momentum-aware trajectory optimization and control for agile quadrupedal locomotion},
  author={Zhou, Ziyi and Wingo, Bruce and Boyd, Nathan and Hutchinson, Seth and Zhao, Ye},
  journal={IEEE Robotics and Automation Letters},
  volume={7},
  number={3},
  pages={7755--7762},
  year={2022},
  publisher={IEEE}
}

@book{baier2008principles,
  title={Principles of model checking},
  author={Baier, Christel and Katoen, Joost-Pieter},
  year={2008},
  publisher={MIT press}
}

@article{kim2019highly,
  title={Highly dynamic quadruped locomotion via whole-body impulse control and model predictive control},
  author={Kim, Donghyun and Di Carlo, Jared and Katz, Benjamin and Bledt, Gerardo and Kim, Sangbae},
  journal={arXiv preprint arXiv:1909.06586},
  year={2019}
}

@inproceedings{toussaint2015logic,
  title={Logic-Geometric Programming: An Optimization-Based Approach to Combined Task and Motion Planning.},
  author={Toussaint, Marc},
  booktitle={IJCAI},
  pages={1930--1936},
  year={2015}
}

@inproceedings{amatucci2022monte,
  title={Monte carlo tree search gait planner for non-gaited legged system control},
  author={Amatucci, Lorznzo and Kim, Joon-Ha and Hwangbo, Jemin and Park, Hae-Won},
  booktitle={2022 International Conference on Robotics and Automation (ICRA)},
  pages={4701--4707},
  year={2022},
  organization={IEEE}
}

@article{jenelten2022tamols,
  title={TAMOLS: Terrain-aware motion optimization for legged systems},
  author={Jenelten, Fabian and Grandia, Ruben and Farshidian, Farbod and Hutter, Marco},
  journal={IEEE Transactions on Robotics},
  volume={38},
  number={6},
  pages={3395--3413},
  year={2022},
  publisher={IEEE}
}

@article{grandia2023perceptive,
  title={Perceptive Locomotion Through Nonlinear Model-Predictive Control},
  author={Grandia, Ruben and Jenelten, Fabian and Yang, Shaohui and Farshidian, Farbod and Hutter, Marco},
  journal={IEEE Transactions on Robotics},
  year={2023},
  publisher={IEEE}
}

@article{gangapurwala2022rloc,
  title={Rloc: Terrain-aware legged locomotion using reinforcement learning and optimal control},
  author={Gangapurwala, Siddhant and Geisert, Mathieu and Orsolino, Romeo and Fallon, Maurice and Havoutis, Ioannis},
  journal={IEEE Transactions on Robotics},
  volume={38},
  number={5},
  pages={2908--2927},
  year={2022},
  publisher={IEEE}
}

@inproceedings{gilroy2021autonomous,
  title={Autonomous navigation for quadrupedal robots with optimized jumping through constrained obstacles},
  author={Gilroy, Scott and Lau, Derek and Yang, Lizhi and Izaguirre, Ed and Biermayer, Kristen and Xiao, Anxing and Sun, Mengti and Agrawal, Ayush and Zeng, Jun and Li, Zhongyu and others},
  booktitle={2021 IEEE 17th International Conference on Automation Science and Engineering (CASE)},
  pages={2132--2139},
  year={2021},
  organization={IEEE}
}

@inproceedings{agrawal2022vision,
  title={Vision-aided dynamic quadrupedal locomotion on discrete terrain using motion libraries},
  author={Agrawal, Ayush and Chen, Shuxiao and Rai, Akshara and Sreenath, Koushil},
  booktitle={2022 International Conference on Robotics and Automation (ICRA)},
  pages={4708--4714},
  year={2022},
  organization={IEEE}
}

@inproceedings{fankhauser2018robust,
  title={Robust rough-terrain locomotion with a quadrupedal robot},
  author={Fankhauser, P{\'e}ter and Bjelonic, Marko and Bellicoso, C Dario and Miki, Takahiro and Hutter, Marco},
  booktitle={2018 IEEE International Conference on Robotics and Automation (ICRA)},
  pages={5761--5768},
  year={2018},
  organization={IEEE}
}

@article{jenelten2020perceptive,
  title={Perceptive locomotion in rough terrain--online foothold optimization},
  author={Jenelten, Fabian and Miki, Takahiro and Vijayan, Aravind E and Bjelonic, Marko and Hutter, Marco},
  journal={IEEE Robotics and Automation Letters},
  volume={5},
  number={4},
  pages={5370--5376},
  year={2020},
  publisher={IEEE}
}

@inproceedings{kim2020vision,
  title={Vision aided dynamic exploration of unstructured terrain with a small-scale quadruped robot},
  author={Kim, Donghyun and Carballo, Daniel and Di Carlo, Jared and Katz, Benjamin and Bledt, Gerardo and Lim, Bryan and Kim, Sangbae},
  booktitle={IEEE International Conference on Robotics and Automation},
  pages={2464--2470},
  year={2020},
  organization={IEEE}
}

@inproceedings{villarreal2020mpc,
  title={Mpc-based controller with terrain insight for dynamic legged locomotion},
  author={Villarreal, Octavio and Barasuol, Victor and Wensing, Patrick M and Caldwell, Darwin G and Semini, Claudio},
  booktitle={2020 IEEE International Conference on Robotics and Automation (ICRA)},
  pages={2436--2442},
  year={2020},
  organization={IEEE}
}

@inproceedings{grandia2021multi,
  title={Multi-layered safety for legged robots via control barrier functions and model predictive control},
  author={Grandia, Ruben and Taylor, Andrew J and Ames, Aaron D and Hutter, Marco},
  booktitle={2021 IEEE International Conference on Robotics and Automation (ICRA)},
  pages={8352--8358},
  year={2021},
  organization={IEEE}
}

@inproceedings{yu2021visual,
  title={Visual-locomotion: Learning to walk on complex terrains with vision},
  author={Yu, Wenhao and Jain, Deepali and Escontrela, Alejandro and Iscen, Atil and Xu, Peng and Coumans, Erwin and Ha, Sehoon and Tan, Jie and Zhang, Tingnan},
  booktitle={5th Annual Conference on Robot Learning},
  year={2021}
}

@inproceedings{ehlers2016slugs,
  title={Slugs: Extensible \uppercase{GR(1)} synthesis},
  author={Ehlers, Rudiger and Raman, Vasumathi},
  booktitle={International Conference on Computer Aided Verification},
  pages={333--339},
  year={2016},
  organization={Springer}
}

\appendix
\section{Index to Multimedia Extensions}

Archives of IJRR multimedia extensions published prior to 2014 can be found at \url{http://www.ijrr.org}; after 2014 all videos are available on the IJRR YouTube channel at \url{http://www.youtube.com/user/ijrrmultimedia}. A project page with all videos and supporting code is also maintained at \url{https://synthesis-micp.github.io/}.

\vspace{0.5em}
\noindent\textbf{Table of Multimedia Extensions}

\vspace{0.3em}
{\footnotesize
\setlength{\tabcolsep}{3pt}
\noindent\begin{tabular}{@{}c l p{0.65\columnwidth}@{}}
\hline
\textbf{Extension} & \textbf{Media Type} & \textbf{Description} \\
\hline
1 & Video & Demonstrations of experimental results, including:
\newline (1) Simulation maneuvers across unstructured terrain (4-type and 8-type) on the Unitree Go2 and rebar terrain (7-type and 14-type) on the SkyMul Chotu;
\newline (2) Simulation runtime-repair case study on an unforeseen \textit{gap} terrain state in the \textit{Unstructured-4}/$3\times 3$ scenario;
\newline (3) Hardware experiments on unstructured terrain with sparse stepping stones, flat regions, and a $30$\,cm gap, using $3$\,s trotting and $1.5$\,s leaping gaits on the Unitree Go2;
\newline (4) Hardware experiments on a rebar mat with and without a runtime-detected obstacle on the SkyMul Chotu, with runtime re-synthesis producing a detour in the obstacle case;
\newline (5) Hardware benchmark on an extreme-sparse rebar scenario against a heuristics-based footstep planner. \\
\hline
\end{tabular}
}

\end{document}